\documentclass{adobe_research}
\usepackage[T1]{fontenc}
\usepackage{lmodern}
\usepackage{multirow}
\usepackage{array}
\usepackage{soul}
\usepackage{pifont}
\usepackage[T1]{fontenc}

\usepackage{graphicx}
\usepackage{amsmath,amssymb}

\usepackage{makecell}
\usepackage[dvipsnames]{xcolor}
\usepackage{color, colortbl}
\usepackage{caption}
\usepackage{algorithm}

\usepackage{xspace}
\makeatletter
\DeclareRobustCommand\onedot{\futurelet\@let@token\@onedot}
\def\@onedot{\ifx\@let@token.\else.\null\fi\xspace}

\def\eg{\emph{e.g}\onedot}

\makeatother

\newcommand{\heterop}{\texttt{HeteroP}}
\newcommand{\base}[1]{#1_{\mathrm{base}}}
\newcommand{\mN}{m_{W}}
\newcommand{\mNinv}{m_{W}^{-1}}
\newcommand{\mL}[1]{m_{L}^{#1}}
\definecolor{tuneblue}{HTML}{1F4E79}

\usepackage{newtxtext}
\usepackage{wrapfig}

\usepackage{algpseudocode}
\definecolor{catgray}{gray}{0.92}

\usepackage[dvipsnames]{xcolor} 
\definecolor{FutureOrange}{HTML}{EC866D}
\newcommand{\ours}{\textcolor{FutureOrange}{\textbf{Chimera}}}
\newcommand{\algours}{\operatorname{\texttt{Chimera}}}

\title{\ours{}: Designing and Chinchilla-Scaling Hybrid Visual Diffusion Transformers}

\author[*\S\dagger]{Chongjian Ge}
\author[*\S]{Hanwen Jiang}
\author[*\S]{Tianyu Wang}
\author[\S]{Jiuxiang Gu}
\author[\S]{Yiran Xu}
\author[\S]{Ziwen Chen}
\author[]{\\Shaoteng Liu}
\author[]{Jing Shi}
\author[]{Yicong Hong}
\author[]{Zefan Cai}
\author[]{Hailin Jin}
\author[\dagger]{Hao Tan}

\affiliation[]{\textbf{Adobe Research}}

\contribution[*]{Equal Contribution}
\contribution[\S]{Core Contribution}
\contribution[\dagger]{Project Lead}

\abstract{
Visual generation is entering a token-intensive regime: high-resolution images, long videos, and multimodal context make the quadratic cost of full attention in Diffusion Transformers increasingly prohibitive. 
Language models underwent a similar transition, but solutions developed for them cannot be transferred directly to visual diffusion models, which must preserve spatiotemporal locality and support bidirectional interactions across modalities. 
We present \ours{}, a hybrid visual diffusion backbone co-designed with a principled scaling recipe. \ours{} processes text, images, and video tokens in a single stream with three complementary mechanisms:
Kimi Delta Attention (KDA) provides long-context state tracking with efficient  $\mathcal{O}(N)$ computation complexity, interleaved Multi-head Latent Attention (MLA) enables direct global interaction, and modality-aware short convolutions capture local spatiotemporal context. 
Together, these components enable processing of a unified raster-ordered token sequence free of positional embeddings. Sparse Mixture-of-Experts (MoE) layers further expand model capacity while keeping activated compute under control. 
To scale this heterogeneous architecture, we introduce \heterop{}, a module-wise scaling scheme that transfers hyperparameters across width and depth based on each tensor’s functional fan-in and the model’s depth. 
\heterop{} produces a consistently tuned model family, enabling us to fit Chinchilla-style compute-optimal laws for activated model size, training-token count and image–video data ratio. 
Guided by these laws, we train an 11B-parameter \ours{} with 2B activated parameters.
Experiments yield three main findings. First, measured by pretraining diffusion loss, our dense backbone achieves \textbf{1.7$\bm{\times}$} the compute efficiency of a matched full-attention Wan-2.1 (2B) baseline, while the complete system raises this factor to \textbf{7.3$\bm{\times}$}. Second, without length-specific fine-tuning, \ours{} exhibits zero-shot length extrapolation from 5-second training clips to \textbf{30-second} videos, with only a $6.5\%$ degradation in FID over the final five seconds. Third, the fitted scaling laws indicate that compute-optimal image pretraining allocates compute nearly evenly between activated model size and training-token count, whereas video pretraining modestly favors model size at higher compute budgets. Together, we hope these findings provide a principled foundation for designing and scaling efficient long-context diffusion architectures.
}

\date{July 25th, 2026}

\begin{document}

\maketitle

\section{Introduction}
\label{sec:introduction}

\begin{figure}
    \centering
    \includegraphics[width=1\linewidth]{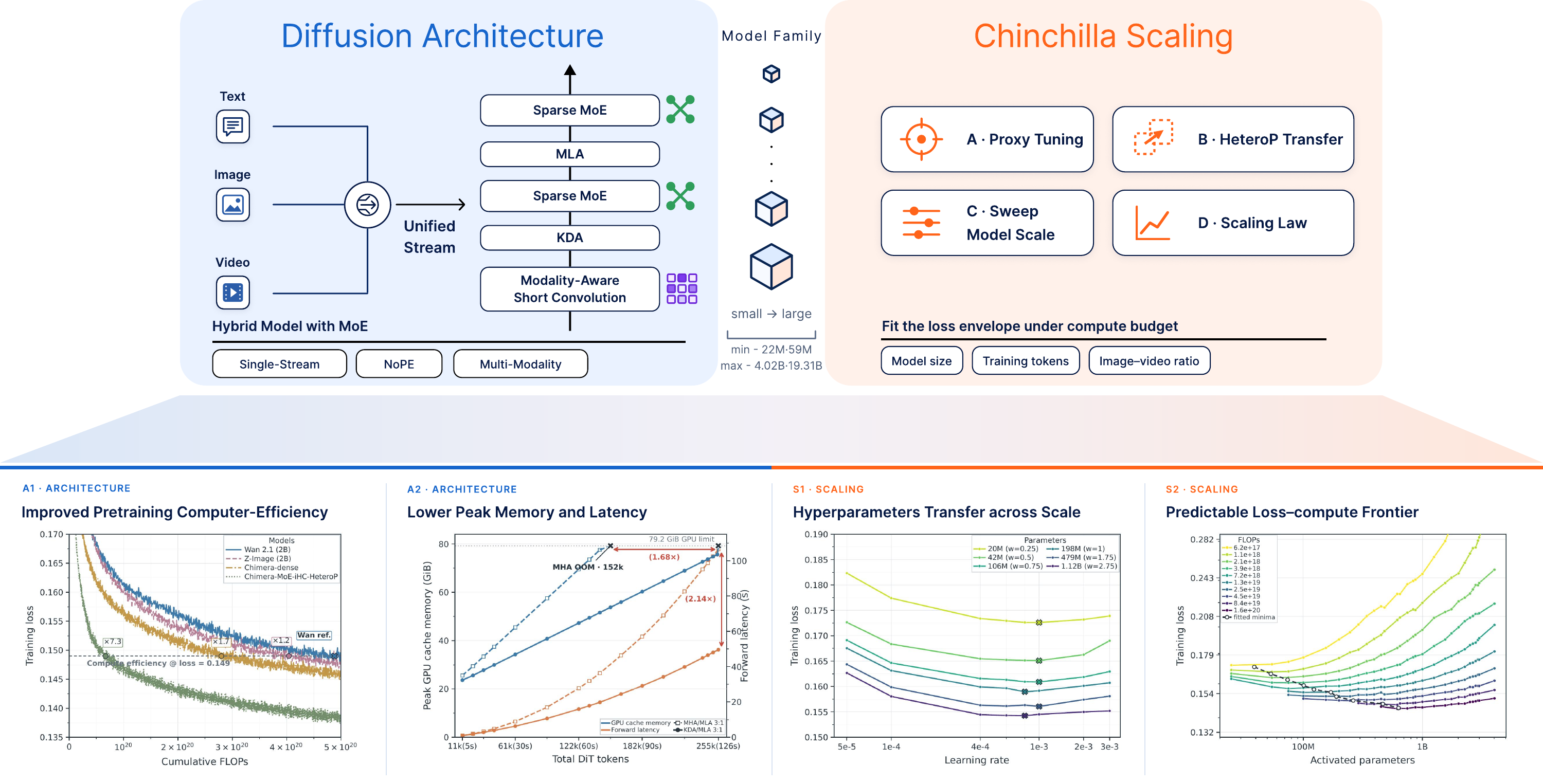}
    \vspace{-0.3in}
    \caption{\textbf{Overview of Chimera's architecture and scaling framework.}  \ours{} couples a hybrid-linear-global diffusion architecture (top left) with a Chinchilla-style scaling methodology (top right).
    The lower panels show both contributions: the architecture improves pre-training compute efficiency (A1) and long-sequence memory/latency efficiency (A2), while the scaling framework enables reliable cross-scale hyperparameter transfer (S1) and yields a predictable loss-compute frontier (S2).}
    \label{fig:teaser}
    \vspace{-1mm}
\end{figure}

Visual generation is becoming a long-sequence problem. A $4096\times4096$ image patchified at $16\times16$ already contains roughly 65K visual tokens, while videos, dense conditions, and multi-turn interactions increase the sequence length further~\citep{rombach2022ldm,peebles2023scalable,videoworldsimulators2024}. We refer to this as the token-extensive regime, in which sequence length becomes a primary determinant of memory, latency, and training cost. Yet progress in this regime is constrained by two coupled gaps. Architecturally, the dominant Diffusion Transformer still relies on full self-attention~\citep{peebles2023scalable}, whose pairwise interactions scale quadratically with sequence length~\citep{vaswani2017attention}. Methodologically, visual generation lacks a systematic framework for scaling such architectures: models and optimization recipes are often designed at a particular scale, without a controlled model family for establishing scaling laws that guide how diffusion-pretraining performance depends on model size, token budget, and data composition~\citep{liang2024scaling,zheng2026scaling}.

Language models entered this regime earlier and offer useful precedents on both fronts. Architecturally, language models have increasingly moved beyond dense attention~\citep{vaswani2017attention} toward sparse attention~\citep{child2019generating,yuan2025native,liu2025deepseekv3p2}, compressed key-value representations~\citep{shazeer2019mqa,ainslie2023gqa,liu2024deepseekv2}, recurrent or linear sequence operators~\citep{katharopoulos2020transformers,sun2023retentive,gu2023mamba,yang2024parallelizing,kimi2025linear}, and sparsely activated capacity~\citep{lepikhin2020gshard,fedus2022switch,dai2024deepseekmoe}. Methodologically, hyperparameter transfer across model scales~\citep{yang2022tensor,dey2026completep} and compute-optimal scaling laws that relate model capacity, data, and training compute~\citep{kaplan2020scaling,hoffmann2022chinchilla} have turned model scaling into a systematic and increasingly predictable process.

However, neither component of the language-model recipe can be transferred directly to visual diffusion training. On the architectural side, visual diffusion must preserve local structure over two- and three-dimensional grids, enable direct interaction among heterogeneous text, image, and video contexts, and remain effective under a denoising objective~\citep{ho2020denoising,ho2022video,ju2025editverse,cai2025zimage}. On the scaling side, hybrid visual backbones contain tensors whose functional input dimensions grow differently, or do not grow at all, as nominal model width increases, so standard rules based on a single global scaling ratio do not preserve comparable optimization dynamics~\citep{yang2022tensor,dey2026completep}. Moreover, visual training compute depends not only on the number of tokens, but also on data composition. The challenge is therefore twofold yet coupled: \textit{to design an efficient context-modeling architecture tailored to the visual data structure, and to develop a systematic framework for scaling it and deriving its scaling laws.}

We present \ours{}, a visual diffusion backbone co-designed with a principled scaling framework for token-extensive generation. As shown in Fig.~\ref{fig:teaser}, \ours{} represents text, images, and videos as a unified token sequence for modeling.
At the sequence scale, most layers use Kimi Delta Attention (KDA)~\citep{kimi2025linear} as a linear-complexity mechanism for content-adaptive long-range state tracking, while periodic Multi-head Latent Attention (MLA) layers~\citep{liu2024deepseekv2} provide direct global token-to-token interaction with compressed key-value representations. 
At the local scale, we propose modality-aware short convolutions aggregate neighbors along the native axes of each modality before every KDA state update. KDA therefore writes locally enriched features, rather than isolated tokens from the flattened sequence, into its recurrent state. 
The local relative-offset cues supplied by the convolutions, together with the ordering inherent in KDA's causal recurrence, make a simple temporal-major raster scan sufficient and allow us to dispense with explicit positional embeddings. 
Orthogonally, \ours{} combines sparse Mixture-of-Experts (MoE), identity hyper-connections (iHC), and sandwich normalization for greater capacity at controlled activated compute and stable signal propagation.

On the training side, to address the challenge of transferring hyperparameters across scales in our heterogeneous architecture~\citep{yang2022tensor,dey2026completep}, we introduce \heterop{}, a hyperparameter-transfer parameterization that derives a separate scaling ratio for each tensor from its functional fan-in and model depth. 
\heterop{} translates these tensor-wise ratios into module-specific transfer rules for learning rates, initialization, and residual scaling, allowing a recipe tuned on a small proxy model to transfer across model widths and depths. 
Together, these rules produce a consistently optimized and mutually comparable family of \ours{} models (59M to 19.3B total parameters) without per-scale retuning. This controlled model family enables fitting Chinchilla-style compute-optimal scaling laws~\citep{hoffmann2022chinchilla} over activated parameter count and visual training tokens. We further extend these laws by treating the image--video data mixture as an explicit scaling variable, capturing how the compute-optimal allocation depends on modality composition.

Across three independent estimators, the training-curve envelope, IsoFLOP profiles, and parametric loss-surface fitting, our scaling laws consistently prescribe nearly balanced growth of activated model size and training tokens for image pre-training(Fig.~\ref{fig:teaser}-S2), with $N_{\mathrm{opt}}\propto C^{0.48\text{--}0.52}$, while video pre-training shifts modestly toward model capacity at higher compute budgets, with $N_{\mathrm{opt}}\propto C^{0.53\text{--}0.56}$. 
Guided by the fitted scaling laws, we train a \ours{} model with 11B total parameters and 2B activated parameters given our training budget. 
Under matched training compute (Fig.~\ref{fig:teaser}-A1), it reaches the training loss of a full-attention Wan 2.1 (2B) baseline~\citep{wan2025wan} with $7.3\times$ fewer FLOPs. 
\ours{} is also competitive with strong image generators such as FLUX.1-dev and Z-Image-Turbo~\cite{cai2025zimage} on the GenEval and DPG-Bench benchmarks. 
Notably, our final model is trained only around 600 H100 Days, substantially less than prior method: Z-Image-Turbo, for example, reports a total training budget of approximately 12.4K H100 Days, 20 times of ours. 

Beyond guiding the final model allocation, \ours{}'s architecture brings concrete advantages in long-sequence generation. Benefiting from its position-embedding-free design, \ours{} generates $30$-second videos despite being trained only on $5$-second clips, requiring no length-specific fine-tuning and extending the generation horizon by $6\times$ while increasing frame-wise FID by only $6.5\%$, compared with an increase of more than $50\%$ for prior arts~\citep{wan2025wan}.
Moreover, compared with the matched MHA/MLA full-attention counterpart, our KDA/MLA backbone supports more than $1.68\times$ longer sequences within a single $80$\,GB GPU (NVIDIA A100-SXM4-80GB) and runs $2.14\times$ faster at $255$k tokens (Fig.~\ref{fig:teaser}-A2).
\footnote{We compare matched KDA/MLA and MHA/MLA $3{:}1$ backbones with approximately 2B activated parameters, using batch size one, BF16 activations, $512$ text tokens, and $18{\times}28$ visual tokens per temporal slice. For the roll-out-cache memory test, we allocate, touch, and retain the cache tensors on device during one complete denoiser forward. FlashAttention does not write the full $N{\times}N$ attention matrix to high-bandwidth memory: it streams blocks of $Q$, $K$, and $V$ through on-chip SRAM, computes $QK^\top$ tile by tile, and accumulates the exact output using online softmax. This removes the quadratic attention workspace, but not the quadratic arithmetic. The 24 MHA layers therefore have sequence-dependent BF16 key--value cache memory of $\mathcal{O}(NHd_h)=\mathcal{O}(ND)$, where $H$ and $d_h$ are the number and width of the attention heads and $D=Hd_h$. In contrast, the 24 KDA layers retain $H$ fixed-size FP32 recurrent states of shape $d_h{\times}d_h$, requiring $\mathcal{O}(Hd_h^2)$ memory independent of $N$; this is the $D{\times}D$ state-memory term when $D$ denotes the per-head state width. Both variants retain the same eight BF16 MLA latent key--value caches. We report peak allocated CUDA memory on one NVIDIA A100-SXM4-80GB and stop at the first out-of-memory input. For speed, we exclude cache construction, use FlashAttention for MHA, and report the median of three CUDA-synchronized post-warm-up forwards.}
Together, these results support jointly considering architecture and scaling for visual generation.

Our contributions are threefold.
First, we introduce \ours{}, a hybrid single-stream visual diffusion backbone for token-extensive generation. It represents text, image, and video as a unified sequence.
Our model combines KDA linear attention with periodic MLA-based global attention;
uses modality-aware causal short convolutions to support a single raster-order scan without positional embeddings; 
and adopts MoE layers, identity hyper-connections, and sandwich normalization for capacity and training stability.
Second, we develop, to our knowledge, 
the first comprehensive hyperparameter-transfer parameterization (\heterop{}) for heterogeneous visual diffusion
backbones.
\heterop{} allows optimization hyperparameters tuned on a small proxy model to transfer reliably across model widths and depths.
Third, we fit Chinchilla-style compute-optimal scaling laws for image and video pre-training over activated model size  and visual training tokens,  and extend the analysis to the image--video data composition.

\section{\ours{} Overview}
\label{sec:overview}

We aim to modernize diffusion models, particularly DiT~\citep{peebles2023scalable}, for the token-extensive visual generation regime.
This objective involves two tightly coupled sub-goals: designing architectures that balance efficiency and capacity, and understanding how their advantages evolve with model size, the number of training tokens, and training-data composition. Accordingly, we study architecture design and scaling jointly: the architecture defines a controlled model family, while the scaling analysis determines how model size, training tokens, and training-data composition should be allocated under a fixed compute budget. Section~\ref{sec:model_arch} details the architecture, while Sec.~\ref{sec:scaling} presents the scaling solution.

\paragraph{\textbf{Model Architecture.}} We summarize the key design choices of \ours{} below, ordered from the overall framework to individual modules:

\begin{itemize}
    \item \textbf{Single-stream Multimodal Backbone.} \ours{} concatenates multimodal tokens from text, images, and video into one sequence and processes them with a single shared backbone. Unlike cross-attention or dual-stream designs~\citep{wan2025wan, wu2025qwenimage}, this single-stream interface enables direct token-level interaction across modalities without modality-specific branches. It also simplifies scaling: a shared stack over one token sequence inherits language-model practices such as hyperparameter transfer and compute-optimal scaling laws.

    \item \textbf{Hybrid Linear and Global Attention.} \ours{} uses KDA~\citep{kimi2025linear} in most layers for $\mathcal{O}(N)$ long-sequence computation and periodically interleaves MLA layers that restore explicit global token interactions through compressed key-value states. Together, they balance long-context efficiency and global modeling capacity. Unlike prior works that rely on complex scanning patterns for visual modeling~\citep{liu2024vmamba,hu2024zigma}, \ours{} applies linear attention through a \textbf{single temporal-major raster scan} for simple and uniform multimodal processing.
    
    \item \textbf{Minimal Modality-aware Inductive Bias and No Positional Embedding (NoPE).} Attention-based models typically use positional embeddings to encode the token positions. For visual data, however, positional embeddings can hinder extrapolation beyond the training length, and factorizing them across spatial (image) and temporal axes (video) requires ad-hoc design choices. The state-tracking property of KDA offers a simple alternative: with fine-grained channel-wise state decay and recurrent delta updates, KDA behaves like local sliding-window attention with \emph{an adaptive window length and causal structure}, which removes the need for positional embeddings, as observed in recent language models~\citep{kimi2025linear}. To adapt this design to multimodal visual data, we retain a single minimal inductive bias: modality-aware short convolutions that serve as a ``Canon Layer'~\citep{allen2026physics} to capture local spatio-temporal dependencies among visual tokens. Order and locality thus come from the scan and the convolution, so the whole backbone processes multimodal tokens through one temporal-major raster scan without positional embeddings.
    
    \item \textbf{Sparse Capacity and Training Stability.} For capacity, \ours{} adopts sparse MoE activation~\citep{fedus2022switch}, which increases total parameters while controlling per-token computation. For stability, it pairs identity hyper-connections~\citep{xie2025mhc}, which stabilize the residual streams, with sandwich normalization~\citep{ding2021cogview}, which regulates the variance of sub-layer outputs.
\end{itemize}


\paragraph{\textbf{Pre-training and Chinchilla Scaling.}} We study how \ours{} scales during pre-training, the stage in which the model acquires its generative knowledge. We argue that the diffusion pre-training loss should be the reliable metric for measuring knowledge acquisition at this stage, while other vision-related evaluation metrics\footnote{FID, DPG~\citep{hu2024ella}, GenEval~\citep{ghosh2023geneval}, etc} are better treated as post-training metrics rather than pre-training objectives. This perspective allows us to treat visual pre-training analogously to language-model pre-training and fit compute-optimal scaling laws. We outline the key ingredients below.

\begin{itemize}
    \item \textbf{Diffusion Loss as a Pre-training Metric.} 
    Under a fixed diffusion objective and noise parameterization, diffusion loss provides a sampler-independent metric for comparing models across scales. Unlike downstream generation metrics such as FID, it does not depend on sampling algorithms, classifier-free guidance scales, or post-training choices. Moreover, diffusion can be interpreted as approximate autoregressive modeling in the frequency domain~\citep{dieleman2024spectralautoregression}: denoising from high to low noise levels predicts visual information from coarse to fine scales, which parallels next-token prediction under progressively longer contexts. This motivates its use for a more faithful pre-training metric than downstream generation scores.

    \item \textbf{Hyperparameter Transfer and Fair Scaling Comparison.} Hyperparameter transfer reduces the cost of hyperparameter search at the target scale and helps every model in a scaling study reach near-optimal performance. We build on $\mu$P-style
    transfer~\citep{yang2022tensor,dey2026completep} across model scales, but \ours{} poses an additional challenge: its heterogeneous modules do not share a single global scaling multiplier. We therefore propose \heterop{} to replace this global-ratio approximation with a module-specific ratio derived from each parameter group's functional fan-in, covering KDA, MLA, sparse MoE, identity hyper-connections, and diffusion conditioning. As a result, the data points used to fit scaling curves come from models trained with scale-appropriate and near-optimal hyperparameters, a factor often not explicitly controlled in prior scaling-law studies~\citep{hoffmann2022chinchilla,liang2024scaling,weng2026scaling}.
    
    \item \textbf{Fitting Chinchilla Scaling Laws.} Using this controlled model family, we fit Chinchilla-style laws over activated model parameters and training tokens to predict their compute-optimal allocation under a fixed compute budget. We also use IsoFLOP analysis to estimate how the optimal image-video mixture changes with compute.
\end{itemize}

\section{\ours{} Architecture} 
\label{sec:model_arch}

\subsection{Preliminaries and Training Objectives}
\label{sec:model_arch_preliminary}

\ours{} follows the rectified-flow framework with $v$-prediction and $v$-loss~\citep{lipman2022flow, liu2022flow}. 
Each image or video is encoded by a frozen VAE from~\citep{wan2025wan} with temporal, height, and width compression factors of $(4,8,8)$; a subsequent patchification layer converts the latents into the clean visual tokens $z_1$. The text prompt is encoded separately by a frozen T5-style encoder~\citep{raffel2020t5}, and we denote the resulting text-token sequence by $c$. During training, we sample Gaussian noise $z_0 \sim \mathcal{N}(0, I)$ and a timestep $\tau$, and linearly interpolate between $z_0$ and $z_1$ to obtain the noised visual tokens $z_\tau = \tau z_1 + (1 - \tau) z_0$. The velocity target is $v_\tau = \frac{d z_\tau}{d\tau} = z_1 - z_0$. We train the denoising model with
\begin{equation}
    \mathcal{L}
    =
    \mathbb{E}_{\tau,z_0,z_1}
    \left\|\algours{}(c,z_\tau,\tau;\theta)-v_\tau\right\|_2^2,
\end{equation}
where $\theta$ denotes the trainable parameters and the loss is calculated only at visual-token positions.

\begin{figure*}[h]
    \centering
    \includegraphics[width=\linewidth]{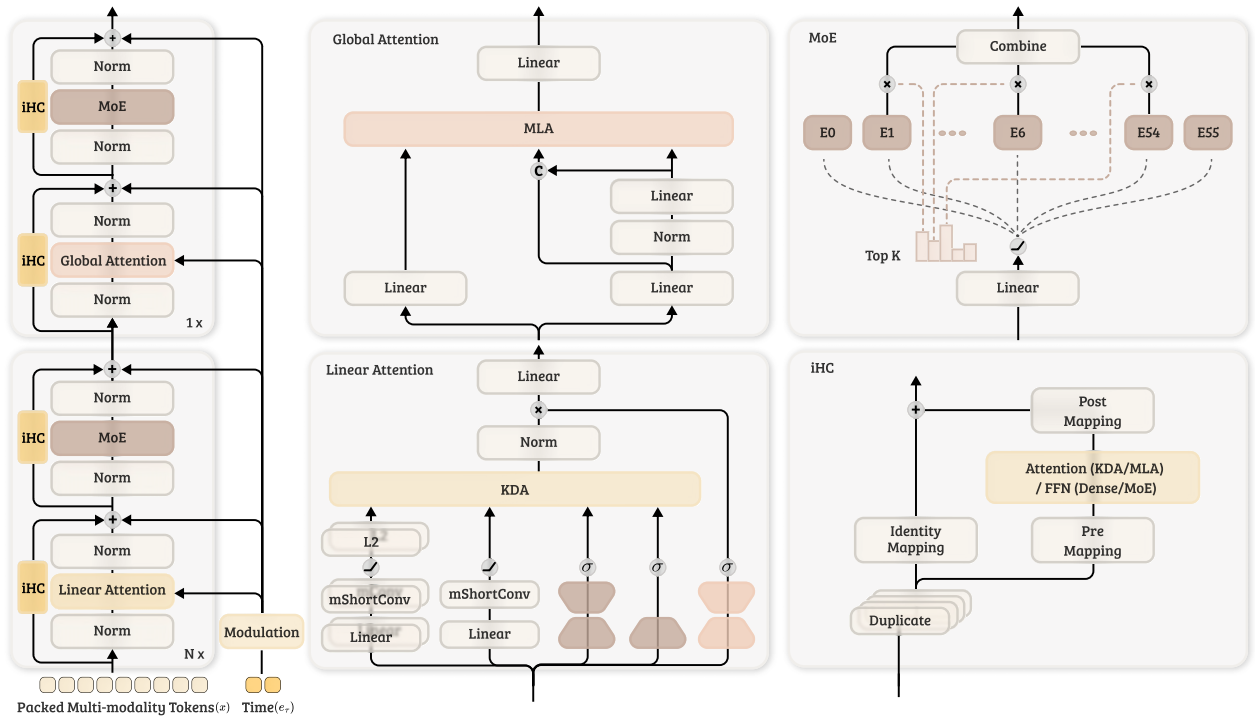}
    \caption{
    \textbf{Illustration of the \ours{} architecture.}
    \ours{} is a single-stream latent diffusion Transformer in which text tokens and noised visual tokens are concatenated into a unified sequence and processed jointly. Each block combines a sparse Mixture-of-Experts feed-forward network with either a KDA layer or a MLA layer. Before each KDA update, a modality-aware short convolution (mShortConv) mixes local context in each modality's native geometry. Timestep information is injected through AdaLN modulation. Each block applies sandwich-style RMSNorm (pre- and post-normalization), while identity hyper-connections (iHC) wrap each sublayer to stabilize signal propagation.
    }
    \label{fig:model_arch}
\end{figure*}

\subsection{Single-Stream Framework}
\label{sec:single_stream}

\ours{} adopts a single-stream framework in which text and visual tokens interact directly within one shared backbone. Its input sequence is
\begin{equation}
    x = [P_{\mathrm{text}}(c),P_{\mathrm{vis}}(z_\tau)],
\end{equation}
where $P_{\mathrm{text}}$ and $P_{\mathrm{vis}}$ are separate trainable projections into the shared hidden space, and padding tokens are removed from $c$ before concatenation. The visual tokens are flattened frame by frame in temporal-major raster order, traversing each frame in row-major order, as formalized in Eq.~\ref{eq:raster-order}. For the diffusion timestep $\tau$, we construct a sinusoidal embedding and project it to the model dimension, yielding the timestep embedding $e_\tau$, which conditions each block through the modulation pathway defined below.

\ours{} consists of a stack of Transformer-style blocks, each containing an attention sub-layer followed by a feed-forward sub-layer (FFN). The attention operator is either KDA or MLA, following the periodic schedule (Sec.~\ref{sec:attention}). The first and last blocks use dense SwiGLU FFNs~\citep{shazeer2020glu}, while all intermediate blocks use sparse MoE FFNs (Sec.~\ref{sec:moe}). Before
each KDA update, a modality-aware short convolution (mShortConv) injects local structure (Sec.~\ref{sec:short_conv}). Both sub-layers use sandwich normalization~\citep{ding2021cogview} and are wrapped with $M=4$ stream identity hyper-connections (iHC; Sec.~\ref{sec:ihc}). With a slight abuse of notation, we continue to write $x$ for this multi-stream state, whose structure is made explicit in Sec.~\ref{sec:ihc}. Timestep conditioning enters through adaptive layer normalization (AdaLN) modulation and residual gates. Omitting the layer index, one block computes:
\begin{align}
(\delta_A, \gamma_A, g^{A}, g^{F})
    &= \operatorname{MLP}_{\mathrm{mod}}(e_\tau),
    \label{eq:adln_mlp} \\
\tilde{x}_A
    &= \operatorname{iHC\text{-}Read}_A(x),
    \label{eq:ihc_read_a} \\
\hat{x}_A
    &= \operatorname{RMSNorm}^{pre}_A(\tilde{x}_A)
       \odot (1 + \gamma_A) + \delta_A,
    \label{eq:input_modulate} \\
a
    &= \operatorname{RMSNorm}^{post}_A
       \!\left(\operatorname{Attn}(\hat{x}_A)\right),
    \label{eq:attention} \\
x'
    &= \operatorname{iHC\text{-}Write}_A
       \!\left(x,\; g^{A} \odot a\right),
    \label{eq:ihc_attention} \\
\tilde{x}_F
    &= \operatorname{iHC\text{-}Read}_F(x'),
    \label{eq:ihc_read_f} \\
f
    &= \operatorname{RMSNorm}^{post}_F
       \!\left(
       \operatorname{FFN}
       (\operatorname{RMSNorm}^{pre}_F(\tilde{x}_F))
       \right),
    \label{eq:ffn} \\
x''
    &= \operatorname{iHC\text{-}Write}_F
       \!\left(x',\; g^{F} \odot f\right).
    \label{eq:ihc_ffn}
\end{align}
Here, $x$, $x'$, $x'' \in \mathbb{R}^{L \times M \times d}$ denote the multi-stream states, whereas $\tilde{x}_A$, $\hat{x}_A$, $a$, $\tilde{x}_F$, $f \in \mathbb{R}^{L \times d}$. The $\operatorname{iHC\text{-}Read}$ operators collapse the stream axis from $\mathbb{R}^{L \times M \times d}$ to $\mathbb{R}^{L \times d}$, whereas $\operatorname{iHC\text{-}Write}$ distributes each sub-layer update back to the $M$ streams. The timestep-conditioned vectors $e_\tau$, $\delta_A$, $\gamma_A$, $g^A$, $g^F \in \mathbb{R}^{d}$ are broadcast across the $L$ token positions. The state $x''$ becomes the input to the next block.
Equation~\ref{eq:adln_mlp} maps $e_\tau$ to an attention shift $\delta_A$, an attention scale $\gamma_A$, and residual gates $g^A$ and $g^F$ for the two sub-layers. Thus, the attention input receives timestep-conditioned AdaLN modulation~\citep{peebles2023scalable}, while both residual updates are timestep-gated (Eqs.~\ref{eq:ihc_attention} and~\ref{eq:ihc_ffn}). 
With $M=4$ parallel residual streams, both sub-layers share the same \emph{read}--compute--\emph{write} pattern over the streams. Each sub-layer starts with $\operatorname{iHC\text{-}Read}$ (Eqs.~\ref{eq:ihc_read_a} and~\ref{eq:ihc_read_f}), which aggregates the $M$ streams into one sub-layer input.  For attention, this input is pre-normalized and timestep-modulated, passed to the attention operator(Eq.~\ref{eq:input_modulate} and ~\ref{eq:attention}). The resulting update is then gated by $g^{A}$, and written back to all $M$ streams by $\operatorname{iHC\text{-}Write}$ (Eq.~\ref{eq:ihc_attention}). The FFN sub-layer also repeats this pattern (Eqs.~\ref{eq:ihc_read_f}--\ref{eq:ihc_ffn}). Within each sub-layer,  Sandwich Normalization places RMSNorm both before ($\texttt{RMSNorm}^{pre}$) and after ($\texttt{RMSNorm}^{post}$) the operator to regulate the variance for training stability. We define the attention operator $\texttt{Attn}(\cdot)$ in Sec.~\ref{sec:attention}, the feed-forward operator $\texttt{FFN}(\cdot)$ in Sec.~\ref{sec:moe}, and the iHC read/write operators in Sec.~\ref{sec:ihc}.  

After the final block, the $M$ residual streams are averaged into one representation (Sec.~\ref{sec:ihc}), and only visual-token positions enter the output head for rectified-flow velocity prediction. Text tokens condition these predictions through token-level interactions in the shared backbone, while the diffusion loss applies only to visual tokens.

\subsection{Hybrid Attention Mechanism with NoPE} 
\label{sec:attention}

\ours{} periodically interleaves Kimi Delta Attention (KDA)~\citep{kimi2025linear} and Multi-head Latent Attention (MLA)~\citep{liu2024deepseekv2}, at a $3{:}1$ KDA-to-MLA ratio. For a sequence of length $L$, each KDA layer requires only $\mathcal{O}(N)$ sequence complexity because it carries a constant-size recurrent state rather than materializing a full attention map. Each MLA layer retains full bidirectional self-attention with compressed key-value representations. The operators are therefore complementary: KDA provides efficient long-sequence computation, while MLA periodically enables unrestricted bidirectional global interaction. We detail both formulations and the no-positional-embedding (NoPE) design below.

\begin{equation}
    \operatorname{Attn}(\hat{x}_A) =
    \begin{cases}
        \operatorname{KDA}(\hat{x}_A), & \text{Linear Attention}, \\
        \operatorname{MLA}(\hat{x}_A), & \text{Global Attention}.
    \end{cases}
    \label{eq:hybrid_attn}
\end{equation}

\paragraph{\textbf{KDA with a Single Scan Pattern.}}
We traverse video latents frame by frame, using row-major order within each frame. For a latent grid of temporal length $T$, height $H$, and width $W$, this temporal-major raster order is
\begin{equation}
    z^{i,j,k} \mapsto z^m,
    \qquad
    m=i\,HW+j\,W+k,
\label{eq:raster-order}
\end{equation} 
where $i$, $j$, and $k$ index the temporal, height, and width dimensions, respectively. An image is treated as a single-frame latent grid. We omit the timestep subscript $\tau$ below for clarity.

After text tokens are prepended to the flattened visual tokens, every KDA head processes $\hat{x}_A$ using the same scan order.  The recurrent state update is causal in this order: at each visual position, the state summarizes the text prefix and previously scanned visual tokens. The update follows the KDA formulation of Kimi Linear~\citep{kimi2025linear}, with fine-grained channel-wise state decay rather than a single head-wise scalar. For a single head, omitting its index, we have
\begin{align}
    (Q,K,V)
        &= \operatorname{mShortConv}\!\left(W_{\mathrm{qkv}}(\hat{x}_A)\right), \\
    S_t
        &= \left(I-\beta_t k_tk_t^\top\right)\operatorname{Diag}(\alpha_t)S_{t-1}
           +\beta_t k_tv_t^\top,
           \label{eq:kda_state_update} \\
    [\operatorname{KDA}(\hat{x}_A)]_t
        &= W_o\!\left(
           \operatorname{Sigmoid}(W_{\mathrm{gate}}(\hat{x}_{A,t}))
           \odot \operatorname{RMSNorm}(S_t^\top q_t)
           \right),
\end{align}
where $q_t \in \mathbb{R}^{d_q}$, $k_t \in \mathbb{R}^{d_k}$, and $v_t \in \mathbb{R}^{d_v}$ denote the query, key, and value vectors at scan position $t$ in $Q$, $K$, and $V$, respectively, while $S_t \in \mathbb{R}^{d_k \times d_v}$ is the recurrent state after processing that position. The forget gate $\alpha_t \in (0,1)^{d_k}$ applies an independent decay to each key channel, while the head-wise scalar $\beta_t \in (0,1)$ controls both the erase and write terms.
The discrete scan index ranges over $t=1,\ldots,L$ and is distinct from the diffusion timestep $\tau$.
The recurrent state is reset to $S_0=0$ at each packed-sample boundary.
$\operatorname{mShortConv}(\cdot)$ denotes the proposed modality-aware short convolution (Sec.~\ref{sec:short_conv}), which supplies local structure and facilitates context-level information shortcut~\citep{allen2026physics}, while allowing the recurrent KDA update to retain a single scan order across modalities.

\paragraph{\textbf{MLA for Global Context Modeling.}} 
KDA compresses the scanned history into a constant-size recurrent state. This state is efficient to maintain, but its bounded size limits how much information can be recalled exactly from long visual sequences. We therefore interleave a small fraction of  full-attention layers to restore direct content-based interactions across all tokens. This hybrid design is more than a cost-capacity compromise. Recent analysis shows that, under standard complexity assumptions, alternating linear-attention and full-attention layers is strictly more expressive than either type alone on formal sequence task: linear attention contributes state tracking and full attention contributes recall, so their composition solves problems that neither solves by itself~\citep{merrill2026olmo}. To keep these global layers memory-friendly, we instantiate them with MLA, which retains full bidirectional attention while compressing the key-value representations:
\begin{align}
    (Q,U,K^{\mathrm{direct}})
        &= \left(
           W_q(\hat{x}_A),
           W_{\mathrm{kv}}^{\mathrm{down}}(\hat{x}_A),
           W_k^{\mathrm{direct}}(\hat{x}_A)
           \right),
           \label{eq:mla_q} \\
    [K^{\mathrm{lat}},V]
        &= W_{\mathrm{kv}}^{\mathrm{up}}
           \!\left(\operatorname{RMSNorm}(U)\right), \\
    K_i
        &= [K^{\mathrm{direct}},K_i^{\mathrm{lat}}],
           \qquad i=1,\ldots,n_h,
           \label{eq:mla_kv} \\
    \operatorname{MLA}(\hat{x}_A)
        &= W_o\!\left(\operatorname{MHSA}(Q,K,V)\right).
           \label{eq:mla_out}
\end{align}
Here, $n_h$ denotes the number of attention heads. $U$ is the
compressed key--value latent. Both $K^{\mathrm{lat}}$ and $V$ are decoded from
$U$, whereas $K^{\mathrm{direct}}$ is projected directly from $\hat{x}_A$ and
shared across heads. This reduces the key--value cache cost while
retaining full bidirectional attention.
Unlike standard MLA~\citep{liu2024deepseekv2}, we leave $K^{\mathrm{direct}}$ unrotated (NoPE).

\paragraph{\textbf{NoPE v.s. RoPE.}} \ours{} uses no explicit positional encoding (NoPE) in either KDA or MLA layers. To motivate this choice, we first examine what positional encoding contributes to attention. Absent explicit positional modulation, attention is a content-matching operation. That's to say, its logits are inner products between learned query and key features, without direct dependence on token indices. Positional encodings add position-dependent inductive biases to this content pathway. For RoPE~\citep{su2024roformer}, the \textit{de facto} choice in both language and visual generation models, these biases take three forms:
\begin{itemize}
    \item \textbf{Position selection.} Some operations must activate at a specific relative token offset. A canonical example is the previous-token head, which serves as a building block of induction heads~\citep{elhage2021mathematical, olsson2022context}. RoPE enables such \emph{token-index-aware/token-level-shortcut} operations by rotating each query-key channel pair at a fixed frequency, allowing trained heads to exploit high-frequency pairs for precise offset selection. By aligning the phases of its highest-frequency pairs, an attention head can make its logit peak at a chosen small offset~\citep{barbero2024round}.
    
    \item \textbf{Recency decay.} Nearby context should generally receive greater weight than distant context. Positional encoding schemes introduce this bias either explicitly, as with ALiBi's distance-dependent penalty~\citep{press2022train}, or implicitly, as with RoPE. Under the conventional frequency spectrum and simplifying assumptions, RoPE's rotations can induce a distance-decaying envelope on the attention logits. This \emph{long-term decay} was a key motivation for RoPE~\citep{su2024roformer}.
    
    \item \textbf{Multidimensional encoding.} Multimodal data has inherently multidimensional positions: text tokens have a one-dimensional index, whereas video tokens use temporal, height, and width coordinates. Axial RoPE assigns rotary query-key pairs to these axes and rotates each subset by its coordinate~\citep{wang2024qwen2, yang2025cogvideox, wan2025wan}.
\end{itemize}

All three biases are implemented by rotating content-bearing query-key features, forcing positional and semantic information to share query-key capacity. Trained models mitigate this competition by separating the two across the frequency spectrum~\citep{barbero2024round}. Low-frequency pairs remain nearly rotation-invariant over the relevant context range, so their dot products are governed primarily by token content and preserve distance-robust semantic matching.  Short-range position selection, by contrast, are concentrated in the high-frequency pairs.

\begin{figure*}[h]
    \centering
    \begin{subfigure}[b]{0.483\linewidth}
        \centering
        \includegraphics[width=\linewidth]{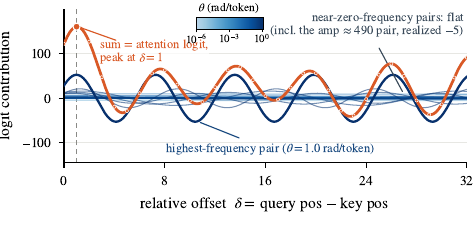}
        \vspace{-0.3in}
        \caption{}
        \label{fig:rope_qwen_decomp}
    \end{subfigure}
    \hfill
    \begin{subfigure}[b]{0.273\linewidth}
        \centering
        \includegraphics[width=\linewidth]{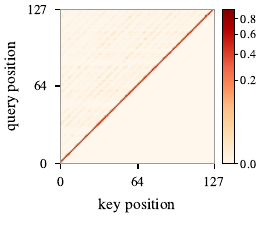}
        \vspace{-0.3in}
        \caption{}
        \label{fig:rope_qwen_attn}
    \end{subfigure}
    \hfill
    \begin{subfigure}[b]{0.228\linewidth}
        \centering
        \includegraphics[width=\linewidth]{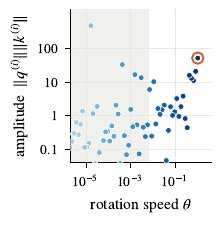}
        \vspace{-0.3in}
        \caption{}
        \label{fig:rope_qwen_amp}
    \end{subfigure}
    \vspace{-0.1in}
    \caption{
    \textbf{RoPE analysis of the strongest previous-token head in Qwen3-4B (layer 0, head 1).}
    \textbf{(a)}~Per-pair logit contributions (darker blue indicates faster rotation) and their sum (orange), which peaks exactly at $\delta{=}1$.
    \textbf{(b)}~Attention map over the first $128$ tokens (power-transformed color scale).
    \textbf{(c)}~Pair amplitude versus rotation frequency: pairs in the shaded region rotate less than one full turn within the context window, and the circled highest-frequency pair creates the peak in (a).
    }
    \label{fig:rope_qwen}
\end{figure*}

To characterize this frequency-wise division of labor, we first analyze a pretrained RoPE-based language model, Qwen3-4B~\citep{yang2025qwen3}, whose previous-token heads provide a clear diagnostic of short-range position selection. We use a $1{,}024$-token window of natural text. Because RoPE rotates each channel pair independently, every attention logit decomposes into a sum of per-frequency contributions, allowing the role of each frequency to be measured directly. Specifically, we average the queries and keys over the window and, for each rotary channel pair $i$, rotate the query component onto the $+x$ axis while absorbing both vector norms into the pair amplitude $|q^{(i)}||k^{(i)}|$. The resulting pairwise cosine terms sum exactly to the attention logit, agreeing with the directly computed value to within $3{\times}10^{-13}$ (Fig.~\ref{fig:rope_qwen_decomp}). We first examine the strongest previous-token head (i.e., layer 0, head 1), for which $98.5\%$ of queries attend most strongly to the immediately preceding token (Fig.~\ref{fig:rope_qwen_attn}). In this head, the highest-frequency pair, which rotates by approximately one radian per token, produces a logit that peaks exactly at offset one: its learned phase places the cosine maximum at $\delta\approx1$. By contrast, the largest-amplitude pairs all lie near zero frequency and contribute almost uniformly across offsets (Fig.~\ref{fig:rope_qwen_amp}). These slowly rotating pairs are therefore modulated primarily by content rather than position. The largest such pair rotates by less than $1^{\circ}$ across the entire window, yet contributes little in this example because its averaged query and key vectors are nearly orthogonal. We then search across all heads for those that attend exactly $n$ tokens into the past, quantified by the fraction of queries whose attention peaks at that offset. Such heads appear only at short distances: the best head attains a fraction of $0.56$ at $n{=}2$, but only $0.07$ at $n{=}10$. These results indicate that RoPE-based position selection operates almost entirely over short ranges.

We then apply the same analysis to visual generation models, FLUX.2~\citep{blackforestlabs2025flux2} and Wan2.2~\citep{wan2025wan}, and observe the same division of labor. That's to say, slowly rotating channel pairs preserve distance-robust semantic matching, whereas rapidly rotating pairs enable short-range position selection (protocol and full results in Appendix~\ref{app:rope_probe}).

From this perspective, RoPE implements each of its three inductive bias  through rotary query-key channels, while introducing several limitations.
(1) First, position selection consumes substantial attention capacity. In the previous-token head above, a single fast-rotating pair fixes the target offset, while the large-amplitude slow pairs contribute only a flat, offset-independent background. As a result, an entire attention head is effectively dedicated to a fixed $n$-token shift, yet reliable selection remains limited to very short ranges. 
(2) Second, recency decay is only implicit. The decaying envelope arises as an average property of the summed rotations rather than from an explicit content-adaptive decay mechanism, and can be bypassed by trained heads~\citep{barbero2024round}.
(3) Third, layout encoding is hand-designed. The channel partition must be specified manually, each additional modality or positional axis further divides the available channels and reduces the frequency resolution per axis, and the resulting allocation remains fixed at design time.

\ours{} instead provides each inductive bias through a dedicated mechanism, none of which consumes attention-channel capacity. Token order is supplied by the recurrent scan itself: KDA processes the sequence in temporal-major raster order, making its state updates inherently order-aware. 
(1) \emph{Position selection} is handled by the modality-aware short convolution (Sec.\ref{sec:short_conv}). A depthwise kernel mixes tokens at explicit index offsets, implementing index-aware operations with only a small number of parameters rather than dedicating an entire attention head to them. Related convolutional mechanisms have also been used to inject positional information into vision Transformers~\citep{chu2023conditional}. Because the analyses above show that RoPE-based position selection is primarily short-range, a short kernel is sufficient to replace it.
(2) \emph{Recency decay} is intrinsic to KDA. The channel-wise forget gate $\alpha_t$ (Eq.~\ref{eq:kda_state_update}) decays the recurrent state at each scan step, linking each token to its predecessors over a learned, content-adaptive window rather than through a fixed spectral envelope.
(3) \emph{Multidimensional encoding} is determined by the structure of the convolution. Text tokens are mixed with a causal 1D kernel along the token index, whereas visual tokens are mixed with a 3D kernel over the temporal-height-width grid. Each modality's native layout is therefore encoded directly by the operator, without partitioning the channels across positional axes.
With these three positional biases assigned to dedicated modules, attention is free to perform its native function: content matching. MLA without RoPE corresponds to the limiting case in which every channel has zero rotary frequency, so its logits depend only on content~\citep{barbero2024round}. The queries and keys in KDA likewise contain no positional phases. This design parallels recent hybrid language models that combine NoPE full attention with gated linear attention~\citep{kimi2025linear,kimiteam2026kimik3openfrontier}. \ours{} applies it to multimodal data via modality-aware convolution.

Beyond this disentanglement, removing RoPE also avoids its dependence on training length. At relative offsets beyond those observed during training, nonzero rotary frequencies can produce phase configurations outside the training distribution.  When a model trained on 5-second clips is used to generate 10- or 20-second videos, these unseen phase configurations can degrade~\citep{kazemnejad2023impact, barbero2024round}. By contrast, our three mechanisms introduce no explicit positional phase tied to the training range: the convolution kernels have fixed support, and the same recurrent update is applied at each scan step. This, to some extent, removes one architectural obstacle to zero-shot extrapolation to longer sequences.

Overall, \ours{} factors RoPE's three positional functions into explicit, specialized modules: a short convolution for position selection, gated state decay for recency, and operator structure for layout encoding. Each inductive bias is provided in a direct and learnable form, without consuming attention capacity for simple positional operations, manually partitioning channels, or coupling the model to its training length.

\subsection{Modality-aware Short Convolution} \label{sec:short_conv}

\begin{figure}[ht]
    \centering
    \includegraphics[width=\linewidth]{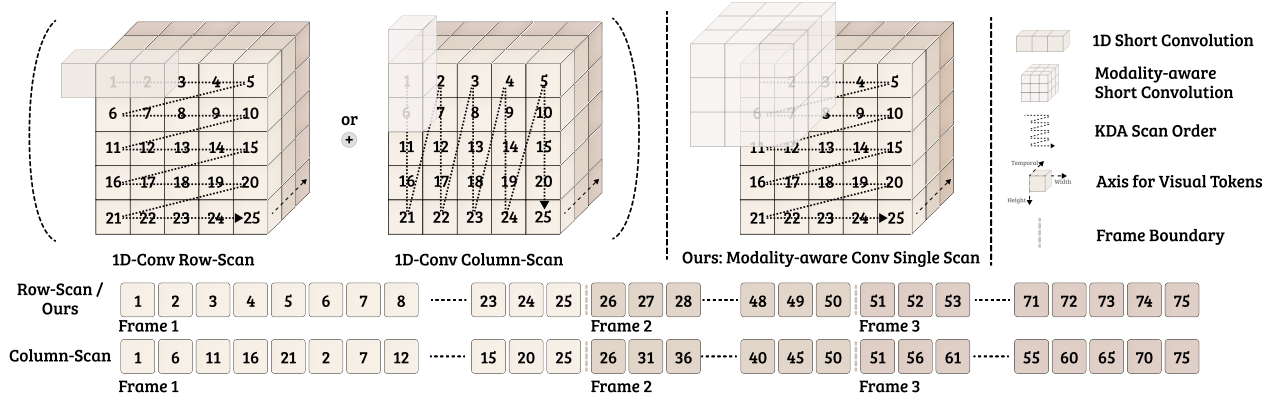}
    \caption{
    \textbf{Conceptual comparison of local-mixing designs in vision modality.}  Conventional 1D short convolution is paired with multi-axis scanning for vision modeling, whereas our modality-aware short convolution models spatial-temporal locality under a single scan pattern
    }
    \label{fig:mconv_arch}
\end{figure}

A short convolution injects local structure into the token features before each linear-attention update, following common practice in the linear-attention literature~\citep{kimi2025linear}. A plain one-dimensional convolution, however, mixes only tokens that are adjacent in the scan order and thus models locality only along a fixed-order axis. To recover the two- and three-dimensional locality of visual data, prior vision models pair such a convolution with multiple hand-designed scan directions~\citep{liu2024vmamba,hu2024zigma,hong2025pushingboundariesstatespace}. We instead make the convolution \emph{modality-aware}, so that it matches each modality's native structure and captures local spatial-temporal correlations directly. This allows \ours{} to retain a single scan pattern. Fig.~\ref{fig:mconv_arch} contrasts our modality-aware convolution this single scan pattern with the conventional pairing of a 1D convolution and multi-axis scanning.

Concretely, the modality-aware convolution uses modality-specific views of text and visual tokens while preserving their positions in the packed single-stream interface (Fig.~\ref{fig:mconv_arch}). Text tokens are gathered from their packed positions and processed with a causal one-dimensional depthwise convolution over the text sequence. Visual tokens are reshaped into their spatial-temporal grid and processed with a depthwise 3D convolution, which models local correlations along the temporal, height, and width axes at once. An image is handled as the special case of a video with temporal length one. The convolutions are applied independently within each modality segment, so no kernel crosses a text-visual or packed-sample boundary. After convolution, the text and visual outputs are scattered back to their original positions in the packed sequence. Because the convolution itself captures multi-dimensional locality, KDA can keep a single temporal-major scan instead of the multi-directional scans discussed above.

Implemented naively, this \emph{gather--convolve--scatter} procedure launches multiple kernels and materializes intermediate buffers per layer. We instead implement a fused Triton kernel that realizes the same text and visual convolutions directly over packed storage, avoiding explicit gather and scatter operations. In operator-level microbenchmarks against the unfused implementation on packed image ($512^2$ and $1024^2$) and video ($21$-frame) batches, the fused kernel achieves a $2.2$--$2.3\times$ forward speedup and a $1.5$--$1.8\times$ forward--backward speedup, while reducing peak activation memory during the forward pass by up to $4\times$.

Along the temporal dimension, the 3D convolution uses causal padding, so that each visual token mixes only with tokens from the current and past frames. We make the short convolution \emph{temporally causal} to keep its local mixing consistent with KDA's causal temporal-major scan, for the following reasons:
\begin{itemize}
    \item \textbf{Consistency with State Tracking.} The KDA operator updates a recurrent state while scanning the sequence, which aligns naturally with the temporal progression of video generation. A non-causal temporal convolution would introduce future-frame information into the local query, key, and value features before each KDA update, weakening this state-tracking interpretation. Causal padding keeps the local visual operator consistent with the causal dynamics of the KDA scan.
    
    \item \textbf{Compatibility with Sequential Generation.} A temporally causal convolution also facilitates adapting the pretrained backbone to downstream streaming generation tasks. For example, if the model is later converted into an autoregressive world model, the KDA pathway, including its short convolution, is already causal in time from pretraining, rather than relying on a causal constraint introduced only during post-training~\citep{huang2026self,hong2025relic}.
    
    \item \textbf{Performance.} Empirically, causal temporal padding in the short convolution has little effect on generation quality compared with the non-causal alternative, so this design choice comes at negligible cost.
\end{itemize}

In summary, KDA follows a causal temporal-major scan, while the short convolution is causal along the temporal axis. This alignment preserves KDA's state-tracking interpretation, while the convolution provides modality-specific local structure under a single scan pattern.

\subsection{Mixture-of-Experts (MoE)} \label{sec:moe}

To enhance the capacity of \ours{} without proportionally increasing activated computation per token, we replace the dense FFN with sparse MoE FFNs~\citep{fedus2022switch,dai2024deepseekmoe} in all except the first and last blocks. This choice complements the hybrid KDA/MLA design: once the attention cost is reduced, FFN computation accounts for a larger share of per-token computation, so enlarging dense FFNs would erode the efficiency gains from hybrid attention. Sparse activation instead increases the total FFN parameters, and hence the model capacity, while keeping the activated computation per token matched to a dense baseline.

Concretely, an MoE FFN contains $N$ routed experts $\{E_i\}_{i=1}^{N}$ and a router $r(\cdot)$. For each token input $u \in \mathbb{R}^{d}$, the router produces expert scores, selects a small top-$K$ set $\mathcal{T}(u)$, and combines only those expert outputs,
\begin{align}
    \mathcal{T}(u)
        &= \operatorname{TopK}(r(u) + b, K),
        \label{eq:moe_topk} \\
    \operatorname{MoE}(u)
        &= \sum_{i \in \mathcal{T}(u)} g_i(u) E_i(u),
        \label{eq:moe_output}
\end{align}
where $K$ is the number of experts activated per token, $b \in \mathbb{R}^{N}$ is a non-gradient balancing bias applied only during expert selection, and $g_i(u)$ is the routing weight of expert $i$. In intermediate blocks, $\operatorname{MoE}(\cdot)$ instantiates the $\operatorname{FFN}(\cdot)$ operator in Eq.~\ref{eq:ffn}. We use \textit{token-choice} routing: each token selects its activated experts independently. Following the fine-grained expert design of DeepSeekMoE~\citep{dai2024deepseekmoe}, each routed expert is a small SwiGLU FFN whose hidden dimension is $1/K$ of the dense FFN hidden width. In our main configuration, every MoE layer has $N=56$ routed experts and activates $K=8$ experts per token, corresponding to an activation ratio of $8/56=1/7$. The total expert parameters are thus enlarged by roughly $56/8=7\times$, while the $8$ activated experts together recover exactly the dense FFN hidden width, \textit{e.g.}, $8\times1024=8192$ in our 2B configuration. We use no shared experts, as we empirically find that the routed experts alone provide sufficient capacity in this visual diffusion setting. The MoE configuration, including expert count and activation ratio, is fixed across experiments, keeping our scaling-study runs directly comparable.

\begin{wrapfigure}[20]{r}{0.4\textwidth}
    \centering
    \vspace{-1em}
    \includegraphics[width=0.38\textwidth]{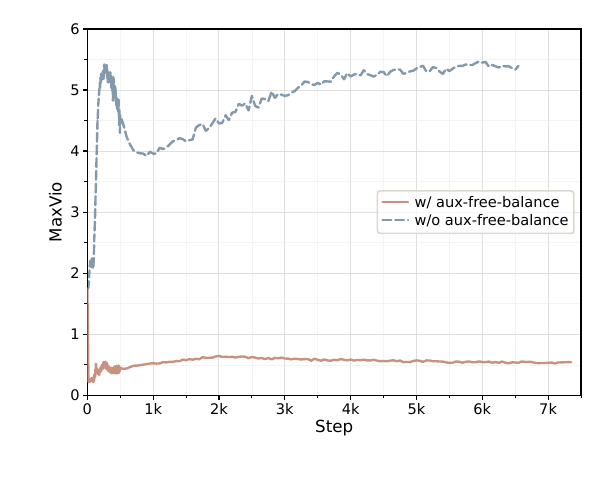}
    \vspace{-2em}
    \caption{\textbf{Expert load balance during training.} We plot the batch-level $\operatorname{MaxVio}$ (Eq.~\ref{eq:maxvio}), averaged over all MoE layers. With auxiliary-loss-free balancing, $\operatorname{MaxVio}$ stays around $0.5$ throughout training; without it, $\operatorname{MaxVio}$ rises above $5$, approaching the collapse bound of $N/K-1=6$.
    }
    \label{fig:moe_balance}
    \vspace{-0.4em}
\end{wrapfigure}
The router is a lightweight two-layer MLP producing sigmoid scores. After top-$K$ selection, the selected scores are normalized and scaled by $\sqrt{K}$ to form the weights $g_i(u)$. To avoid routing collapse and expert-load imbalance during training, we adopt the auxiliary-loss-free balancing strategy~\citep{wang2024auxiliary,liu2024deepseekv3}: the non-gradient bias $b$ enters only the top-$K$ selection, while the weights $g_i(u)$ are always computed from the unbiased router scores. After each training step, the bias of overloaded experts is decreased and that of underloaded experts is increased by a small fixed step, according to the observed expert loads. This keeps the expert load balanced without adding any auxiliary loss to the diffusion objective.

To verify that this strategy indeed balances the load in our visual diffusion setting, we follow \citep{wang2024auxiliary} and quantify the degree of load balance of an MoE layer by its \emph{maximal violation} (MaxVio):
\begin{equation}
    \operatorname{MaxVio} = \frac{\max_{1 \le i \le N} \operatorname{Load}_i - \overline{\operatorname{Load}}}{\overline{\operatorname{Load}}},
    \label{eq:maxvio}
\end{equation}
\WFclear
where $\operatorname{Load}_i$ is the number of tokens routed to expert $E_i$ in a training batch, and $\overline{\operatorname{Load}}$ is the mean load over the $N$ experts, \textit{i.e.}, the per-expert load under perfect balance. MaxVio equals $0$ under a perfectly uniform load and reaches $N/K-1=6$ in our configuration when all tokens select the same $K$ experts. Fig.~\ref{fig:moe_balance} tracks the batch-level MaxVio, averaged over all MoE layers, for two identical runs. With the balancing bias enabled, MaxVio drops below $1$ almost immediately and stabilizes around $0.5$ for the rest of training, \textit{i.e.}, even the most loaded expert receives only about $1.5\times$ its expected share of tokens. Disabling the bias instead leads to severe imbalance: MaxVio rises above $5$ within the first few hundred steps and keeps drifting toward the collapse bound, with the most loaded expert attracting over $6\times$ its expected load. These results confirm that the bias-based mechanism maintains well-balanced expert utilization throughout training without adding a load-balancing term to the diffusion objective.

\subsection{Identity Hyper-Connections (iHC)}
\label{sec:ihc}
Deep networks rely on residual connections to preserve signal propagation across depth. Hyper-Connections (HC)~\citep{zhu2025hyper} generalize this design by expanding the residual state into multiple streams and learning how each sub-layer reads from and writes back to them. mHC~\citep{xie2025mhc} further constrains the residual-stream transition to be doubly stochastic, preserving the approximate identity mapping required for stable large-scale optimization. This multi-stream interface is particularly well suited to \ours{}. KDA, MLA, modality-aware short convolution, and MoE perform complementary computations and produce updates with different statistics, yet a standard residual block forces every sub-layer to read from and write to the same state. As these heterogeneous updates accumulate with depth, a single stream can restrict how operator-specific information is preserved and recombined. Multiple residual streams provide token-dependent read/write pathways that increase routing capacity and maintain stable feature statistics through an identity path.

We adopt Identity Hyper-Connection (iHC), a simplified variant of mHC that fixes the residual-stream transition to the identity while retaining token-dependent read and write mappings. At the input of the denoising backbone, iHC expands the standard residual state into $M$ residual streams $R \in \mathbb{R}^{M \times d}$ by replication, where $d$ is the hidden dimension and $R$ is the per-token multi-stream state. Stacking these states across $L$ token positions yields the sequence-level state $x \in \mathbb{R}^{L \times M \times d}$ used in Eqs.~\ref{eq:ihc_read_a}--\ref{eq:ihc_ffn}. After the final block, the $M$ streams are averaged back into a single stream. Within each block, before each sub-layer, a read vector $h_{\mathrm{pre}}(R) \in \mathbb{R}^{M}$ aggregates the streams into the sub-layer input,
\begin{equation}
    \tilde{x} = \sum_{m=1}^{M} h_{\mathrm{pre},m}(R) R_m,
\end{equation}
which instantiates the $\operatorname{iHC\text{-}Read}$ operator in Eqs.~\ref{eq:ihc_read_a} and~\ref{eq:ihc_read_f}.
Let $\Delta \in \mathbb{R}^{d}$ denote the gated sub-layer
update, with $\Delta=g^{A}\odot a$ for attention
(Eq.~\ref{eq:ihc_attention}) and $\Delta=g^{F}\odot f$ for the FFN
(Eq.~\ref{eq:ihc_ffn}).
A write vector $h_{\mathrm{post}}(R) \in \mathbb{R}^{M}$ independently scales and adds the shared update to each residual stream,
\begin{equation}
    R'_m = R_m + h_{\mathrm{post},m}(R)\,\Delta,
    \qquad m=1,\ldots,M,
    \label{eq:ihc_write_rule}
\end{equation}
which instantiates the $\operatorname{iHC\text{-}Write}$ operator. The token-dependent read and write vectors are generated by
\begin{equation}
    \begin{bmatrix}
        h_{\mathrm{pre}}(R) \\
        h_{\mathrm{post}}(R)
    \end{bmatrix}
    =
    \begin{bmatrix}
        \mathbf{1}_M \\
        2\mathbf{1}_M
    \end{bmatrix}
    \odot
    \sigma\!\left(
        \alpha_h W_h \operatorname{RMSNorm}(\operatorname{vec}(R))
        +
        \begin{bmatrix}
            b_{\mathrm{pre}} \\
            b_{\mathrm{post}}
        \end{bmatrix}
    \right).
    \label{eq:ihc_coefficients}
\end{equation}
Here, $W_h \in \mathbb{R}^{2M \times Md}$, $b_{\mathrm{pre}},b_{\mathrm{post}} \in \mathbb{R}^{M}$, and $\alpha_h$ is a learnable scalar. $\mathbf{1}_M$ and $\mathbf{0}_M$ denote the all-ones and all-zeros vectors in $\mathbb{R}^{M}$, respectively, and $\sigma$ is applied elementwise. This gives $h_{\mathrm{pre}}(R) \in (0,1)^M$ and $h_{\mathrm{post}}(R) \in (0,2)^M$. We initialize $b_{\mathrm{pre}}=\operatorname{logit}(1/M)\mathbf{1}_M$ and $b_{\mathrm{post}}=\mathbf{0}_M$, so suppressing the input-dependent term yields $h_{\mathrm{pre}}=\frac{1}{M}\mathbf{1}_M$ and $h_{\mathrm{post}}=\mathbf{1}_M$. Because $\alpha_h$ is initialized to a small value and the streams begin as replicas of the input, each block starts close to a standard residual update. Input-adaptive multi-stream routing then emerges gradually during training. As described in Sec.~\ref{sec:single_stream}, this \emph{read}--compute--\emph{write} pattern is applied separately around the attention and the FFN sub-layers, with $M=4$ residual streams in our main model.

The key simplification relative to mHC lies in the residual-stream transition. In the general HC/mHC formulation, the write step additionally mixes the streams themselves through a residual transition matrix $H_{\mathrm{res}} \in \mathbb{R}^{M \times M}$, \textit{i.e.}, $R'_m = \sum_{m'=1}^{M} (H_{\mathrm{res}})_{mm'} R_{m'} + h_{\mathrm{post},m}(R)\,\Delta$, where mHC learns a token-dependent $H_{\mathrm{res}}$ and projects it onto the doubly-stochastic manifold with Sinkhorn iterations~\citep{xie2025mhc}. iHC instead fixes $H_{\mathrm{res}}=I_M$, yielding the write rule in Eq.~\ref{eq:ihc_write_rule}. This removes the per-token $M \times M$ residual mixing matrix and the Sinkhorn projection required by mHC. Beyond maintaining, reading, and writing the $M$ residual streams, the coefficient generator in Eq.~\ref{eq:ihc_coefficients} requires one RMSNorm, one $Md \to 2M$ projection, and elementwise gates per sub-layer. For fixed $M=4$, this routing arithmetic scales linearly with $d$, compared with the quadratic-in-$d$ projection cost of attention and FFNs. iHC therefore retains adaptive read and write mappings over multiple residual streams while preserving an exact identity residual transition.

\section{Scaling Recipe}
\label{sec:scaling}

This section presents the scaling recipe for \ours{}. Sec.~\ref{sec:hp_transfer} develops \heterop{}, a module-wise hyperparameter-transfer parameterization that yields a consistently tuned model family across width and depth. Section~\ref{sec:chinchilla} then uses this family to fit Chinchilla-style compute-optimal laws over model size and training tokens and extends the analysis to the image-video data ratio.

\subsection{Hyperparameter Transfer}
\label{sec:hp_transfer}

Studying the scaling behavior of a new architecture requires training a family of models of increasing scales. This immediately raises a practical question: which hyperparameters should be used at each scale? Re-tuning the learning rate, initialization, and weight decay independently at every scale is prohibitively expensive. Reusing the values tuned at a small scale is cheap but often suboptimal, because the optimal learning rate drifts with model width and depth under standard parameterizations~\citep{yang2022tensor,dey2026completep}. \emph{Hyperparameter transfer} resolves this tension by reparameterizing the model and optimizer so that a set of base hyperparameters remains approximately stable across width and depth. One can tune these base values on a small proxy model and then transfer them to larger models

Existing hyperparameter-transfer methods address complementary scaling axes. The maximal-update parameterization ($\mu$P) provides width-transfer rules and has been validated on Transformers and ResNets~\citep{yang2022tensor}, while CompleteP extends transfer across depth for Transformers~\citep{dey2026completep}. Recent diffusion-specific work further adapts $\mu$P to full-attention diffusion Transformers~\citep{zheng2026scaling}. These methods provide the principles on which we build, but do not specify a complete parameterization for a heterogeneous visual diffusion backbone that jointly combines KDA linear attention, MLA global attention, modality-aware short convolutions, sparse MoE layers, iHC, sandwich normalization, and diffusion timestep conditioning. We call our resulting method \textbf{\heterop{}} (Heterogeneous Parameterization).

\paragraph{\textbf{Why transfer is necessary for a fair scaling study.}}
Beyond reducing tuning cost, hyperparameter transfer makes a scaling study \textbf{\emph{meaningful}}. A scaling curve compares the losses of models at different sizes to estimate how performance improves with compute. If each size is trained with scale-mismatched hyperparameters, differences in measured loss may reflect both model scale and suboptimal tuning, obscuring the true effect of scale. For example, a larger model trained with a suboptimal learning rate can even underperform a smaller one, making the resulting curve unreliable for design conclusions. Hyperparameter transfer reduces this confound effect, making differences in loss primarily reflect scale and architecture rather than tuning noise. Earlier diffusion scaling studies do not explicitly control this factor with a width- and depth-transfer parameterization~\citep{peebles2023scalable,liang2024scaling}. In Sec.~\ref{sec:exp_hp}, we verify that the optimal learning rate stays stable across width and depth under our recipe, but shifts under standard parameterization.

\paragraph{\textbf{Why heterogeneous backbones require module-wise ratios.}}
A homogeneous Transformer is often widened by multiplying a common hidden dimension, making a single global width ratio a reasonable shorthand when instantiating transfer rules. In \ours{}, ``width'' is instead a collection of distinct dimensions. A projection may take as input the model hidden state, the MLA KV-compression latent, a KDA head, an MoE expert hidden state, the router state, or the timestep-conditioning state. These dimensions may scale with the model width, grow at different rates, or remain fixed. Applying the global model-width ratio to every tensor would therefore mis-scale parameters whose updates are controlled by another fan-in. \heterop{} resolves this mismatch by computing the target-to-proxy ratio separately for every parameter group from the dimension that controls its activation and update scale. The result is a fine-grained, dimension-accurate parameterization that preserves the intended transfer rule for structurally heterogeneous backbones.

\paragraph{\textbf{\heterop{} Parameterization.}}
Let $\mathcal{M}^{(0)}$ denote a small proxy model on which the base hyperparameters are tuned, and let $\mathcal{M}$ denote a target model to which they are transferred. The two models share the same architecture but may differ in width and depth, with $n_{\mathrm{blk}}^{(0)}$ and $n_{\mathrm{blk}}$ blocks, respectively. For each target parameter group $W$, let $W^{(0)}$ denote its proxy counterpart. We write the transfer compactly as:
\begin{align}
    \bar{\mathbf{h}}
        &= \left(
           \sigma_{\mathrm{base}}^{2},
           \eta_{\mathrm{base}},
           \lambda_{\mathrm{base}},
           \epsilon_{\mathrm{base}}
           \right),
           \label{eq:transfer_base} \\
    (m_W,m_L)   
        &= \left(
           \frac{\operatorname{fan\text{-}in}(W)}
                {\operatorname{fan\text{-}in}(W^{(0)})},
           \frac{n_{\mathrm{blk}}}{n_{\mathrm{blk}}^{(0)}}
           \right),
           \label{eq:transfer_multipliers} \\
    \mathcal{P}_W
        &= \mathcal{T}_{\rho(W)}
           \!\left(\bar{\mathbf{h}};m_W,m_L\right).
           \label{eq:transfer_map}
\end{align}
Here, $\sigma_{\mathrm{base}}^{2}$, $\eta_{\mathrm{base}}$, $\lambda_{\mathrm{base}}$, and $\epsilon_{\mathrm{base}}$ denote the base initialization variance, learning rate, weight decay, and AdamW $\epsilon$. $\mathcal{P}_W$ collects the initialization, optimizer, forward, and residual-scaling settings for $W$, while $\rho(W)$ denotes its functional role. $m_W$ is a module-specific width multiplier determined by the fan-in of $W$ and is independent of the number of blocks, whereas $m_L$ is the global depth multiplier determined by the block count. In tuning \ours{}, unlike previous works that adopt a single global scale ratio, $m_W$ may be determined by the model hidden width, the MLA KV-compression rank, KDA head width, MoE expert width, router width, or timestep-conditioning width. The depth multiplier $m_L$ is shared because it is determined by the target-to-proxy block-count ratio. Our proxy has model width $512$ and depth $4$ ($22$M activated parameters, Table~\ref{tab:model_family}), while the largest model has model width $2048$ and depth $32$. Hence $m_W=4$ for parameter groups whose fan-in is the model width,  while other groups receive their own ratios, and $m_L=8$. Operationally, \heterop{} (i) tunes $\bar{\mathbf h}$ once on the proxy, (ii) assigns each target tensor a role $\rho(W)$ and computes $m_W$, and (iii) applies the role-dependent map $\mathcal T_{\rho(W)}$ together with $m_L$. Eqs.~\ref{eq:transfer_base}--\ref{eq:transfer_map} define these three ingredients, and Table~\ref{tab:heterop} gives their module-wise instantiation.

\begin{table}[t]
\centering
\small
\setlength{\tabcolsep}{10pt}
\renewcommand{\arraystretch}{0.8}
\caption{
\textbf{Module-wise hyperparameter-transfer rules of \heterop{} for \ours{}, compared with standard parameterization (SP).}
We jointly control width and depth: $\mN$ is the target-to-proxy fan-in ratio computed separately for parameter group $W$ (\eg, from the model width, MLA KV-compression rank, KDA head width, MoE expert width, or router width), and $\mL{}$ is the target-to-proxy depth ratio raised to the listed power.
$\base{\eta}, \base{\sigma}^{2}, \base{\lambda}, \base{\epsilon}$ denote the learning rate, initialization variance, weight decay, and AdamW $\epsilon$ tuned once on the proxy model and transferred to all scales.
}
\label{tab:heterop}
\vspace{-1em}
{
\begin{tabular*}{\textwidth}{@{\extracolsep{\fill}}lll@{}}
\toprule
\textbf{Module / Hyperparameter} & \textbf{SP} & \textbf{\heterop{}} \\
\midrule

\multicolumn{3}{@{}l}{\textit{Input adapters} (patch / text / timestep embedding)} \\
\quad Init. Var. & $\base{\sigma}^{2}$ & $\base{\sigma}^{2}$ \\
\quad LR (AdamW) & $\base{\eta}$ & $\base{\eta}$ \\
\midrule

\multicolumn{3}{@{}l}{\textit{Sandwich RMSNorm} (pre-/post-norm gains)} \\
\quad Init. Var. & $\base{\sigma}^{2}$ & $\base{\sigma}^{2}$ \\
\quad LR (AdamW) & $\base{\eta}$ & $\base{\eta}$ \\
\midrule

\multicolumn{3}{@{}l}{\textit{Hidden weights} (KDA/MLA projections, FFN \& MoE experts, router)} \\
\quad Init. Var. & $\base{\sigma}^{2}$ & $\base{\sigma}^{2} \cdot \mNinv$ \\
\quad LR (AdamW) & $\base{\eta}$ & $\base{\eta} \cdot \mNinv$ \\
\quad Bias LR (AdamW) & $\base{\eta}$ & $\base{\eta}$ \\
\quad WD (AdamW) & $\base{\lambda}$ & $\base{\lambda} \cdot \mN$ \\
\midrule

Attention residual (KDA, MLA; iHC)
& $\mathbf{X}^{l} + \mathrm{Attn}(\mathrm{RMSNorm}(\mathbf{X}^{l}))$
& $\mathbf{X}^{l} + \mL{-1} \cdot \mathrm{Attn}(\mathrm{RMSNorm}(\mathbf{X}^{l}))$ \\

FFN / MoE residual (iHC)
& $\mathbf{Z}^{l} + \mathrm{FFN}(\mathrm{RMSNorm}(\mathbf{Z}^{l}))$
& $\mathbf{Z}^{l} + \mL{-1} \cdot \mathrm{FFN}(\mathrm{RMSNorm}(\mathbf{Z}^{l}))$ \\
\midrule

\multicolumn{3}{@{}l}{\textit{Final norm} (pre-readout)} \\
\quad Init. Var. & $\base{\sigma}^{2}$ & $\base{\sigma}^{2}$ \\
\quad LR (AdamW) & $\base{\eta}$ & $\base{\eta}$ \\
\midrule

\multicolumn{3}{@{}l}{\textit{Diffusion readout head} (velocity prediction)} \\
\quad Init. Var. & $\base{\sigma}^{2}$ & $\base{\sigma}^{2}$ \\
\quad LR (AdamW) & $\base{\eta}$ & $\base{\eta}$ \\
\quad Forward
& $\mathbf{X}^{L}\mathbf{W}_{\mathrm{head}}^{\top}$
& $\mathbf{X}^{L}\mathbf{W}_{\mathrm{head}}^{\top} \cdot \mNinv$ \\
\midrule

AdamW $\epsilon$ (residual blocks)
& $\base{\epsilon}$
& $\base{\epsilon} \cdot \mNinv \cdot \mL{-1}$ \\

AdamW $\epsilon$ (input adapters \& readout)
& $\base{\epsilon}$
& $\base{\epsilon} \cdot \mNinv$ \\
\bottomrule
\end{tabular*}
}\label{tab:mup_arceus}
\vspace{-1.0em}
\end{table}

\paragraph{\textbf{Width and depth transfer rules.}}
The \heterop{} rules in Table~\ref{tab:heterop} combine module-wise hyperparameter transfer across both width and depth. Under width transfer, hidden matmul weights use initialization variance and learning rate scaled by $\mNinv$, together with weight decay scaled by $\mN$.
These initialization and learning-rate corrections preserve width-consistent activation and update scales, while the paired scaling $\eta \propto \mNinv$ and $\lambda \propto \mN$ keeps $\eta\lambda$ invariant.
Input adapters, normalization gains, biases, and the readout keep their base initialization and learning rate, matching the input/output treatment of $\mu$P. The readout is additionally multiplied by $\mNinv$ in the forward pass, and AdamW $\epsilon$ is scaled by $\mNinv$ to track the shrinking gradient scale.

\heterop{} incorporates the depth correction of CompleteP~\citep{dey2026completep}: each attention or FFN/MoE sub-layer output is scaled by $\mL{-1}$ before the corresponding iHC write, keeping the aggregate residual contribution on a depth-independent scale. Residual-block tensors receive the same additional $\mL{-1}$ correction to AdamW $\epsilon$, while their learning rates require no additional depth correction, as listed in Table~\ref{tab:heterop}.

\paragraph{\textbf{Module-wise assignments.}}
Mapping \heterop{} onto \ours{} requires assigning each learnable tensor to a rule. For each parameter group, we identify its functional role and, where applicable, the dimension that determines its fan-in. The resulting assignment is summarized below.
\begin{itemize}
    \item \textbf{Input adapters and readout.} The patch, text, and timestep embedding projections are input layers, and the velocity-prediction head is the output layer. Both keep the base learning rate and initialization, and the readout additionally uses the $\mNinv$ forward multiplier. The blockwise timestep-conditioning MLP in Eq.~\ref{eq:adln_mlp}, however, follows the hidden rule; its input projection is parameterized by the width of $e_\tau$.
    \item \textbf{Hybrid attention.} All projections that read the hidden state, including the KDA query--key--value projection and the MLA query and key--value down-projections, scale with the model width. The MLA key--value up-projection scales with the KV-compression rank, while the KDA gating and the output projections of both operators scale with the corresponding head widths. The short-convolution kernels and the KDA low-rank gate up-projections have a fixed fan-in per output channel. They therefore use $\mN=1$, retaining the base learning rate and weight decay while using the residual-block $\epsilon$ scaling. The complementary gate down-projections read the hidden state, so they follow the hidden rule.
    \item \textbf{FFN and MoE.} The up- and gate-projections of dense SwiGLU FFNs and MoE experts scale with the model width, and the down-projections scale with their respective hidden widths. The router input projection likewise follows the model width, while the router hidden width is held fixed across scales. This keeps the routing logits on a consistent scale as the model grows. The routing sparsity is also fixed, as discussed below.
    \item \textbf{iHC and Sandwich Normalization.} The iHC gains and biases, together with the RMSNorm gains, are control tensors and follow the norm treatment: base initialization, base learning rate, and the hidden $\epsilon$. The lightweight projection that produces the iHC read/write vectors reads the concatenated residual streams, whose size grows with the model width, so it follows the hidden rule. This preserves residual stability across width and depth.
\end{itemize}

\paragraph{\textbf{What is held fixed when scaling.}}
When constructing the model family, we vary the module-specific widths and block count represented by $\mN$ and $\mL{}$, respectively, while holding the structural settings listed below fixed. The MoE routing configuration is constant across scales: every model uses $56$ routed experts with $8$ activated per token, an expert activation ratio of $1/7$ (Sec.~\ref{sec:moe}). The MLA KV-compression ratio, defined as the compression rank divided by the model width, is likewise constant as width grows, so compression acts as a fixed architectural property rather than a hidden scaling variable.\footnote{Whether the optimal compression ratio is itself scale-dependent is an interesting question that we leave to future work. We use the proxy value.} The same holds for the KDA-to-MLA layer ratio, the iHC residual-stream count, and the patchification factors. Fixing these settings prevents structural choices from being entangled with the scaling study and keeps all data points mutually comparable.

\subsection{Chinchilla Laws for \ours{}} \label{sec:chinchilla}

\heterop{} yields mutually comparable \ours{} models with scale-appropriate hyperparameters. Using this family, we fit Chinchilla-style compute-optimal scaling laws~\citep{hoffmann2022chinchilla} for visual diffusion.

\paragraph{\textbf{Why a compute-optimal law is needed.}}
A central practical question in scaling is allocation. Given a fixed training budget of $C$ FLOPs, how large should the model $N$ be, and on how many tokens $D$ should it process? Training an overly large model for too few steps wastes capacity, whereas training an overly small model for too long spends compute on a saturated model. \citet{hoffmann2022chinchilla} show that, for language models, the loss-minimizing split between $N$ and $D$ follows a predictable trade-off, and that ignoring it yields substantially undertrained models. For visual diffusion, this trade-off has not been characterized on a model family with hyperparameters controlled across scales through \heterop{} (Sec.~\ref{sec:hp_transfer}). Our goal is therefore to make the size-versus-data trade-off of \ours{} explicit and predictive, so that the configuration of a large target run can be chosen \emph{a priori} from the fitted law rather than by trial and error.

\paragraph{\textbf{Formulation.}}
We adopt the parametric form of \citet{hoffmann2022chinchilla}, modeling the training loss as a function of the model size $N$ and the amount of training data $D$:
\begin{equation}
    \widehat{\mathcal{L}}(N, D) = E + \frac{A}{N^{\,a}} + \frac{B}{D^{\,b}},
    \label{eq:chinchilla}
\end{equation}
where $E$ denotes the irreducible loss under the fixed data distribution and diffusion objective, \textit{i.e.}, the Bayes-optimal diffusion loss on the data distribution, and the two power-law terms capture the finite-size penalties from limited capacity and limited data. Three measurement choices adapt this form to our setting. First, the loss $\widehat{\mathcal{L}}$ is the \ours{} \emph{diffusion loss}, the rectified-flow $v$-prediction objective of Sec.~\ref{sec:model_arch_preliminary}, rather than a downstream generation score such as FID. Under the fixed objective and evaluation protocol used here, diffusion loss provides a sampler-independent metric for comparing pre-training across model scales. Second, because \ours{} is sparsely activated, $N$ counts \emph{activated} parameters, so the law reflects the compute actually spent per token rather than the total parameter store.\footnote{The load balance across experts does not affect computational cost, but it can affect model quality and hence the losses used for fitting. We therefore conduct this study under near-perfect load balance, as demonstrated in Fig.~\ref{fig:moe_balance}.} Third, $D$ denotes the cumulative training-data volume, measured by the number of visual latent positions processed, and we relate it to compute through the standard accounting $C \approx 6ND$ for the dominant matmul cost,\footnote{This approximation remains valid for the hybrid backbone. Most layers use linear attention with constant per-token cost, and the few MLA layers attend over compressed key--value states, so parameter matmuls dominate the overall compute.} which lets us express the law and its optimum directly in terms of compute.

\paragraph{\textbf{Fitting and usage.}}
All runs share the same base AdamW configuration and a warmup-then-constant learning-rate schedule without decay. Thus, the loss measured at an intermediate step is not affected by a learning-rate decay phase tailored to that training duration. A single run can therefore provide comparable loss measurements across multiple data budgets $D$. We further ensure that all runs remain in the infinite-data regime: no run processes more than one epoch of the training corpus. Given a set of runs spanning model sizes and durations, we estimate the compute-optimal allocation with the three estimators of \citet{hoffmann2022chinchilla}: the lower envelope of the loss-versus-compute curves, IsoFLOP profiles that compare losses across model sizes at fixed compute budgets, and a direct parametric fit of Eq.~\ref{eq:chinchilla} that minimizes the Huber loss between predicted and observed log-losses over a grid of initializations. The fitted law yields the compute-optimal frontier $N_{\mathrm{opt}}(C)$ and $D_{\mathrm{opt}}(C)$ in closed form. We use this frontier to select the configuration of \ours{}'s largest runs, and report the fits and their cross-estimator agreement in Sec.~\ref{sec:exp_law}.

\paragraph{\textbf{Extending the law to the image--video data ratio.}}
Unlike scaling laws for single-modality text data, visual generation trains on multimodal data that mixes images and videos. The two differ in information content and in the number of tokens each sample contributes, so the mixture shifts both the learning signal and how the compute budget is spent. The standard law over $(N, D)$ leaves this axis uncontrolled. We therefore extend the scaling law with an additional dimension, the image--video data ratio, and model how its optimal value changes as the training budget increases.

Specifically, we generalize the IsoFLOP methodology of \citet{hoffmann2022chinchilla}. We train models of multiple sizes under multiple image--video ratios and record their diffusion losses at fixed compute budgets, tracking the image and video losses separately. At each budget, instead of fitting a one-dimensional quadratic profile over model size alone, we fit a quadratic surface over model size and data ratio for each loss. We then select the optimum of every fitted surface and fit the compute-optimal frontier to these per-budget optima, obtaining the optimal model size and image--video ratio as functions of compute. Sec.~\ref{sec:exp_ratio} presents the fitted surfaces and the resulting frontier.

\section{Experiments} 
\label{sec:experiment}

We evaluate \ours{} along the two axes developed in this paper. We first validate the scaling methodology: Sec.~\ref{sec:exp_hp} verifies \heterop{} transfer across width and depth, Sec.~\ref{sec:exp_law} presents compute-optimal scaling laws for image and video generation, and Sec.~\ref{sec:exp_ratio} extends the law to the image-video data mixture. We then evaluate the models: Sec.~\ref{sec:exp_quality} compares \ours{} with strong baselines on generation quality, and Sec.~\ref{sec:exp_ablation} presents component ablations.

\subsection{Implementation details.}
\label{sec:implementation_details}

\begin{figure*}[t]
    \centering
    \includegraphics[width=0.95\linewidth]{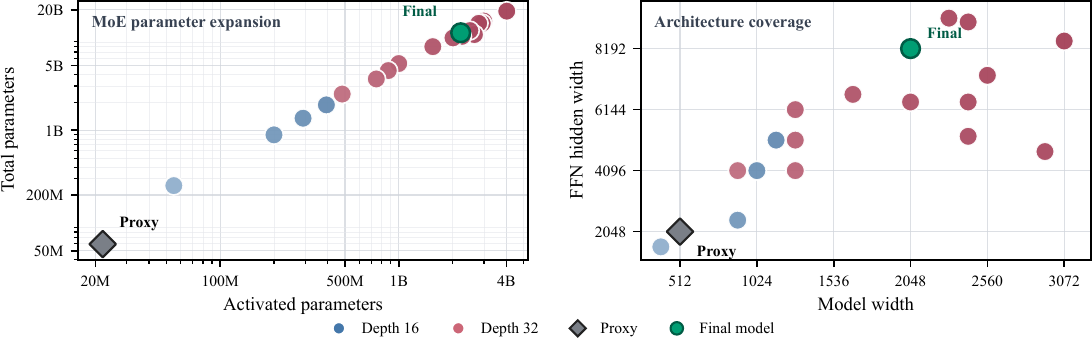}
    \vspace{-1em}
    \caption{
    {\textbf{Representative \ours{} model family.} Left: total versus activated parameters, showing the parameter expansion. Right: model width versus FFN hidden width, showing the configurations covered by the scaling study. The family is not a strict fixed-aspect-ratio scaling ladder; instead, it spans feasible combinations of width, depth, and FFN capacity while keeping the high-level architecture fixed. We primarily fix the depth and study the scaling of width. All markers use the same diameter; hue denotes depth, while within-depth color intensity increases logarithmically with activated parameter count. The proxy and final training model are highlighted separately. Exact configurations are provided in Appendix~\ref{app:model_family}.}
    }
    \label{fig:model_family}
    \vspace{-0.5em}
\end{figure*}

All models instantiate the architecture of Sec.~\ref{sec:model_arch} and differ only in width, FFN width, and depth.
Fig.~\ref{fig:model_family} visualizes representative configurations spanning $55$M to $4$B activated parameters, of which the exact specifications are listed in Appendix~\ref{app:model_family}.
As specified in Sec.~\ref{sec:hp_transfer}, all structural ratios are held fixed across scales, including the MoE configuration, the KDA-to-MLA ratio, the residual-stream count, the MLA KV-compression ratio, and the patchification factors.

\paragraph{\textbf{Frozen components.}}
Visual inputs are encoded by a frozen causal 3D VAE~\citep{wan2025wan} with $16$ latent channels and $(4,8,8)$ compression along the temporal, height, and width axes. A clip of $4k+1$ frames at resolution $H\times W$ is encoded into $k+1$ latent frames of size $\tfrac{H}{8}\times\tfrac{W}{8}$, and a single image is the special case $k=0$. Encoded latents are standardized with the per-channel mean and standard deviation of the VAE latent space, so the rectified-flow objective operates on approximately zero-mean, unit-variance targets. We then apply $(1,2,2)$ patchification. Text prompts are encoded by a frozen umT5-XXL encoder~\citep{chung2023unimax} with a maximum length of $512$ tokens. Padding positions are dropped and the remaining embeddings are packed into a variable-length sequence that is prepended to the visual tokens through a separate trainable projection (Sec.~\ref{sec:single_stream}).

\paragraph{\textbf{Training recipe.}}
All models are trained with the rectified-flow objective of Sec.~\ref{sec:model_arch_preliminary}, and the loss is the mean squared error on the velocity target, unweighted across timesteps. One timestep is drawn per sample and shared by all of its tokens, from a logit-normal distribution~\citep{esser2024scaling} that concentrates supervision on intermediate noise levels. Since a fixed timestep destroys more information in a longer token sequence, we then shift it toward higher noise by $\tau' = \tau / (\tau + s(1-\tau))$, with $s = \sqrt{L_{\mathrm{vis}} / 256}$ for a sample of $L_{\mathrm{vis}}$ visual tokens, taking the $256^2$ image bucket as the reference. Because $s$ is computed per sample, images and video clips of different sizes are shifted independently within the same batch.

Text conditioning is dropped with probability $0.1$ to enable classifier-free guidance at inference. We optimize with AdamW using a base learning rate $\base{\eta} = 10^{-3}$, betas $(0.95, 0.95)$~\citep{orvieto2026search,fernandez2026adam,cattaneo2026effect}, $\base{\epsilon} = 10^{-8}$, and a base weight decay $\base{\lambda} = 0.01$. \heterop{} instantiates the parameter-group learning rates, initialization variances, weight decays, and AdamW $\epsilon$ values at each scale according to Table~\ref{tab:heterop}. The base learning rate warms up linearly from $10^{-6}$ over $2{,}000$ steps and then remains constant. The constant schedule keeps intermediate checkpoints comparable across training durations, which the scaling study in Sec.~\ref{sec:exp_law} relies on.

Gradients are clipped to a global norm of $1.0$. As a stability guard, we skip the optimizer and EMA update at any step whose pre-clipping gradient norm is non-finite or exceeds $10$. Expert load is balanced by the auxiliary-loss-free bias update of Sec.~\ref{sec:moe}, with the selection bias adjusted by $10^{-3}$ per step. No load-balancing term is therefore added to the training objective, and the reported loss is exactly the flow-matching loss above, which matters for the scaling study in Sec.~\ref{sec:exp_law} where this loss is the primary metric. We additionally maintain an exponential moving average of the trainable parameters with decay $0.9999$, updated every step.

\paragraph{\textbf{Data.}}
We train on a large-scale curated visual corpus.
Training uses multi-resolution pixel buckets: an early stage mixes $256^2$, $512^2$, and $1024^2$ buckets with sampling ratios $4{:}2{:}2$. Mixed-resolution batches yield highly variable sequence lengths, so a data sequence balancer~\citep{zhang2025knapformer} redistributes packed sequences across devices to balance the per-GPU load. Samples with aspect ratio above $4{:}1$ or below $1{:}4$ are discarded. 
During mixed image--video training, we retain the three image buckets and add approximately 5-second video clips at 180p and 360p, using loader sampling ratios $2{:}1{:}1{:}2{:}2$ for $256^2$, $512^2$, $1024^2$, 180p video, and 360p video, respectively. Each clip contains 81 frames sampled at 16 fps and is assigned to an area-preserving crop bucket according to its aspect ratio. Additionally, we conduct all trials in Secs.~\ref{sec:exp_hp} and~\ref{sec:exp_law} using a deterministic data loader and under an infinite-training-data regime.

\paragraph{\textbf{Infrastructure.}}
We train \ours{} with Hybrid Sharded Data Parallel (HSDP)~\citep{zhao2023pytorch}, sharding model states across the eight GPUs within each node while replicating them across nodes. HSDP is applied at block granularity to overlap parameter communication with computation. Parameters are gathered in \texttt{bf16}, gradients are reduced in \texttt{fp32}. We further apply activation checkpointing to every block. The dominant per-layer operators are implemented directly over the packed token representation using custom Triton kernels. These include a fused multimodal short convolution, chunked KDA with fused gating and normalization, and an MoE implementation based on expert-wise token sorting and grouped matrix multiplications.

Mixed image--video training can produce substantial load imbalance because different ranks may process samples with widely varying computational costs. We therefore introduce two lightweight workload balancers. For VAE encoding, ranks within each node exchange sample-shape metadata and estimate encoding cost from pixel count. Samples are assigned greedily in descending order of cost to the currently least-loaded rank, transferred in a single all-to-all, and encoded latents are returned to their original ranks and order through a second all-to-all. We additionally apply a DiT balancer before the first transformer block, following \citep{zhang2025knapformer}. It estimates the cost of each packed sequence using a model that captures both the linear cost of KDA-dominated computation and the quadratic cost of periodic MLA layers, and redistributes complete sequences through a differentiable all-to-all so that outputs and gradients are routed back to their owners. Both assignments are recomputed at every step and applied only when the estimated reduction in the maximum per-rank workload exceeds a small threshold.

\paragraph{\textbf{Evaluation protocol.}}
We evaluate \ours{} at two levels. For the scaling study, the primary metric is the \ours{} diffusion training loss of Sec.~\ref{sec:model_arch_preliminary}. Under a fixed objective and noise parameterization, this loss is sampler-independent and therefore comparable across model sizes, training durations, and data mixtures, which the analyses in Sec.~\ref{sec:exp_law} rely on. For the final models, we additionally report downstream generation quality. We assess image generation on the GenEval~\citep{ghosh2023geneval} and DPG-Bench~\citep{hu2024ella} benchmarks, and video generation with FID and FVD under the length-extrapolation protocol of Sec.~\ref{sec:exp_quality}. To keep comparisons controlled, we hold the sampling configuration fixed across the variants under test within each such comparison, so that measured differences reflect the models rather than the sampling settings. We detail the setup specific to each experiment in the corresponding subsection.

\subsection{\heterop{} Hyperparameter Transfer}
\label{sec:exp_hp}

\begin{figure}[t]
    \centering
    \vspace{-0.8em}

    \begin{subfigure}[h]{0.49\linewidth}
        \centering
        \includegraphics[width=\linewidth]{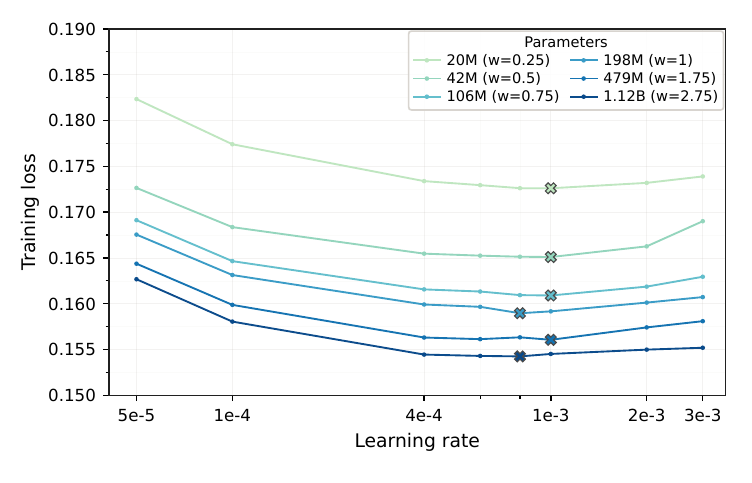}
        \vspace{-0.35in}
        \caption{\heterop{}: width transfer.}
        \label{fig:heterop_width}
    \end{subfigure}
    \hfill
    \begin{subfigure}[h]{0.49\linewidth}
        \centering
        \includegraphics[width=\linewidth]{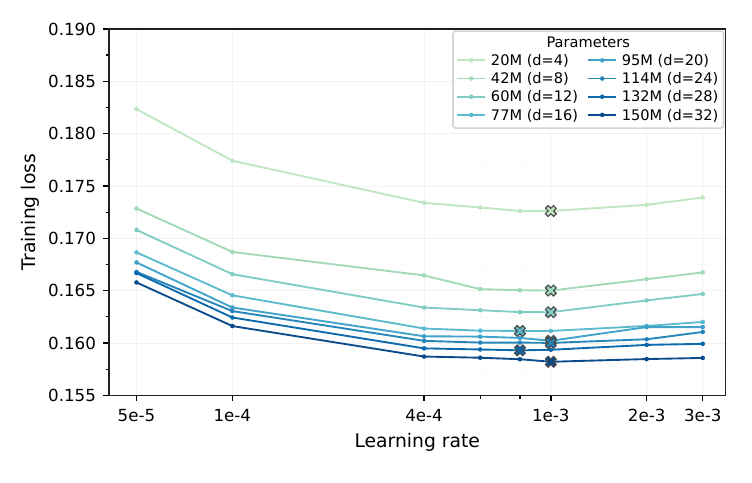}
        \vspace{-0.35in}
        \caption{\heterop{}: depth transfer.}
        \label{fig:heterop_depth}
    \end{subfigure}
    \\
    \vspace{-0.2em}
    \begin{subfigure}[h]{0.49\linewidth}
        \centering
        \includegraphics[width=\linewidth]{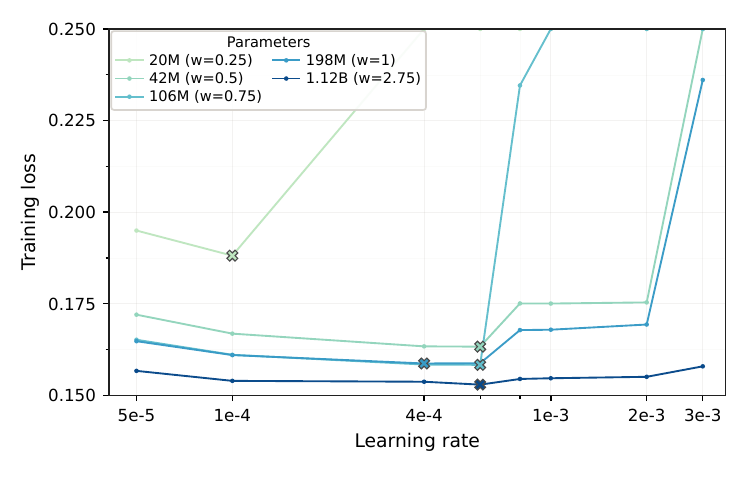}
        \vspace{-0.35in}
        \caption{SP: width sweep.}
        \label{fig:sp_width}
    \end{subfigure}
    \hfill
    \begin{subfigure}[h]{0.49\linewidth}
        \centering
        \includegraphics[width=\linewidth]{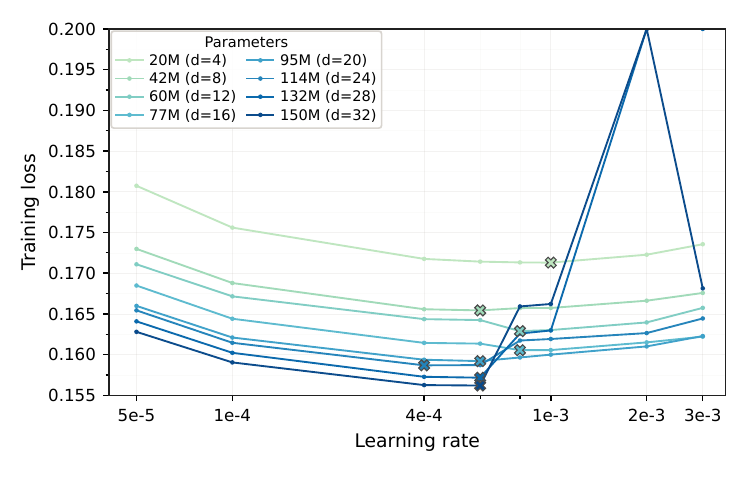}
        \vspace{-0.35in}
        \caption{SP: depth sweep.}
        \label{fig:sp_depth}
    \end{subfigure}
    \vspace{-0.7em}
    \caption{
    \textbf{\heterop{} transfers learning rates across width and depth.}
    Top row: \heterop{}; bottom row: standard parameterization (SP).
    Panels (a,c) vary width 0.25× to 2.75× (20M to 1.12B activated parameters) and panels (b,d) vary depth from 4 to 32 layers at fixed width (20M to 150M activated parameters). The crosses mark the minimum after $50$k steps.
    The horizontal axis denotes the base learning rate $\eta_{\mathrm{base}}$ under our recipe and the unscaled global learning rate under SP.
    {With \heterop{}, the optimal learning rate stays at ${\approx}10^{-3}$ across $56\times$ in model size and $8\times$ in depth}. Under SP, it shifts with scale, and high-rate runs can become unstable.
    }
    \label{fig:heterop_lr_transfer}
    \vspace{-1.5em}
\end{figure}

We first verify the central premise of the scaling study: under \heterop{}, the base learning rate $\eta_{\mathrm{base}}$ tuned on the proxy model remains near-optimal at larger scales.

\paragraph{\textbf{Setup.}}
We sweep the base learning rate $\eta_{\mathrm{base}}$ from $5\times10^{-5}$ to $3\times10^{-3}$ on two model families that vary one scaling axis at a time. Under our recipe, every model is trained for $50$k steps at each candidate $\eta_{\mathrm{base}}$, with the parameter-group learning rates and other scaled optimizer values instantiated according to Table~\ref{tab:mup_arceus}. For comparison, we repeat the same sweeps under standard parameterization (SP), where the swept learning rate is applied globally without \heterop{}'s transfer rules. The width sweep spans $20$M to $1.12$B activated parameters. The depth sweep varies the number of layers from $4$ to $32$, spanning $20$M to $150$M activated parameters. Across both sweeps, all other architectural and training settings are held fixed as specified in Sec.~\ref{sec:implementation_details}.

\paragraph{\textbf{Results.}}
Fig.~\ref{fig:heterop_lr_transfer} plots the training loss after $50$k steps against the base learning rate $\eta_{\mathrm{base}}$ for both families. Two observations support the \heterop{} transfer recipe. First, the optimal base learning rate is approximately stable across scales: every configuration attains its minimum near $\eta_{\mathrm{base}} = 10^{-3}$, drifting by at most one point on the sweep grid across a $56\times$ range of model sizes (Fig.~\ref{fig:heterop_width}) and an $8\times$ range of depths (Fig.~\ref{fig:heterop_depth}), and the loss basin around the optimum remains wide and flat at all scales. Second, the curves are cleanly ordered: at every learning rate, wider and deeper models achieve strictly lower loss with no crossovers, indicating that the parameterization preserves the benefit of scale rather than trading it for stability. Together, these results support transferring the base learning rate tuned on the proxy model across both axes, and we therefore fix $\eta_{\mathrm{base}} = 10^{-3}$ for every run in the scaling study of Sec.~\ref{sec:exp_law}. In contrast, under SP, the optimal global learning rate shifts with scale and several high-rate runs become unstable (Figs.~\ref{fig:sp_width} and~\ref{fig:sp_depth}), leading to suboptimal training and less reliable measurements for fitting the scaling law (discussed later in Sec.~\ref{sec:exp_ablation}).

\begin{figure*}[t]
    \centering

    \begin{subfigure}[t]{0.32\textwidth}
        \centering
        \includegraphics[width=\linewidth]{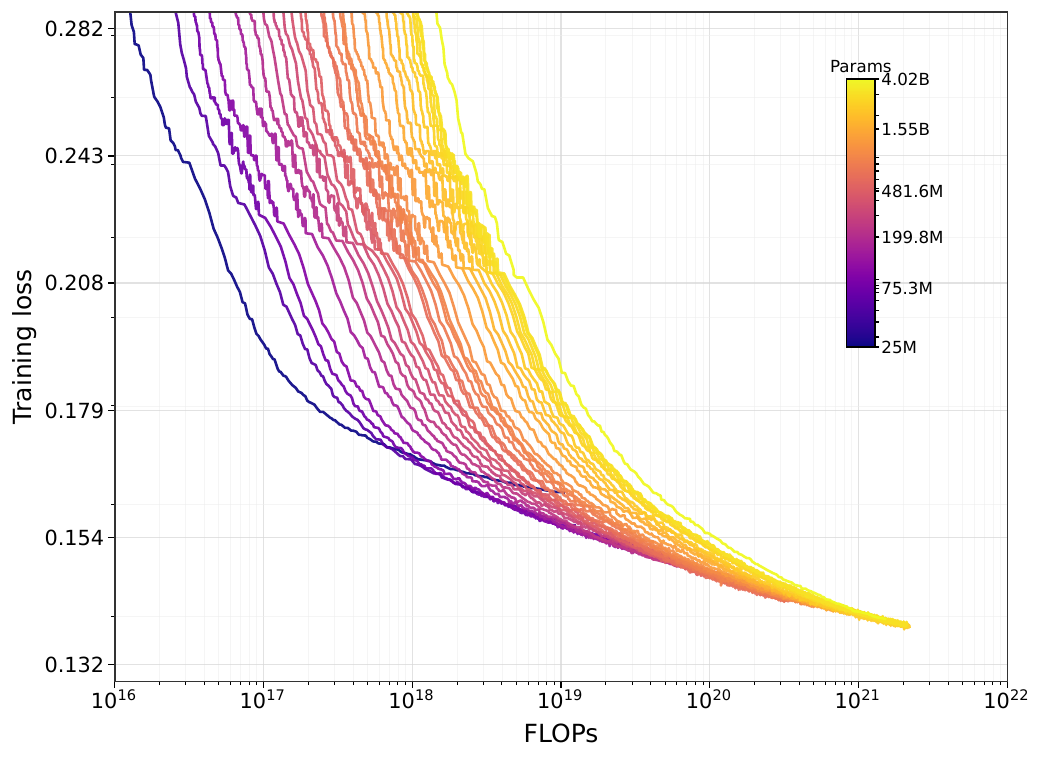}
        \caption{Training-loss envelope.}
        \label{fig:loss_flops_256}
    \end{subfigure}
    \hfill
    \begin{subfigure}[t]{0.32\textwidth}
        \centering
        \includegraphics[width=\linewidth]{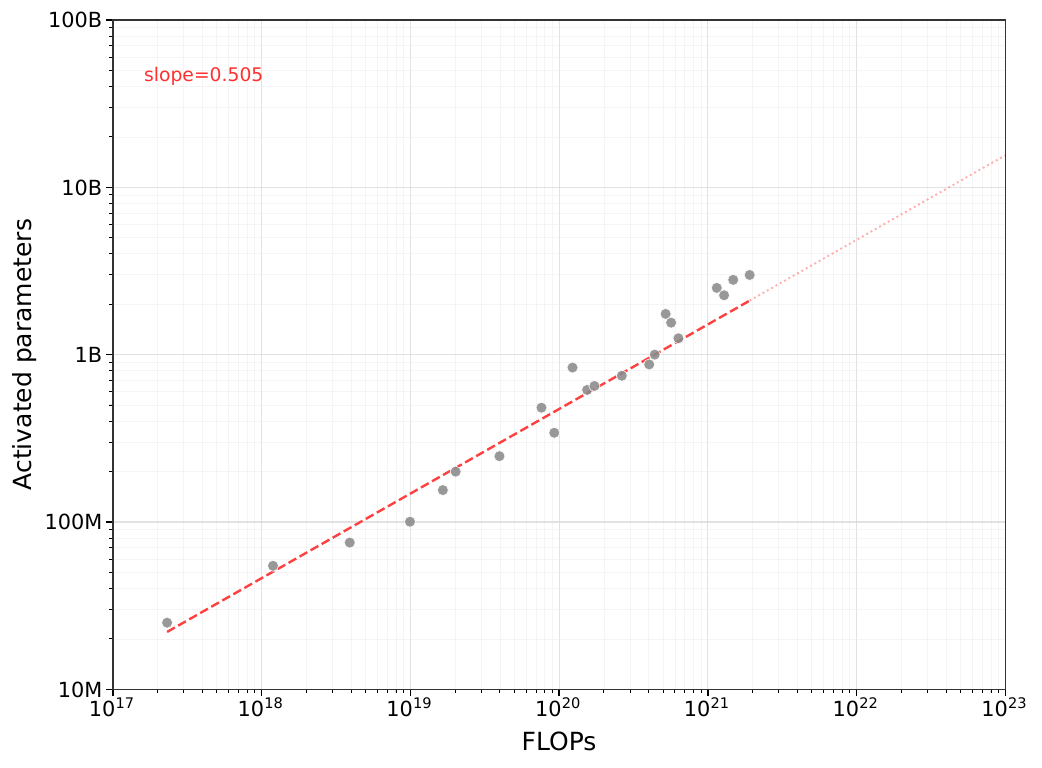}
        \caption{Envelope-optimal model size.}
        \label{fig:params_flops_256}
    \end{subfigure}
    \hfill
    \begin{subfigure}[t]{0.32\textwidth}
        \centering
        \includegraphics[width=\linewidth]{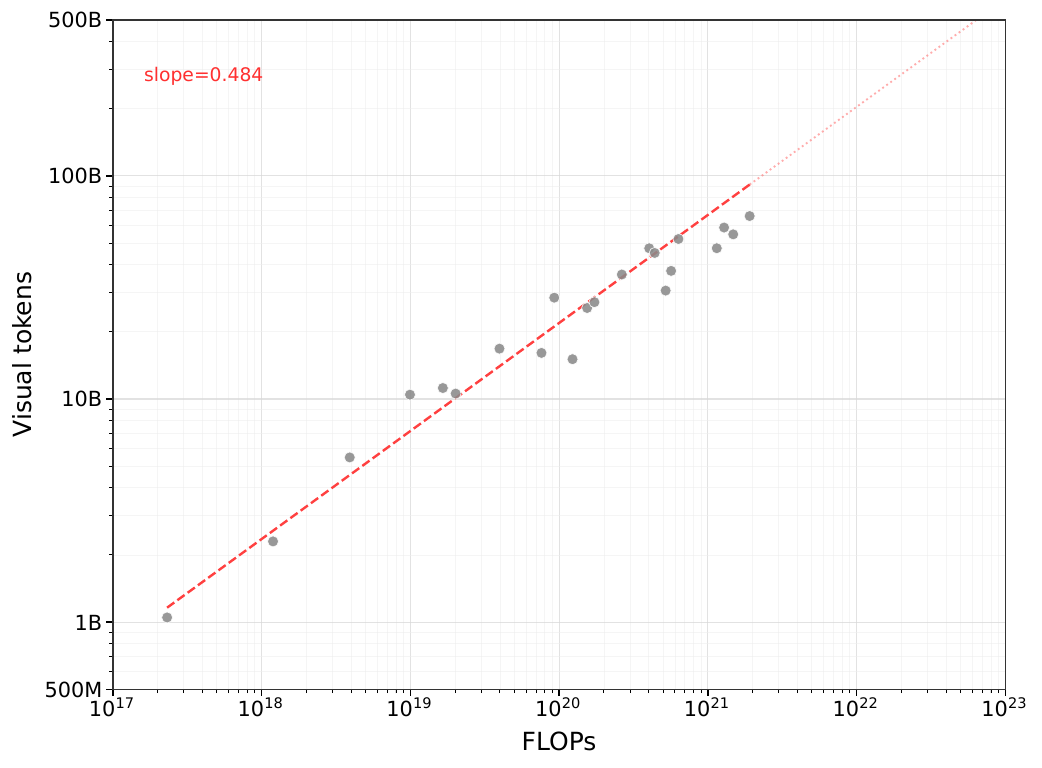}
        \caption{Envelope-optimal training tokens.}
        \label{fig:tokens_flops_256}
    \end{subfigure}

    \par\vspace{0.75em}

    \begin{subfigure}[t]{0.32\textwidth}
        \centering
        \includegraphics[width=\linewidth]{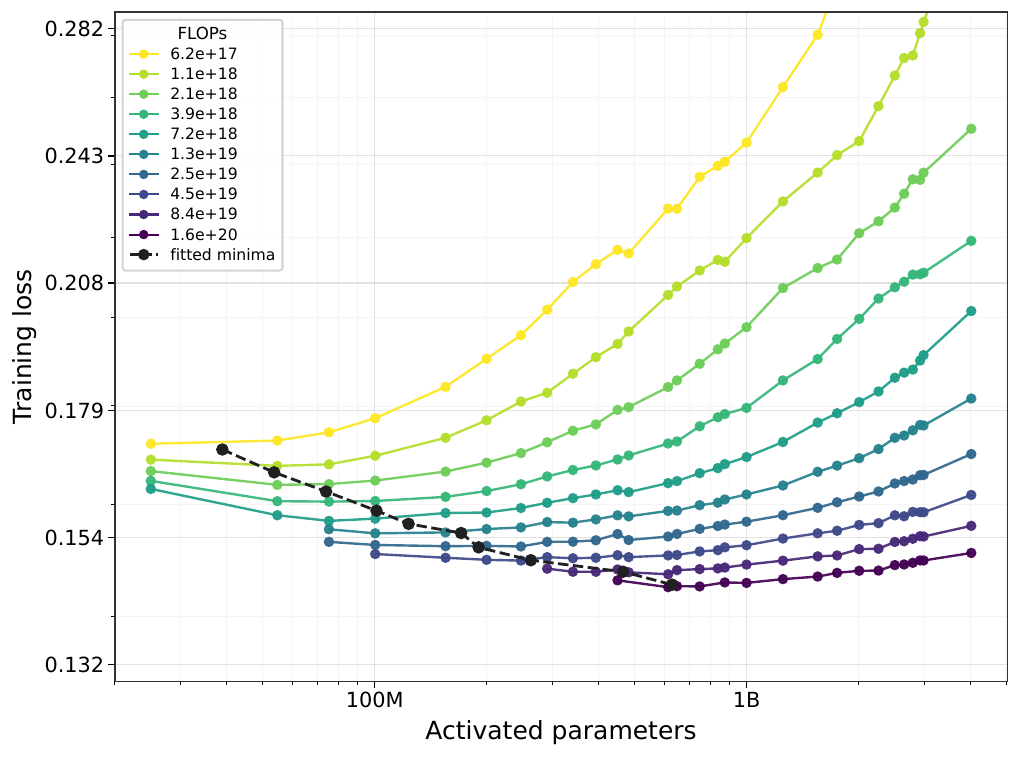}
        \caption{IsoFLOP loss profiles.}
        \label{fig:iso_loss_params_256}
    \end{subfigure}
    \hfill
    \begin{subfigure}[t]{0.32\textwidth}
        \centering
        \includegraphics[width=\linewidth]{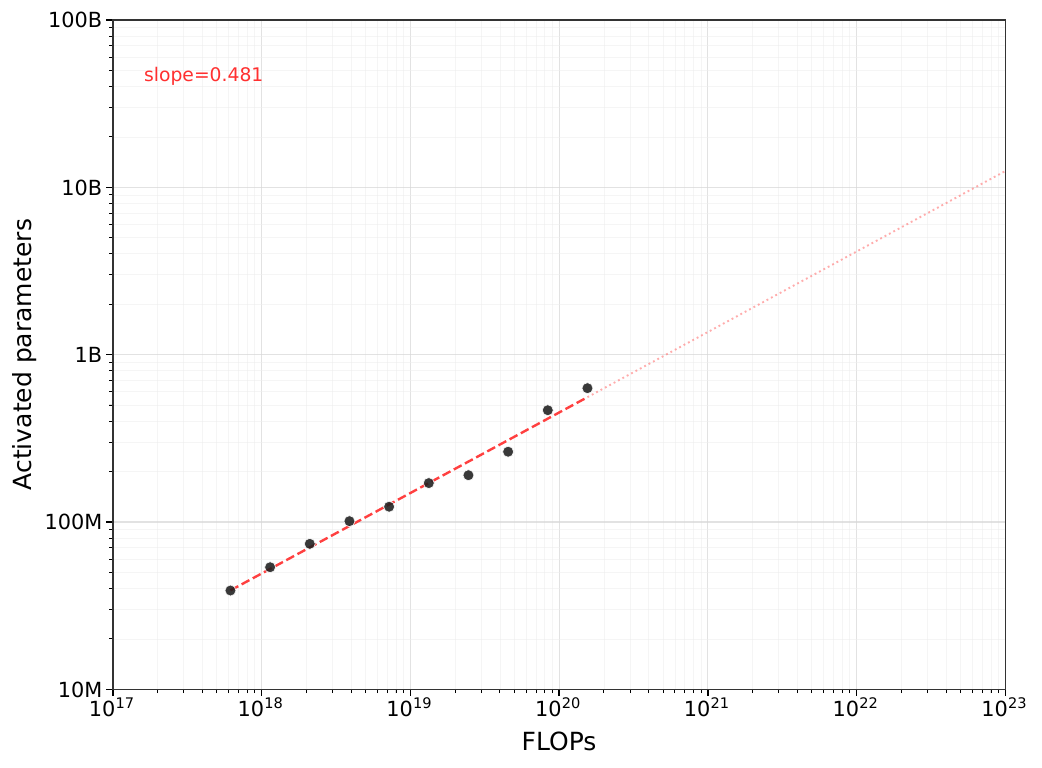}
        \caption{IsoFLOP-optimal model size.}
        \label{fig:iso_params_flops_256}
    \end{subfigure}
    \hfill
    \begin{subfigure}[t]{0.32\textwidth}
        \centering
        \includegraphics[width=\linewidth]{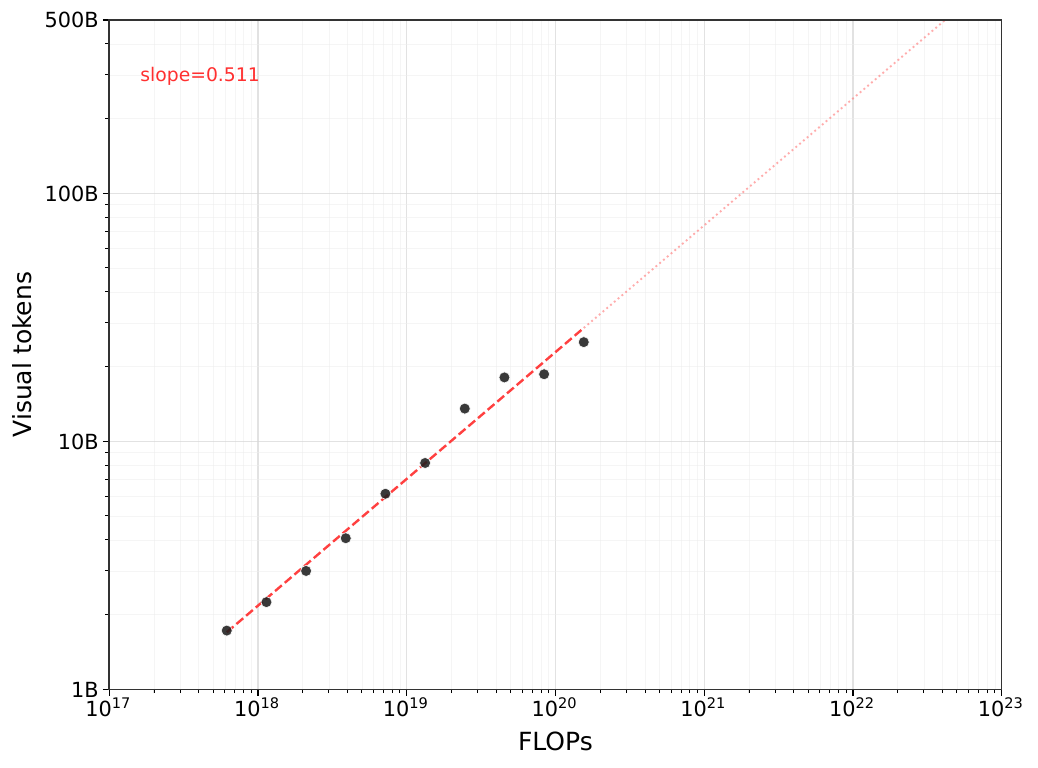}
        \caption{IsoFLOP-optimal training tokens.}
        \label{fig:iso_tokens_flops_256}
    \end{subfigure}

    \caption{
    \textbf{Compute-optimal scaling of \ours{} for $256^2$ image pre-training.}
    \textbf{Top row, training-loss envelope:}
    (a)~Diffusion training loss versus cumulative training FLOPs; the lower envelope identifies the best checkpoint at each compute budget.
    (b)~Compute-optimal activated model size $N_{\mathrm{opt}}(C)$ from the envelope-winning checkpoints.
    (c)~The corresponding compute-optimal training tokens $D_{\mathrm{opt}}(C)$, obtained using each model's logged per-step FLOPs.
    \textbf{Bottom row, IsoFLOP profiles:}
    (d)~Diffusion loss versus activated model size at fixed compute budgets; the marked profile minima identify the optimal model at each budget.
    (e)~Compute-optimal activated model size inferred from these minima.
    (f)~The corresponding compute-optimal training tokens.
    Dashed lines show the fitted power laws, with their exponents annotated in the scaling panels.
    }
    \label{fig:chinchilla_256img}
    \label{fig:iso_chinchilla_256img}
    \vspace{-1em}
\end{figure*}

\begin{figure*}[t]
    \centering

    \begin{subfigure}[t]{0.32\textwidth}
        \centering
        \includegraphics[width=1.05\linewidth]{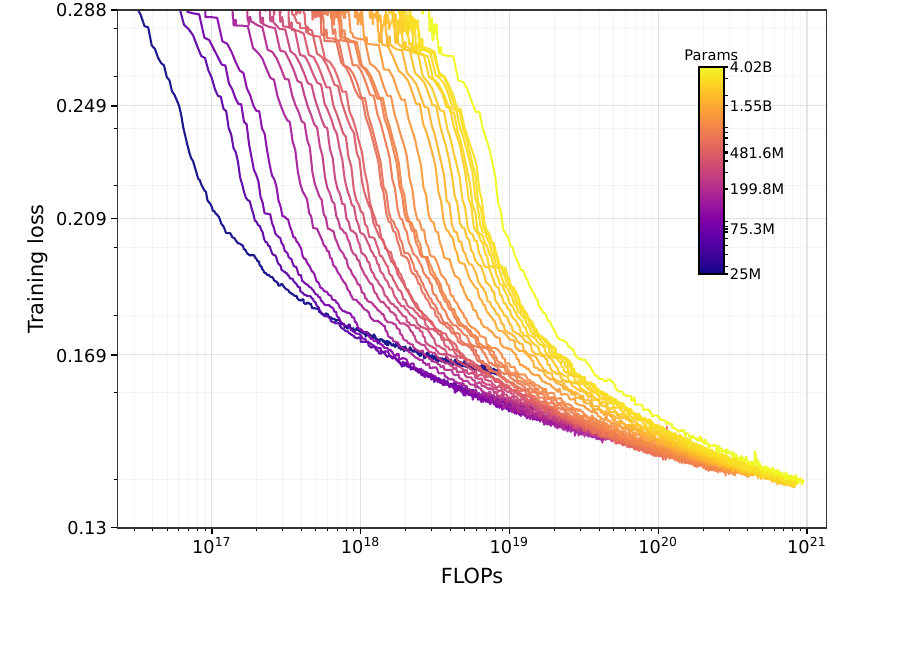}
        \vspace{-0.3in}
        \caption{Training-loss envelope.}
        \label{fig:loss_flops_180pvideo}
    \end{subfigure}
    \hfill
    \begin{subfigure}[t]{0.32\textwidth}
        \centering
        \includegraphics[width=1.05\linewidth]{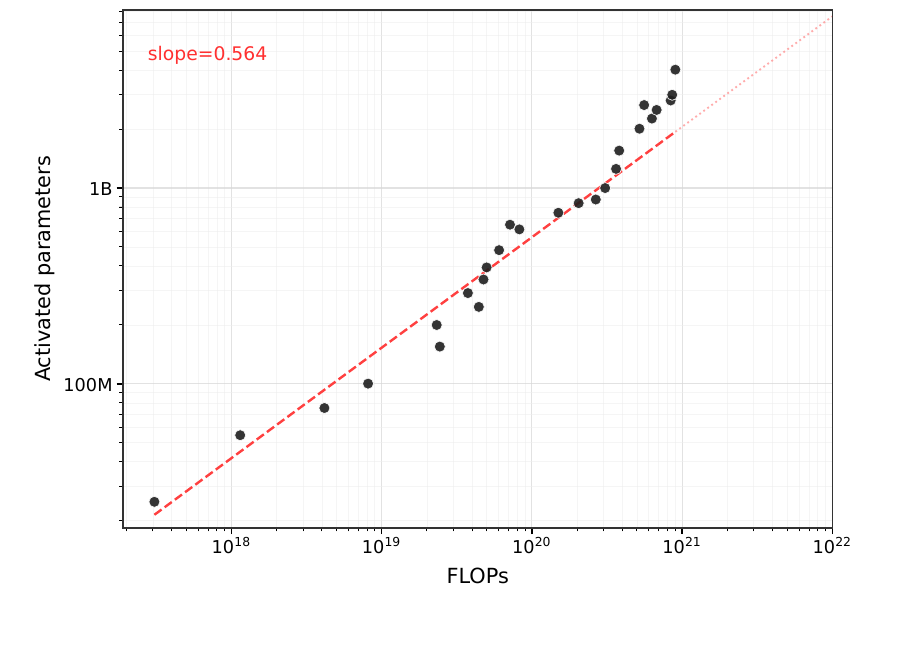}
        \vspace{-0.3in}
        \caption{Envelope-optimal model size.}
        \label{fig:params_flops_180pvideo}
    \end{subfigure}
    \hfill
    \begin{subfigure}[t]{0.32\textwidth}
        \centering
        \includegraphics[width=\linewidth]{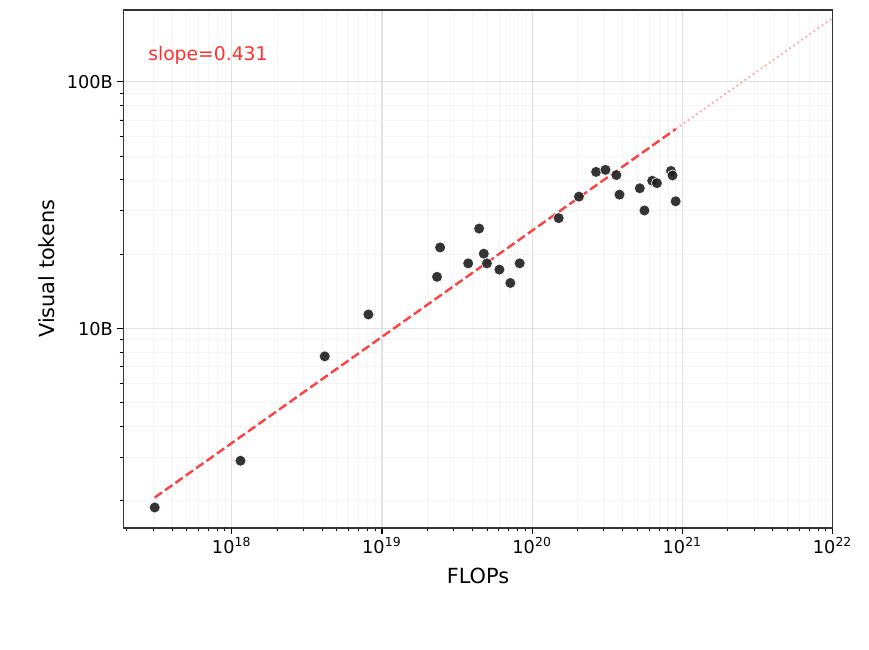}
        \vspace{-0.3in}
        \caption{Envelope-optimal training tokens.}
        \label{fig:tokens_flops_180pvideo}
    \end{subfigure}

    \par\vspace{0.75em}

    \begin{subfigure}[t]{0.32\textwidth}
        \centering
        \includegraphics[width=1.05\linewidth]{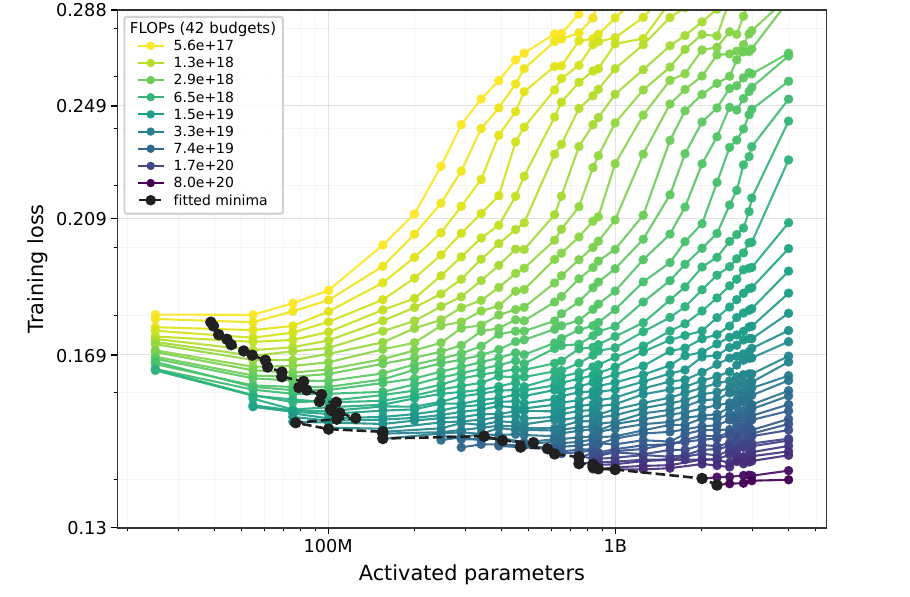}
        \caption{IsoFLOP loss profiles.}
        \label{fig:iso_loss_params_180pvideo}
    \end{subfigure}
    \hfill
    \begin{subfigure}[t]{0.32\textwidth}
        \centering
        \includegraphics[width=1.05\linewidth]{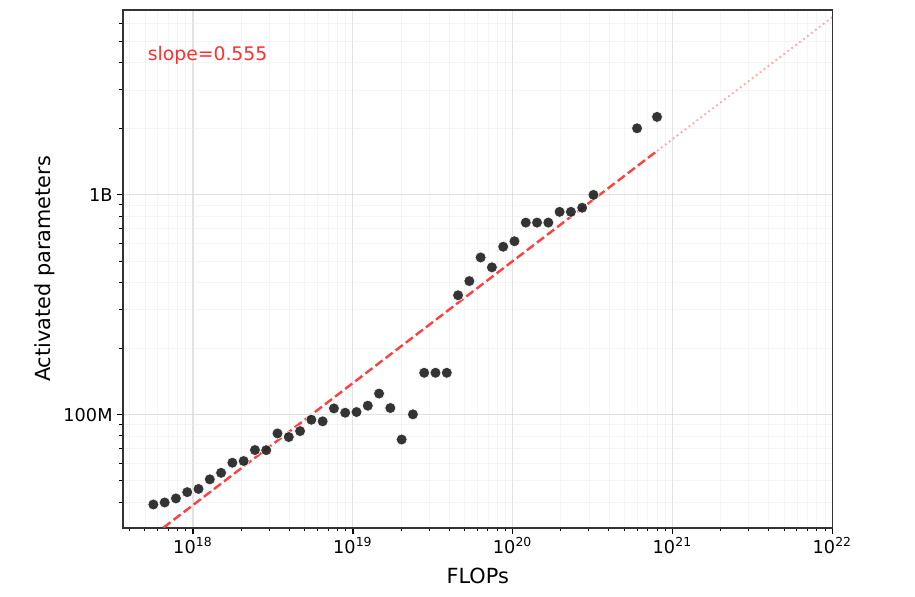}
        \caption{IsoFLOP-optimal model size.}
        \label{fig:iso_params_flops_180pvideo}
    \end{subfigure}
    \hfill
    \begin{subfigure}[t]{0.32\textwidth}
        \centering
        \includegraphics[width=\linewidth]{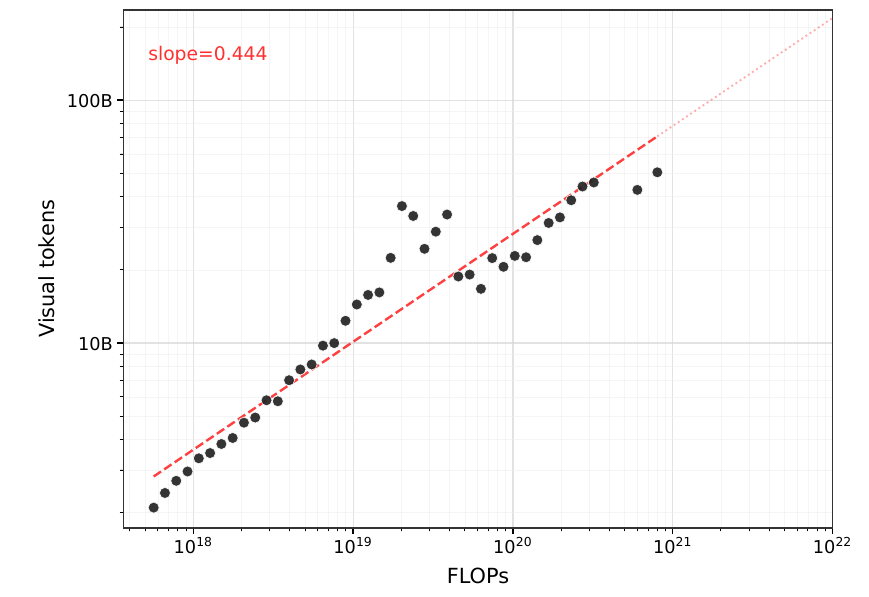}
        \caption{IsoFLOP-optimal training tokens.}
        \label{fig:iso_tokens_flops_180pvideo}
    \end{subfigure}
    \vspace{-0.05in}
    \caption{
    \textbf{Compute-optimal scaling of \ours{} for $180$p video pre-training.}
    \textbf{Top row, training-loss envelope:}
    (a)~Diffusion training loss versus cumulative training FLOPs; the lower envelope identifies the best checkpoint at each compute budget.
    (b)~Compute-optimal activated model size $N_{\mathrm{opt}}(C)$ from the envelope-winning checkpoints.
    (c)~The corresponding compute-optimal training tokens $D_{\mathrm{opt}}(C)$, obtained using each model's logged per-step FLOPs.
    \textbf{Bottom row, IsoFLOP profiles:}
    (d)~Diffusion loss versus activated model size at fixed compute budgets; quadratic-profile minima, or observed minima for boundary slices, identify the optimal model at each budget.
    (e)~Compute-optimal activated model size inferred from these minima.
    (f)~The corresponding compute-optimal training tokens.
    Dashed lines show the fitted power laws, with their exponents annotated in the scaling panels.
    }
    \vspace{-0.1in}
    \label{fig:chinchilla_180pvideo}
    \label{fig:iso_chinchilla_180pvideo}\
    \vspace{-1em}
\end{figure*}

\subsection{Compute-Optimal Scaling Laws}
\label{sec:exp_law}

We estimate the compute-optimal allocation between parameters and training tokens with all three of their estimators: the training-curve envelope, IsoFLOP profiles, and a direct parametric fit of the loss surface (Eq.~\ref{eq:chinchilla}). In all cases, we start by training the model family in Table~\ref{tab:model_family} with different numbers of tokens, and ensure all models are trained in the infinite-data regime. All fits use activated parameters for $N$, since this is the quantity that determines the per-token compute of our sparse MoE backbone, and use the diffusion training loss defined in Sec.~\ref{sec:model_arch_preliminary}. The warmup-then-constant base-learning-rate schedule from Sec.~\ref{sec:implementation_details} lets us treat intermediate checkpoints from a long run as comparable shorter-horizon runs.

Specifically, the envelope estimate follows the lower envelope of loss--FLOPs trajectories and records the checkpoint that attains the lowest loss at each compute budget. Although cumulative FLOPs grow with $D$ for a fixed model, the same FLOP budget corresponds to different visual-token counts across model sizes because the per-token cost varies with $N$. Each envelope-winning checkpoint therefore jointly determines $N_{\mathrm{opt}}(C)$ and $D_{\mathrm{opt}}(C)$. The latter is the optimal training volume associated with the selected model size, rather than an independent compute axis. We use the logged per-step FLOPs to map each checkpoint to $C$. The IsoFLOP estimate instead slices the trajectories at fixed FLOPs, fits a local quadratic profile of loss against $\log N$, and uses the profile minimum to infer $N_{\mathrm{opt}}(C)$ and $D_{\mathrm{opt}}(C)$. We then fit power laws to these optima. This mirrors the first two empirical estimators in \citet{hoffmann2022chinchilla}, while adapting them to visual diffusion loss and activated-parameter accounting.

\paragraph{\textbf{Scaling Laws on Image generation.}}
For $256^2$ image pre-training, the current fit uses $36$ usable runs covering $29$ activated model sizes and $1.2\times10^5$ logged loss points. The envelope spans $8.6\times10^{16}$ to $2.2\times10^{21}$ FLOPs, while the IsoFLOP profiles use interior budgets from $6.2\times10^{17}$ to $1.55\times10^{20}$ FLOPs. Fig.~\ref{fig:chinchilla_256img} (\subref{fig:loss_flops_256}-\subref{fig:tokens_flops_256}) shows that the training curves are cleanly ordered and that the lower envelope moves from smaller to larger models as the budget increases. The resulting frontier is close to balanced: the envelope fit gives
$N_{\mathrm{opt}}(C)\propto C^{0.505}$ and $D_{\mathrm{opt}}(C)\propto C^{0.484}$. The exponents sum to $0.989 \approx 1$, consistent with the approximate compute relation $C \propto N D$. The compute is divided nearly evenly between activated model size and visual-token count.

Fig.~\ref{fig:iso_chinchilla_256img} (\subref{fig:iso_loss_params_256}-\subref{fig:iso_tokens_flops_180pvideo}) gives an independent view of the same law. Each fixed-compute slice forms a smooth, approximately convex profile in model size, with a minimum that shifts rightward as compute increases. Fitting these minima yields
$N_{\mathrm{opt}}(C)\propto C^{0.481}$ and $D_{\mathrm{opt}}(C)\propto C^{0.511}$.
The agreement between the envelope and IsoFLOP estimates indicates that additional image-training compute should be allocated nearly evenly between activated parameters and visual tokens, rather than spent mostly on either larger models or longer training alone.

\paragraph{\textbf{Scaling Laws on Video generation.}}
We repeat the study on $180$p videos using $30$ independent training-from-scratch runs on the same $29$-model activated-parameter grid. Video curves cover $1.8\times10^{17}$ to $9.5\times10^{20}$ FLOPs on the envelope, and the IsoFLOP analysis uses budgets from $5.6\times10^{17}$ to $3.2\times10^{20}$ FLOPs, with the high-compute frontier anchored by the $6\times10^{20}$ and $8\times10^{20}$ FLOP slices.

The video frontier is again close to balanced, but the largest budgets now available tilt mildly toward larger activated models. The full envelope fit in Fig.~\ref{fig:chinchilla_180pvideo} (\subref{fig:loss_flops_180pvideo}-\subref{fig:tokens_flops_180pvideo}) gives
$N_{\mathrm{opt}}(C)\propto C^{0.564}$ and $D_{\mathrm{opt}}(C)\propto C^{0.431}$. The IsoFLOP profiles in Fig.~\ref{fig:iso_chinchilla_180pvideo} (\subref{fig:iso_loss_params_180pvideo}-\subref{fig:iso_tokens_flops_180pvideo}) give
$N_{\mathrm{opt}}(C)\propto C^{0.555}$ and $D_{\mathrm{opt}}(C)\propto C^{0.444}$ when the high-compute slices are included.

Compared with image generation, video generation is consistently more model-heavy across both estimators: the envelope/IsoFLOP model-size exponents increase from $0.505/0.481$ for images to $0.564/0.555$ for videos, while the corresponding token exponents decrease from $0.484/0.511$ to $0.431/0.444$. Thus, as the compute budget grows, compute-optimal video training allocates a modestly larger share of additional compute to model capacity rather than training tokens. One plausible explanation for this shift is that video generation places a broader representational burden on the backbone. In addition to the appearance and semantic structure required for images, a video model must capture frame correspondence, including object dynamics, temporal correspondences, and camera motion.  This may reduce the marginal return from scaling $D$ relative to scaling $N$, consistent with the observed shift of the video frontier.

\paragraph{\textbf{Parametric loss fit.}}
The envelope and IsoFLOP estimators use only the per-budget optima. As a third, independent route, we fit the full loss surface of Eq.~\ref{eq:chinchilla} to all measurements at once, following the third approach of \citet{hoffmann2022chinchilla}: the five parameters $\{E, A, B, a, b\}$ are obtained by minimizing a Huber loss ($\delta=10^{-3}$) on the residuals between predicted and observed log-losses, using L-BFGS-B from a grid of $4{,}500$ initializations (doubled for video with a second grid centered on the observed loss scale). For images, the fit uses the de-spiked training trajectories of the same $36$ runs, sampling $24$ geometrically spaced checkpoints per model ($696$ points spanning $3.2\times10^{17}$ to $2.2\times10^{21}$ FLOPs); for video, it uses the final checkpoint of each of the $30$ runs. The fitted surfaces are
\begin{equation}
    \hat{L}_{\mathrm{image}}(N, D) = 0.126 + 5.28\,N^{-0.315} + 33.8\,D^{-0.336},
    \label{eq:parametric_img}
\end{equation}
\begin{equation}
    \hat{L}_{\mathrm{video}}(N, D) = 0.124 + 8.07\,N^{-0.330} + 145.2\,D^{-0.394}.
    \label{eq:parametric_vid}
\end{equation}
Both fits are accurate: the image surface reproduces the observed log-losses with $R^{2}=0.993$ and a median absolute relative error of $0.19\%$ ($0.22\%$ under leave-one-model-out validation), and the video surface predicts the held-out terminal losses with a median error of $0.49\%$. 

Minimizing the fitted surfaces under the constraint $C \approx 6ND$ yields
$N_{\mathrm{opt}}(C)\propto C^{0.516}$, $D_{\mathrm{opt}}(C)\propto C^{0.484}$ for images and
$N_{\mathrm{opt}}(C)\propto C^{0.544}$, $D_{\mathrm{opt}}(C)\propto C^{0.456}$ for video.
These closed-form exponents agree closely with the two non-parametric estimators above: all three routes place the image frontier within $[0.48, 0.52]$ and the video frontier within $[0.53, 0.56]$ for the model-size exponent. A run-level bootstrap ($100$ resamples over $80\%$ of the model settings) gives an $80\%$ interval of $[0.503, 0.546]$ for the image exponent; the corresponding video interval is wider ($[0.47, 0.80]$), since the terminal checkpoints span only $0.7$ decades in $D$, so we read the video fit as confirming the direction of the allocation rather than sharpening its exponent. Notably, the fitted irreducible losses are nearly identical across the two modalities ($E=0.126$ versus $0.124$). 
The near equality of $E$ is consistent with using the same models, including the denoiser and VAE, for both modalities, as architecture changes tend primarily to shift the offset of a scaling curve~\citep{hestness2017deep}.
Overall, the three estimators consistently indicate that image pre-training compute should be split almost evenly between activated parameters and visual tokens, while video pre-training currently favors allocating a modestly larger share to model size.

\subsection{Optimal Image-Video Data Ratio}
\label{sec:exp_ratio}

We extend the compute-optimal analysis along the mixture axis introduced in Sec.~\ref{sec:chinchilla} and estimate how the optimal image--video data ratio evolves with the training budget.

\begin{figure*}[!t]
    \centering
    \includegraphics[width=\textwidth]{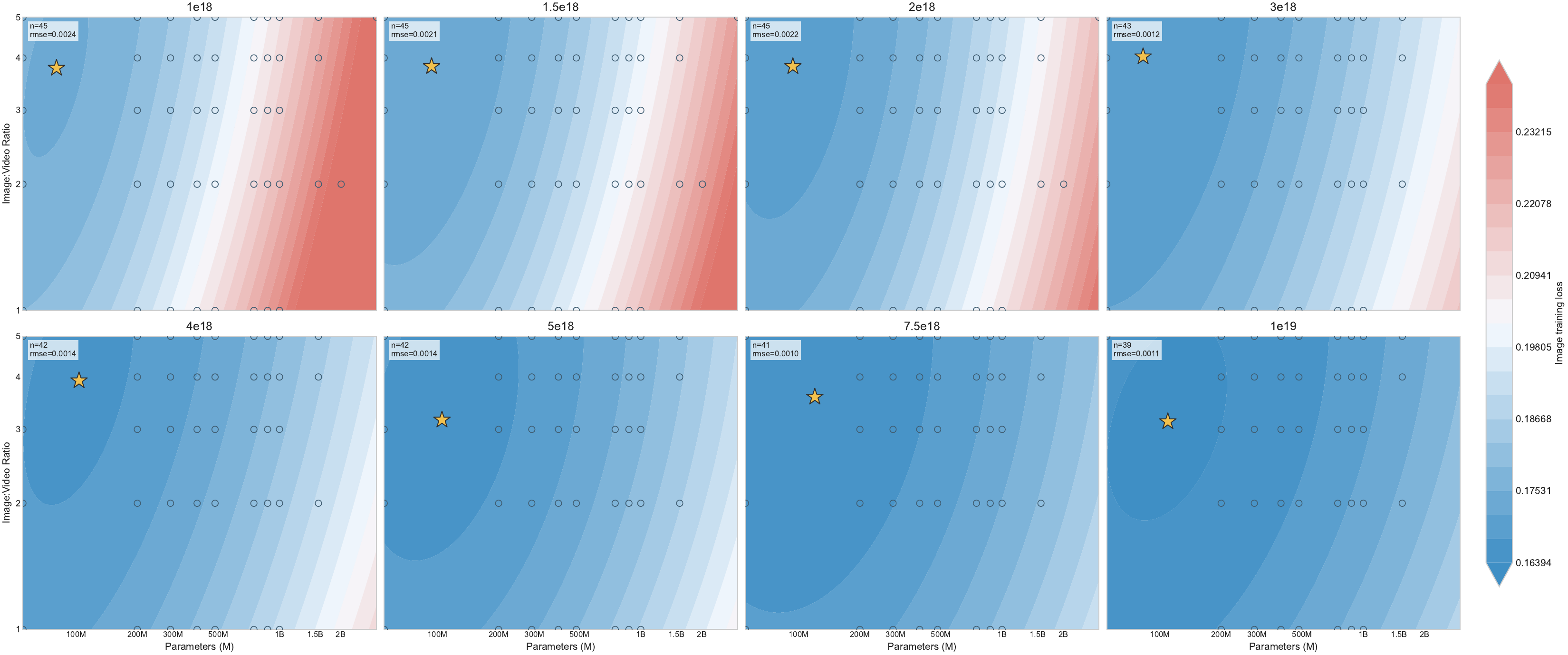}
    \caption{\textbf{IsoFLOP surfaces of the image diffusion loss over model size and image:video data sample ratio.} For each of eight compute budgets from $10^{18}$ to $10^{19}$ FLOPs, we fit a quadratic surface to the losses of runs spanning both model sizes and mixture ratios. Each circle marks an experimental data point, and the star marks the fitted optimum under that budget.
    }
    \vspace{-0.1in}\label{fig:ivratio_loss_surface}
\end{figure*}


\paragraph{\textbf{Setup.}}
Unlike the single-modality studies above, these runs train on image and video data jointly. We train models spanning roughly $50$M to $2$B activated parameters under several image:video mixture ratios between $1{:}1$ and $5{:}1$, following the recipe of Sec.~\ref{sec:implementation_details}, and record the image and video diffusion losses separately. Note that this ratio searching range is decided by the actual amount of image and video data, considering two facts: i) Video data is usually less than image data; ii) The selected ratio should be ideally maintain the infinite-data regime (less video data reuse). 
In detail, we fix the total batch size during training and balance the number of image and video samples. We follow the previous $256^2$ and $180$p settings for image and video data, respectively.
The analysis covers eight compute budgets that span $10^{18}$ to $10^{19}$ FLOPs. Slicing all training trajectories in a given budget yields $39$--$45$ measurements per budget, each corresponding to one model-size and ratio configuration. Following the extended IsoFLOP methodology of Sec.~\ref{sec:chinchilla}, we fit a quadratic surface of each loss over model size and mixture ratio at every budget, and take the surface optimum as the compute-optimal configuration for that budget.

\paragraph{\textbf{IsoFLOP loss surfaces.}}
Fig.~\ref{fig:ivratio_loss_surface} shows the fitted image-loss surfaces. The quadratic surfaces reproduce the measurements closely at all budgets, with root-mean-square residuals of at most $2.4\times10^{-3}$, about $1\%$ of the measured loss. The surfaces are smooth and approximately convex, and their optima (stars) lie in the interior of the ratio range. We notice two consistent patterns. First, the optimal model size shifts toward larger models as the budget grows, in line with the compute-optimal frontier of Sec.~\ref{sec:exp_law}. Second, near the optimum the loss varies far more strongly along the model-size axis than along the ratio axis. Choosing the model size correctly therefore matters most, while moderately off-optimal mixtures cost little.

\begin{wrapfigure}{r}{0.4\textwidth}
    \centering
    \vspace{-0.1in}
    \includegraphics[width=0.4\textwidth]{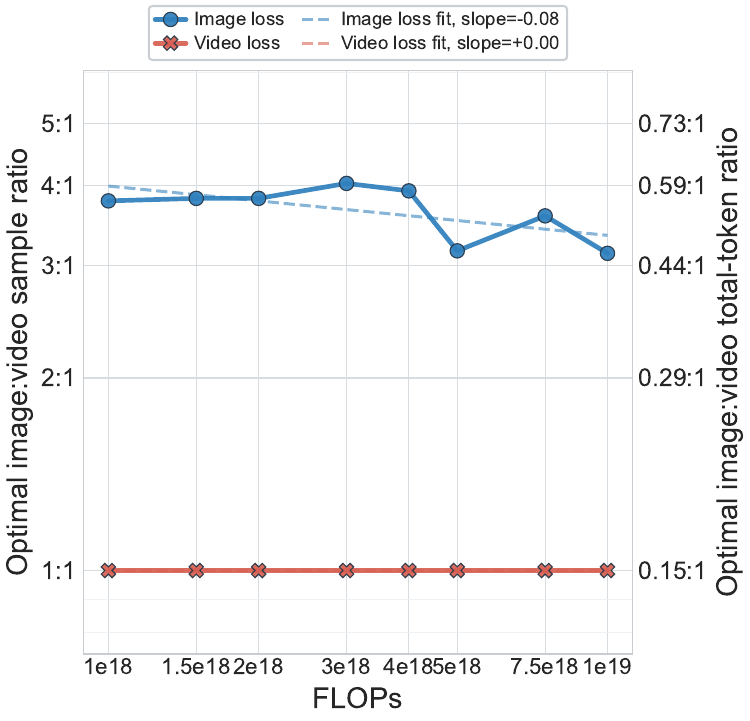}
    \vspace{-1.7em}
    \caption{\textbf{Compute-optimal image:video ratio versus training compute.}  The image-loss optimum drifts slowly from about $4{:}1$ toward $3{:}1$ (exponent $-0.08$), while the video-loss optimum stays at $1{:}1$, the most video-heavy mixture in the fitted range.
    }
    \label{fig:optimal_iv_ratio_fit}
    \vspace{-1em}
\end{wrapfigure}

\paragraph{\textbf{Compute-optimal data ratio.}}
Fig.~\ref{fig:optimal_iv_ratio_fit} tracks the optimal image-to-video ratio at each compute budget for both image and video losses, together with power-law fits. The two losses lead to clearly different optima. For image loss, the optimum is consistently image-heavy across all examined budgets, lying between $3{:}1$ and $4{:}1$. It shifts only gradually toward video as compute increases, moving from roughly $4{:}1$ at $10^{18}$ FLOPs to roughly $3{:}1$ at $10^{19}$ FLOPs. The fitted exponent is $-0.08$, meaning that the optimal ratio decreases by about $17\%$ per decade of compute. For video loss, the fitted optimum remains at $1{:}1$, the video-heavy boundary of our search range, for all budgets, with a fitted exponent of $0.00$. Because this optimum lies on the boundary, we interpret it as a one-sided result: within the mixtures explored here, video loss improves monotonically as the video share increases.

These results suggest a simple principle for data mixing: the compute-optimal mixture is image-heavy, but shifts slowly toward video as the training budget grows. This behavior is intuitive given the different roles and costs of image and video data. Images provide appearance and semantic supervision at a much lower token cost per sample, making image-heavy mixtures more compute-efficient in the low-compute regime. As compute increases, the optimal mixture allocates a modestly larger share to video, suggesting that additional budget can be used to learn temporally rich and long-tail dynamics after the model has acquired sufficient appearance priors from images. The small negative slope further indicates that this transition is gradual rather than abrupt: video becomes increasingly valuable with scale, but does not eliminate the token-efficiency advantage of image data.

We also convert the sample ratio into a total-token ratio, including text tokens. Under this view, the same optima correspond to a much more video-heavy allocation in terms of training tokens, reflecting the substantially higher token cost of each video sample. Together with the size--data frontier in Sec.~\ref{sec:exp_law}, these fits specify the compute-optimal model size, token budget, and data mixture for our largest runs.

\subsection{Generation Quality and Comparisons}
\label{sec:exp_quality}

\paragraph{\textbf{Loss comparisons under matched compute.}}

\begin{wrapfigure}{r}{0.42\textwidth}
    \centering
    \vspace{-0.2in}
    \includegraphics[width=0.42\textwidth]{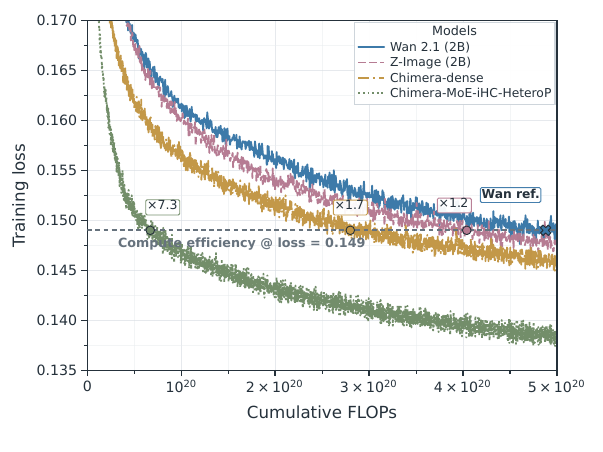}
    \vspace{-2.5em}
    \caption{
    \textbf{Compute efficiency under matched training conditions.}
    We calculate the compute efficiency across 2B instantiations of Wan2.1, Z-Image, Chimera-dense, and the complete Chimera configuration trained with the deterministic dataloader and a compute budget of $5\times10^{20}$ FLOPs.
    }
    \label{fig:flops_loss_compute_efficiency}
    \vspace{-1.2em}
\end{wrapfigure}

We construct 2B instantiations of the Wan2.1~\citep{wan2025wan} and Z-Image~\citep{cai2025zimage} architectures and train them alongside two 2B \ours{} variants on the same data, with the same diffusion objective, and to a shared compute budget of $5\times10^{20}$ FLOPs. The first \ours{} variant, Chimera-dense, uses dense FFNs and standard parameterization, providing a matched comparison without sparse capacity or \heterop{}. The complete variant adds MoE, iHC, and \heterop{}.

Fig.~\ref{fig:flops_loss_compute_efficiency} shows a clear convergence advantage for \ours{}. At the common training loss of $0.149$, Wan requires $4.29\times10^{20}$ FLOPs and Z-Image requires $3.75\times10^{20}$ FLOPs. Chimera-dense reaches the same loss with $2.55\times10^{20}$ FLOPs, corresponding to $1.7\times$ higher compute efficiency than Wan despite using neither MoE nor \heterop{}. The complete \ours{} configuration reaches the target with only $6.27\times10^{19}$ FLOPs, a $6.8\times$ compute-efficiency gain over Wan. Its curve remains below all dense baselines throughout the shared compute range. These results separate two sources of improvement: the dense \ours{} architecture already converges more efficiently than matched full-attention baselines, while sparse capacity, iHC, and \heterop{} compound this advantage.

\paragraph{\textbf{Qualitative results.}}

\begin{figure*}[t]
    \vspace{-4em}
    \centering
    \includegraphics[width=0.31\textwidth]{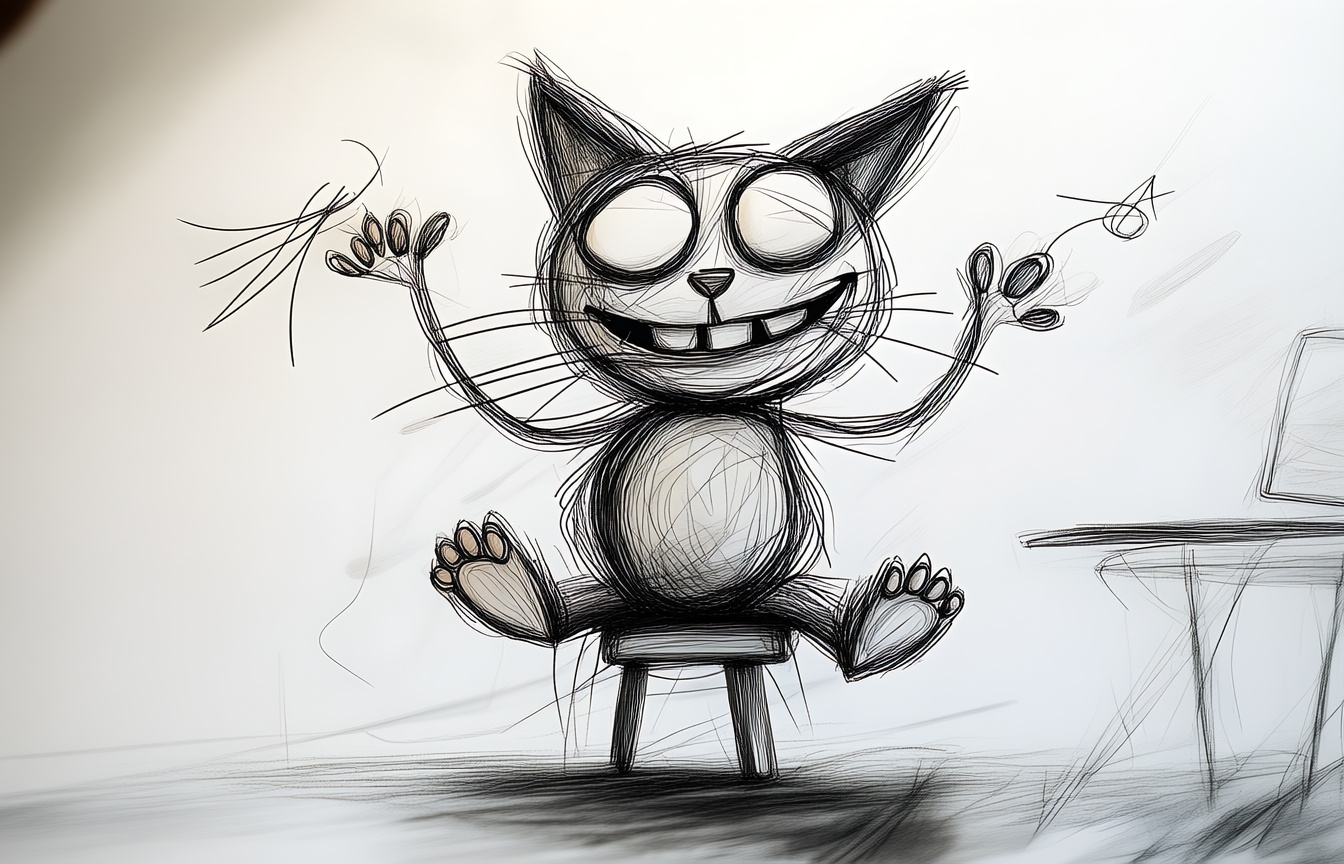}\hfill
    \includegraphics[width=0.31\textwidth]{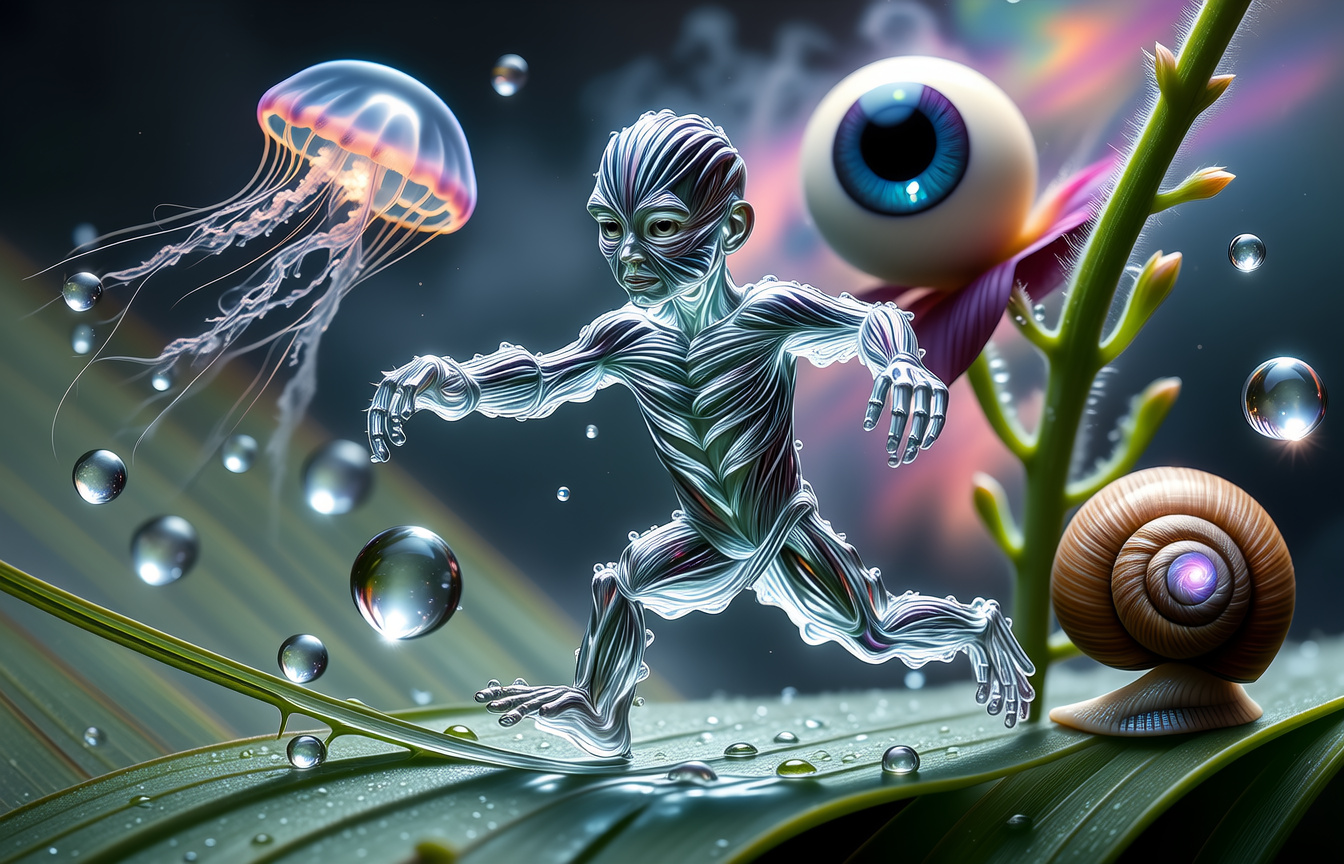}\hfill
    \includegraphics[width=0.31\textwidth]{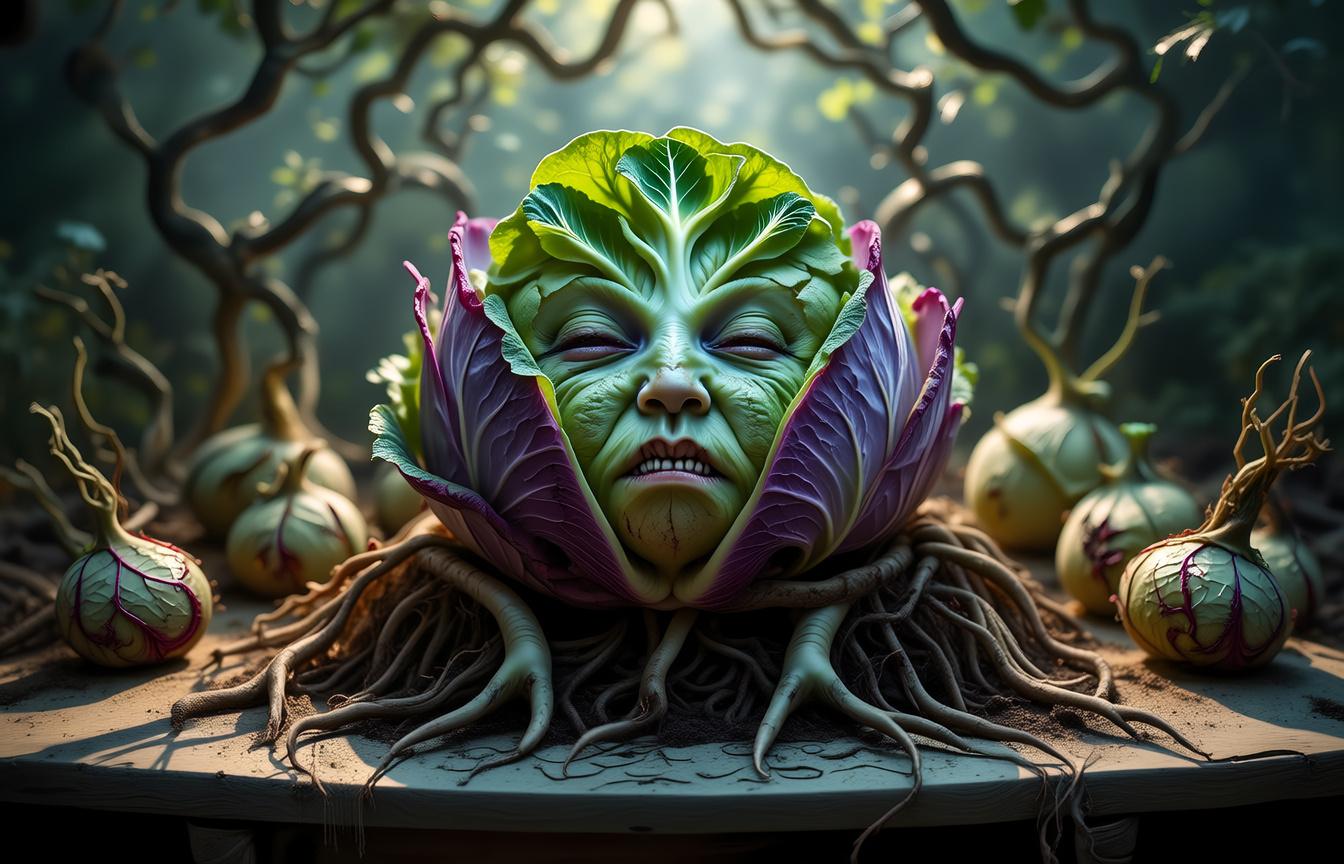}\\[1pt]
    \includegraphics[width=0.31\textwidth]{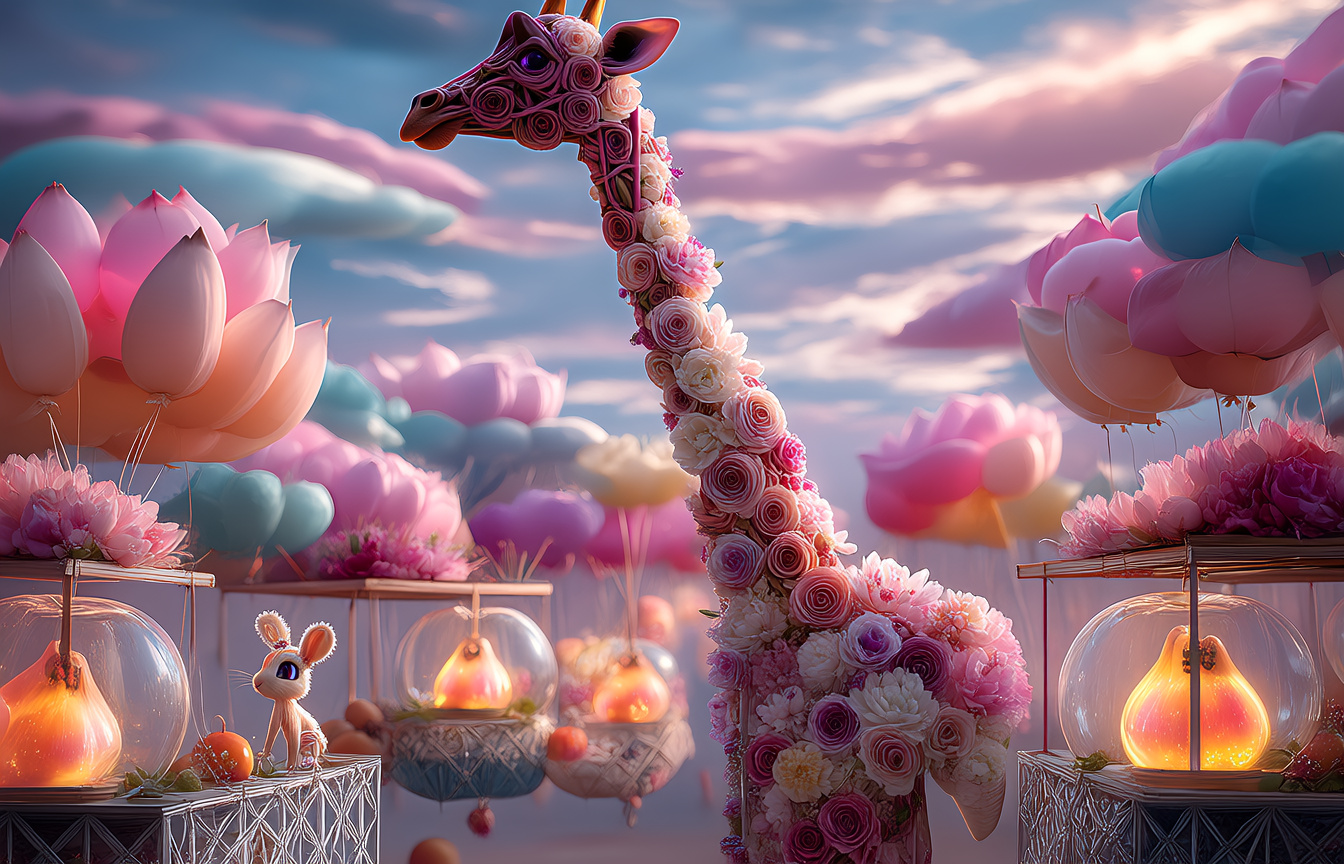}\hfill
    \includegraphics[width=0.31\textwidth]{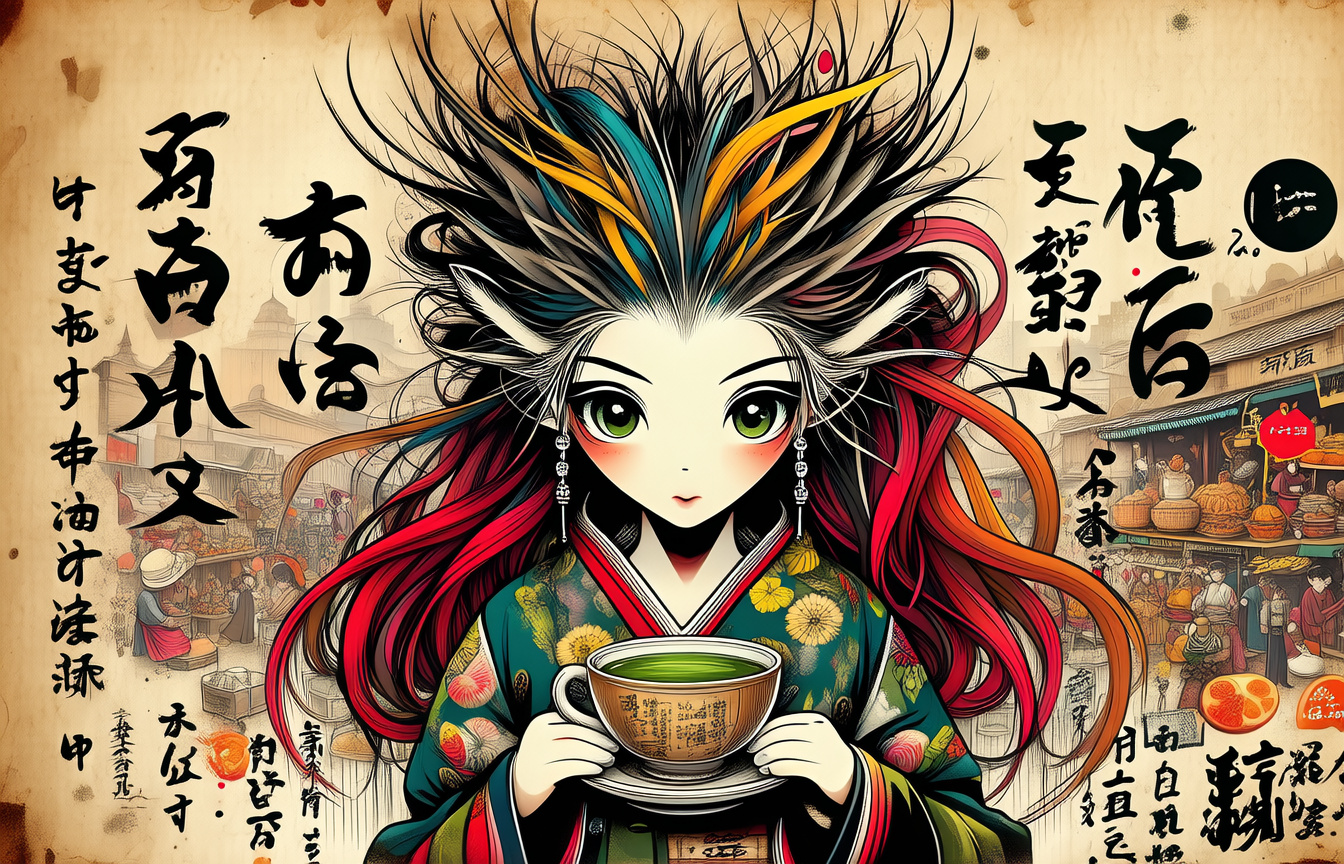}\hfill
    \includegraphics[width=0.31\textwidth]{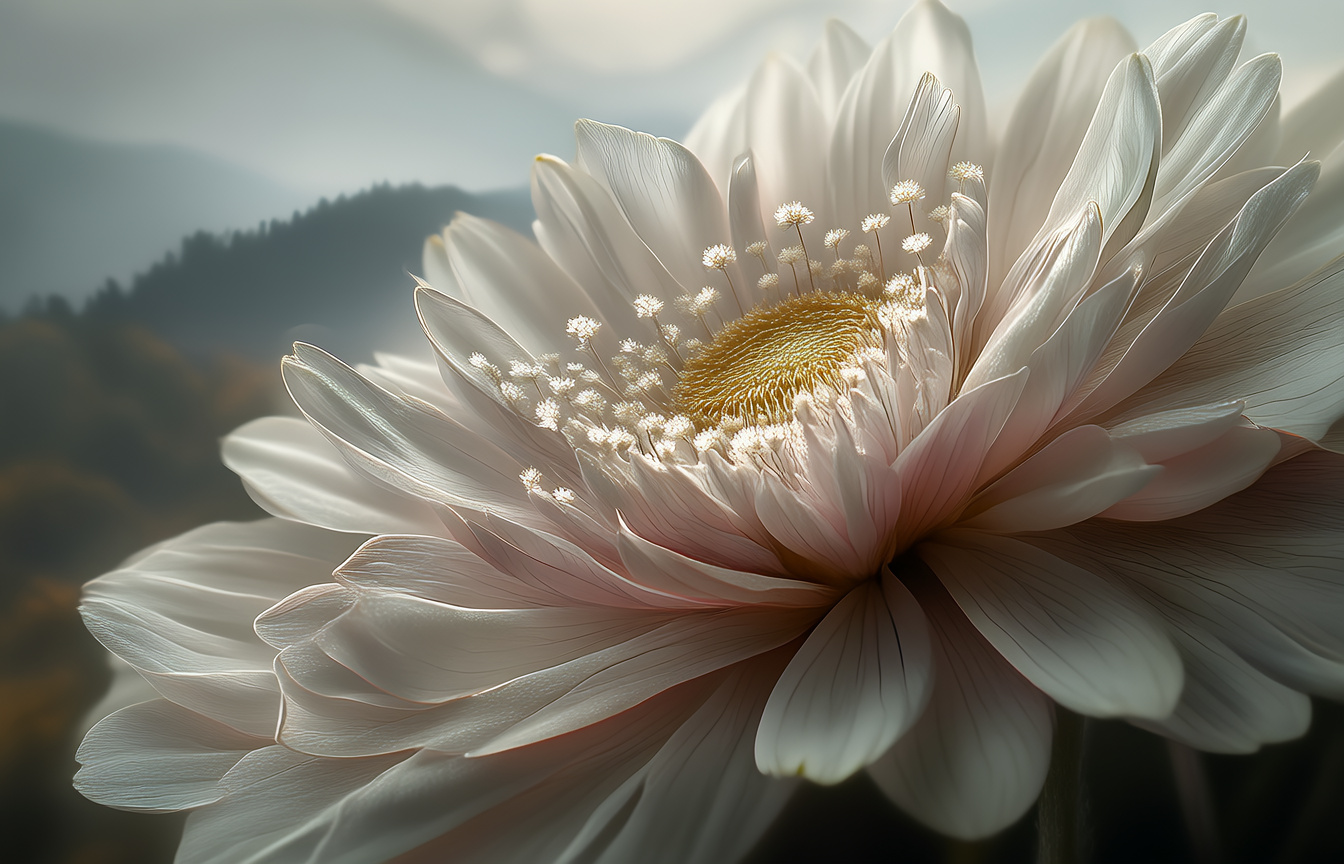}\\[1pt]
    \includegraphics[width=0.31\textwidth]{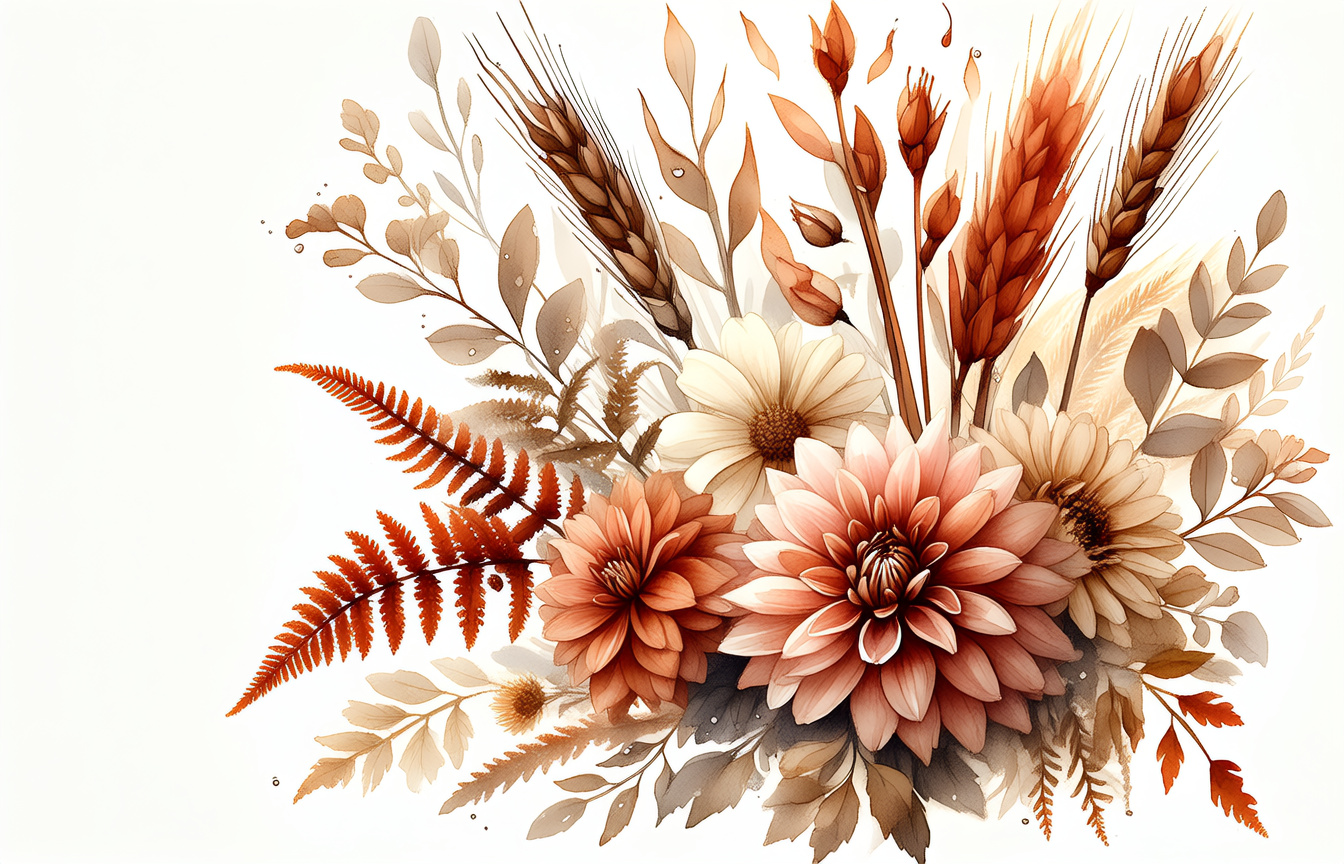}\hfill
    \includegraphics[width=0.31\textwidth]{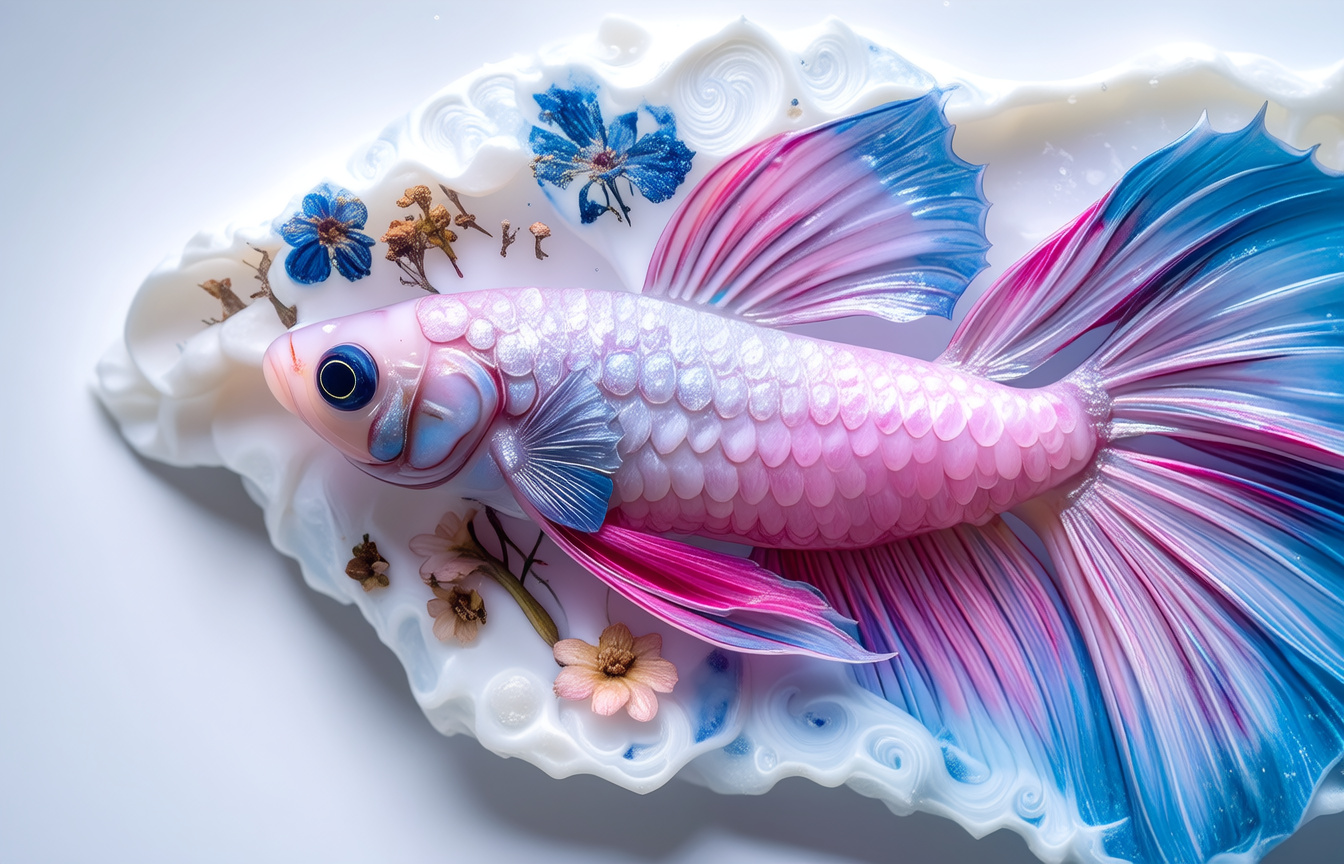}\hfill
    \includegraphics[width=0.31\textwidth]{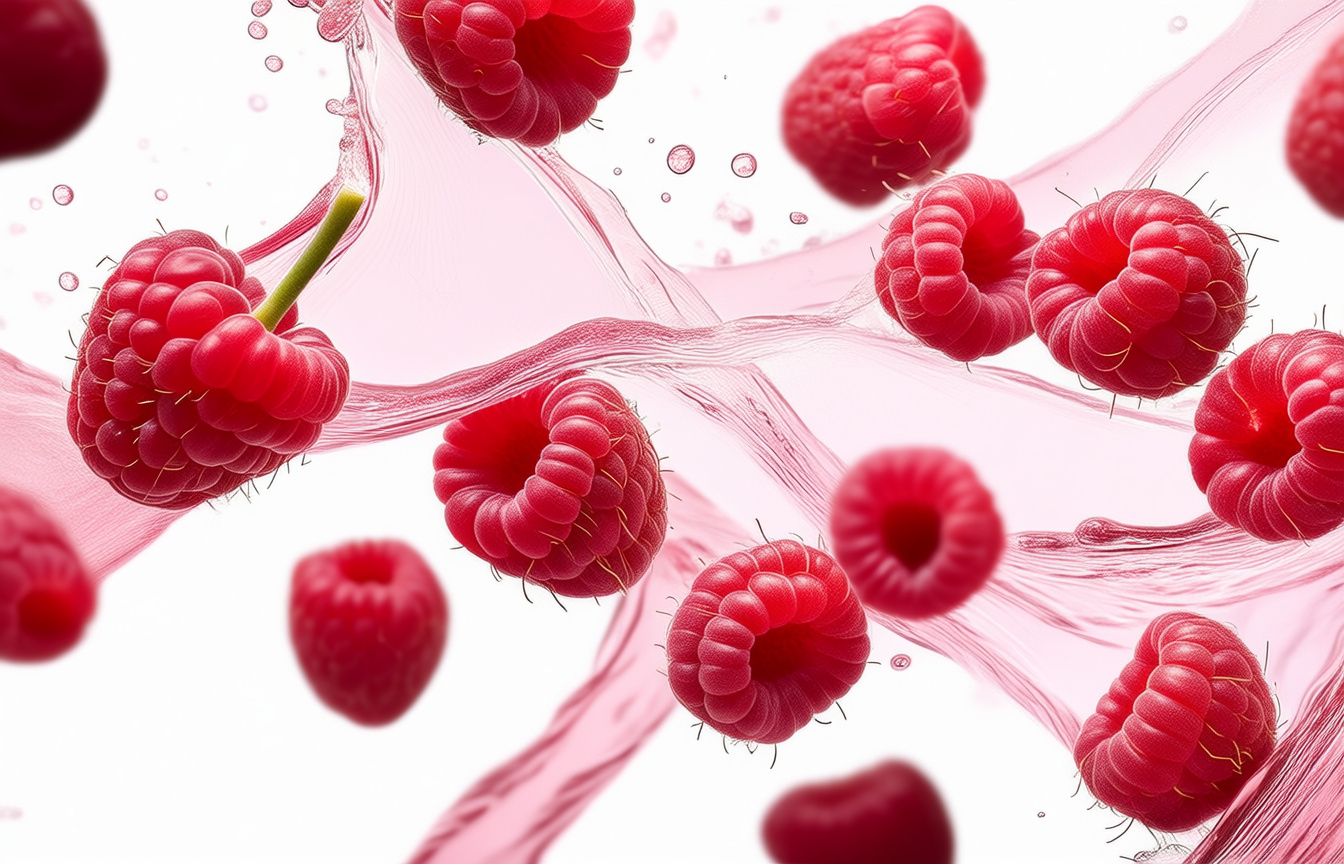}\\[1pt]
    \includegraphics[width=0.31\textwidth]{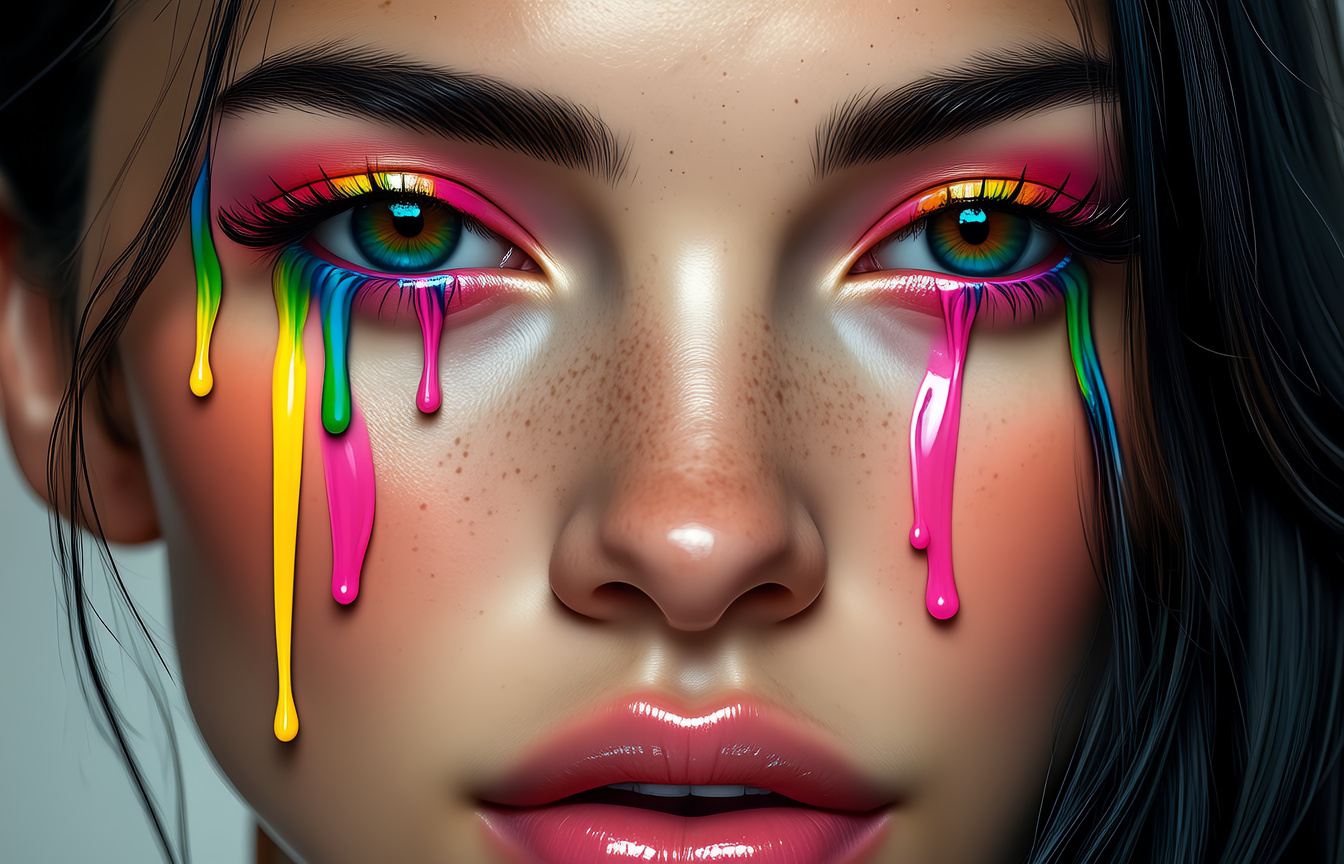}\hfill
    \includegraphics[width=0.31\textwidth]{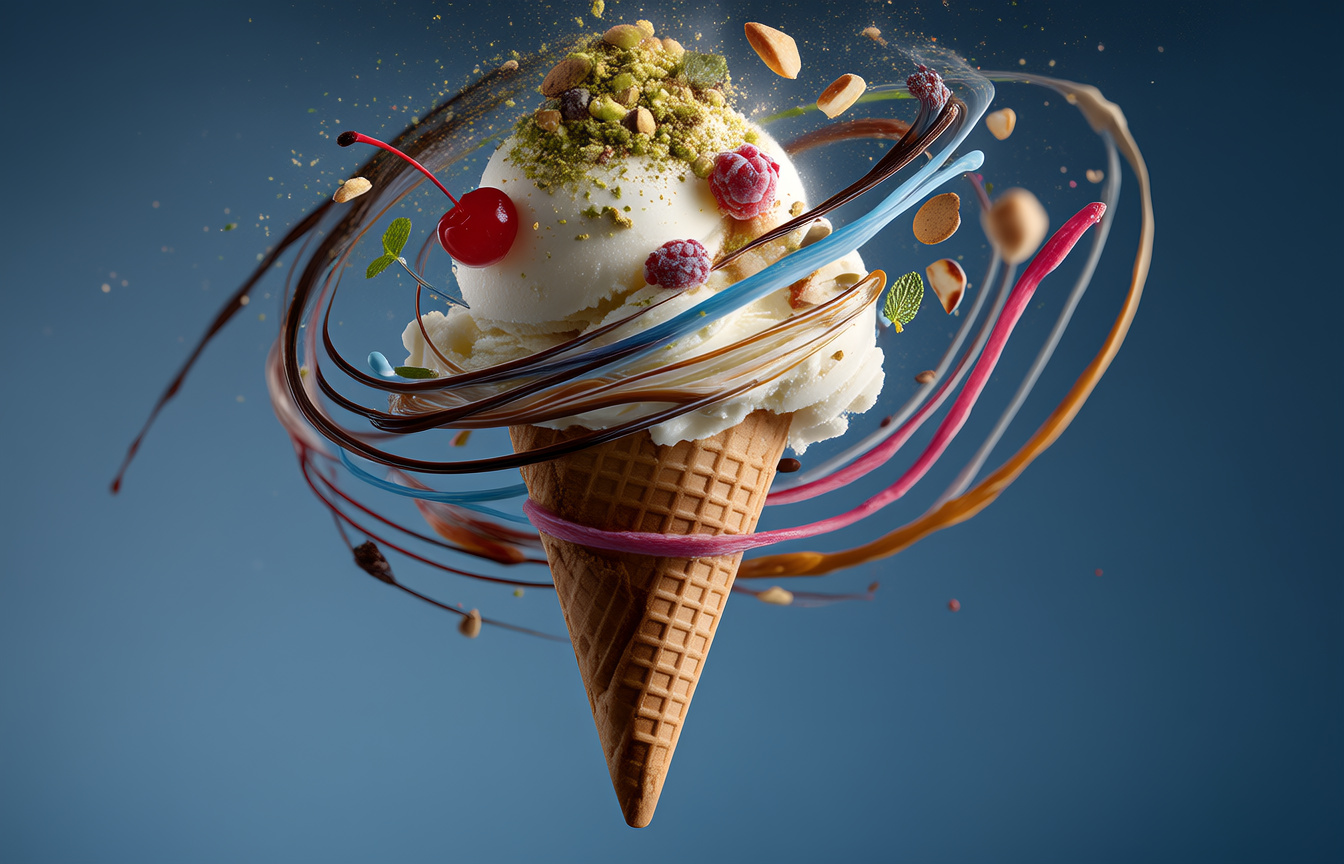}\hfill
    \includegraphics[width=0.31\textwidth]{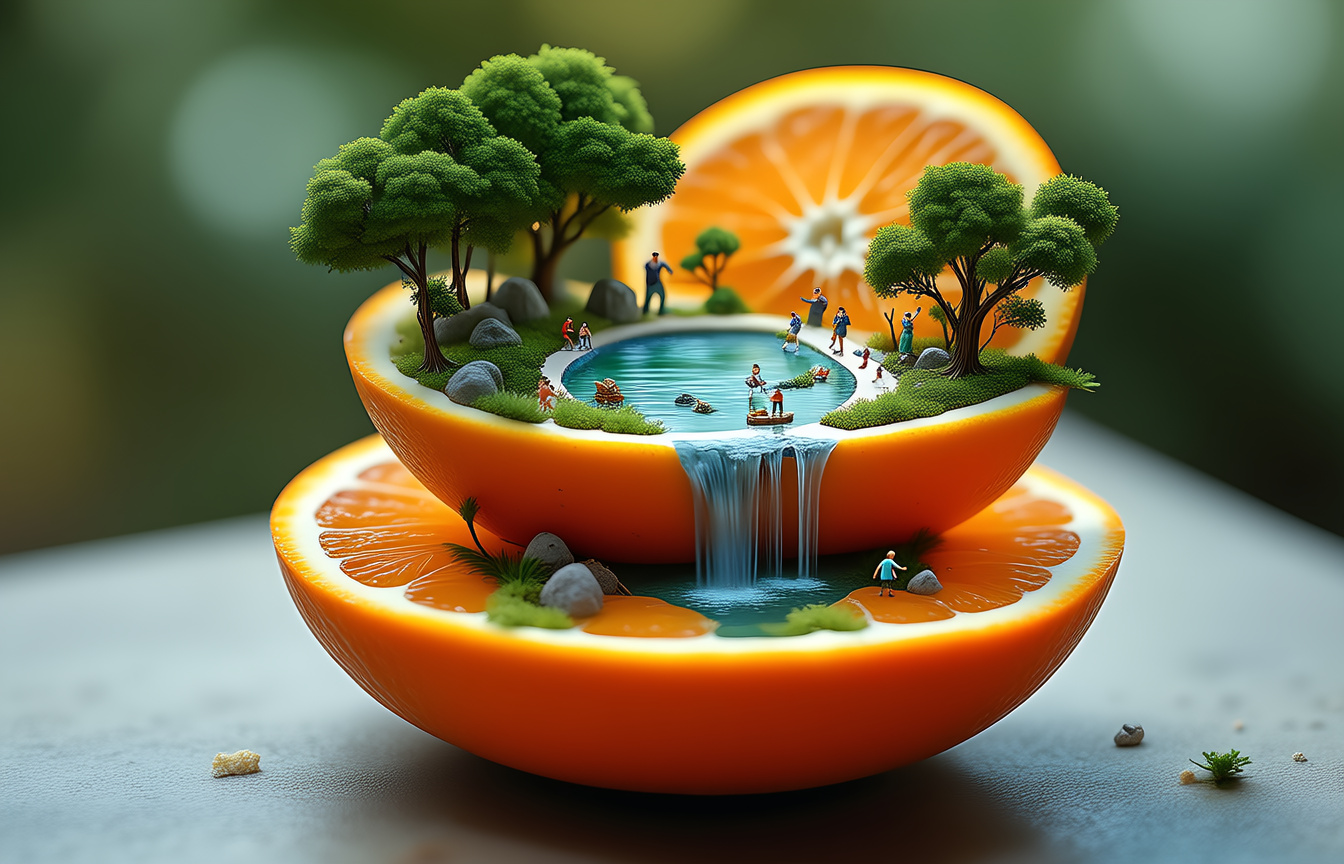}\\[1pt]
    \includegraphics[width=0.31\textwidth]{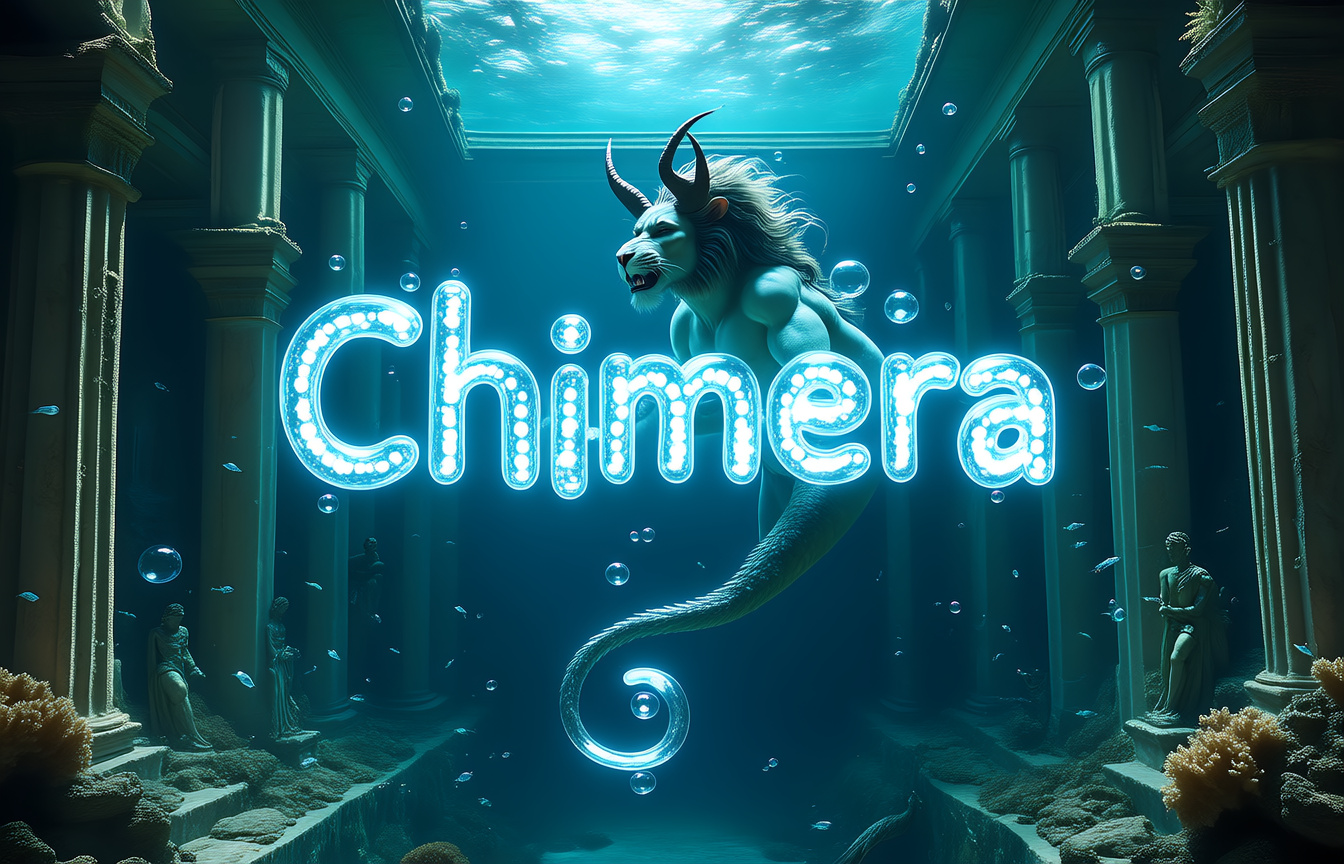}\hfill
    \includegraphics[width=0.31\textwidth]{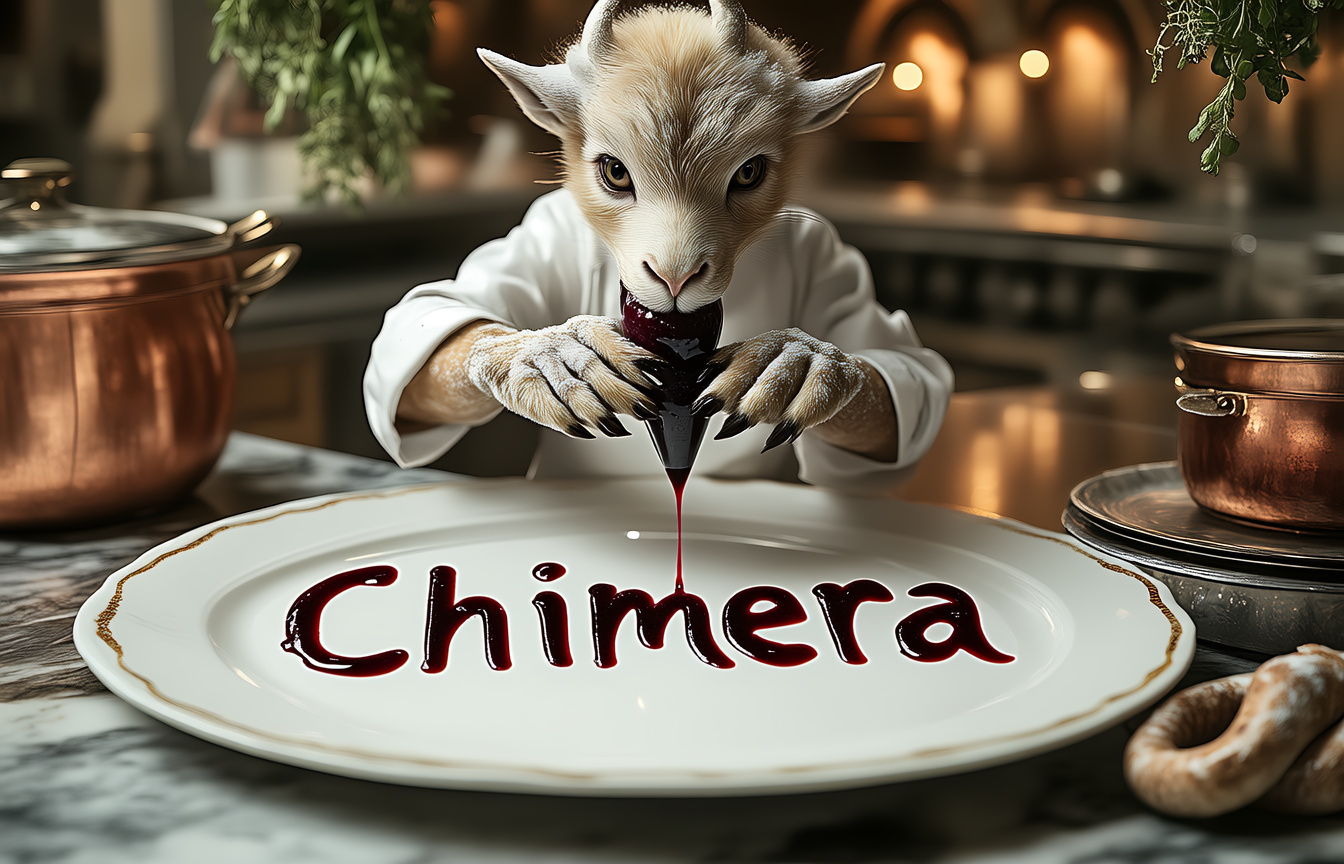}\hfill
    \includegraphics[width=0.31\textwidth]{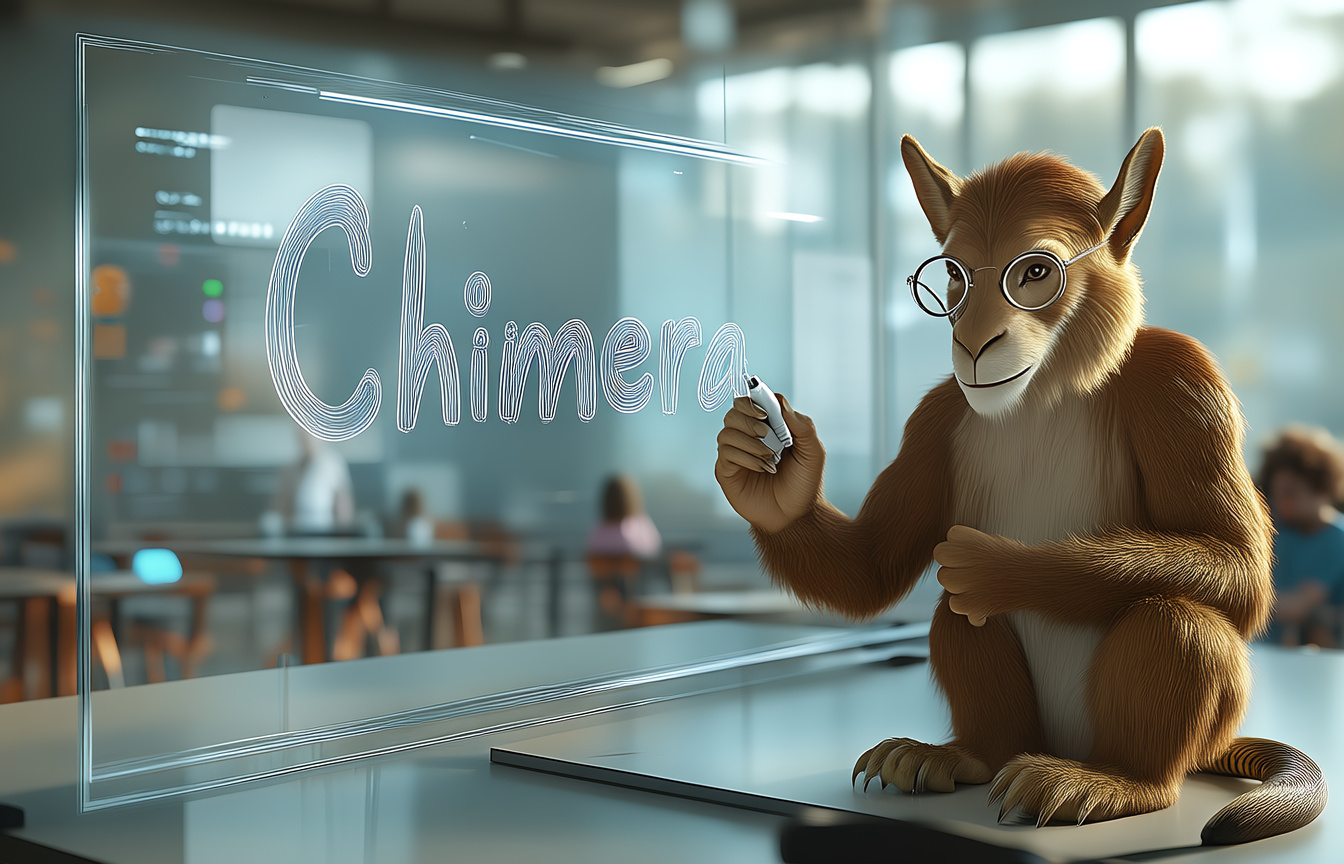}
    \caption{\textbf{Qualitative text-to-image samples from \ours{}.}
    All images are generated at $1344\times864$ resolution with classifier-free guidance scale $4$, using prompts that range from short phrases to long, detailed descriptions.
    }
    \label{fig:gen_samples}
    \vspace{-1em}
\end{figure*}

\begin{figure*}[t]
    \centering
    \begin{minipage}[c]{0.662\textwidth}
        \includegraphics[width=\linewidth]{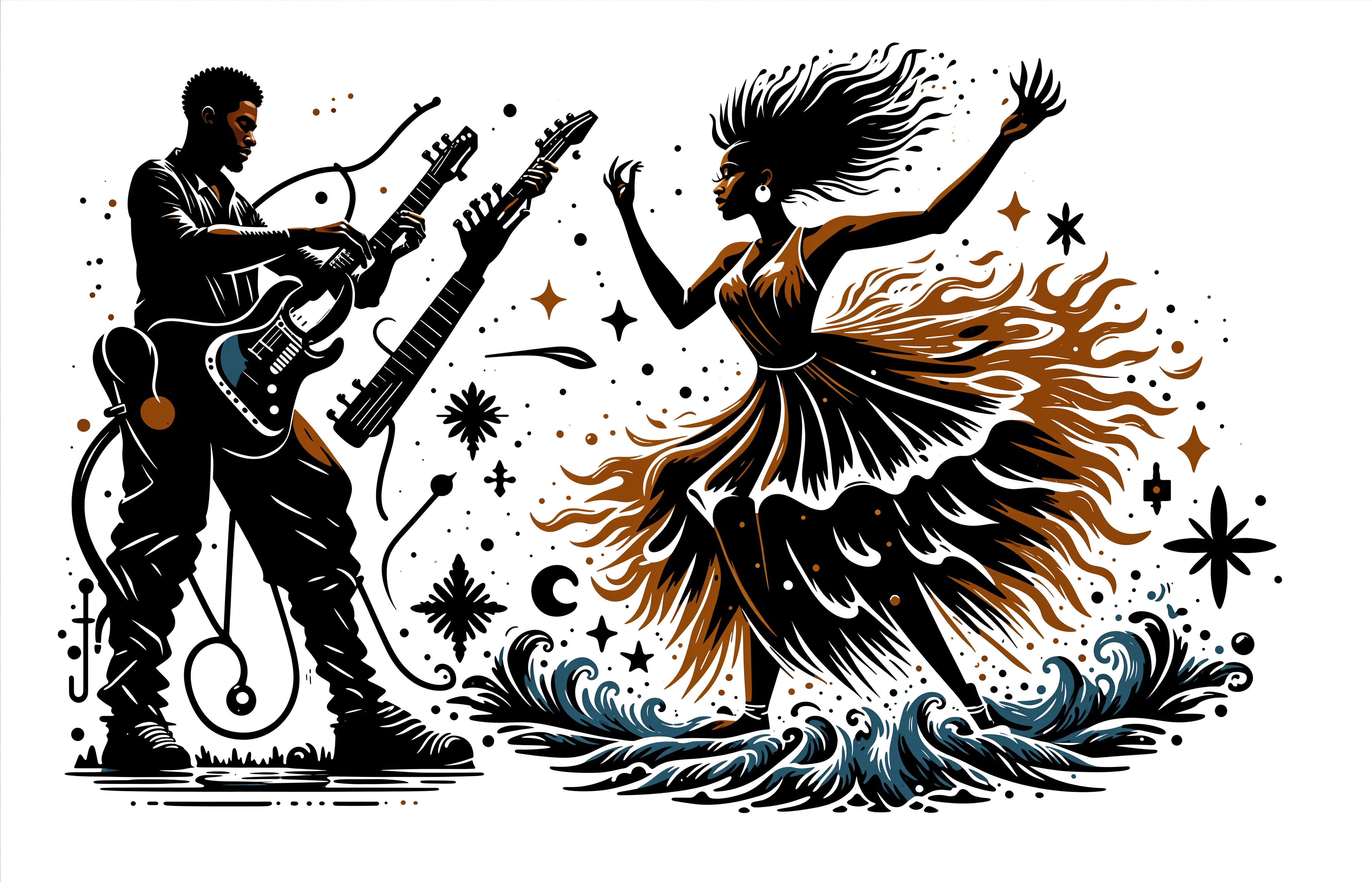}
    \end{minipage}\hfill%
    \begin{minipage}[c]{0.331\textwidth}
        \includegraphics[width=\linewidth]{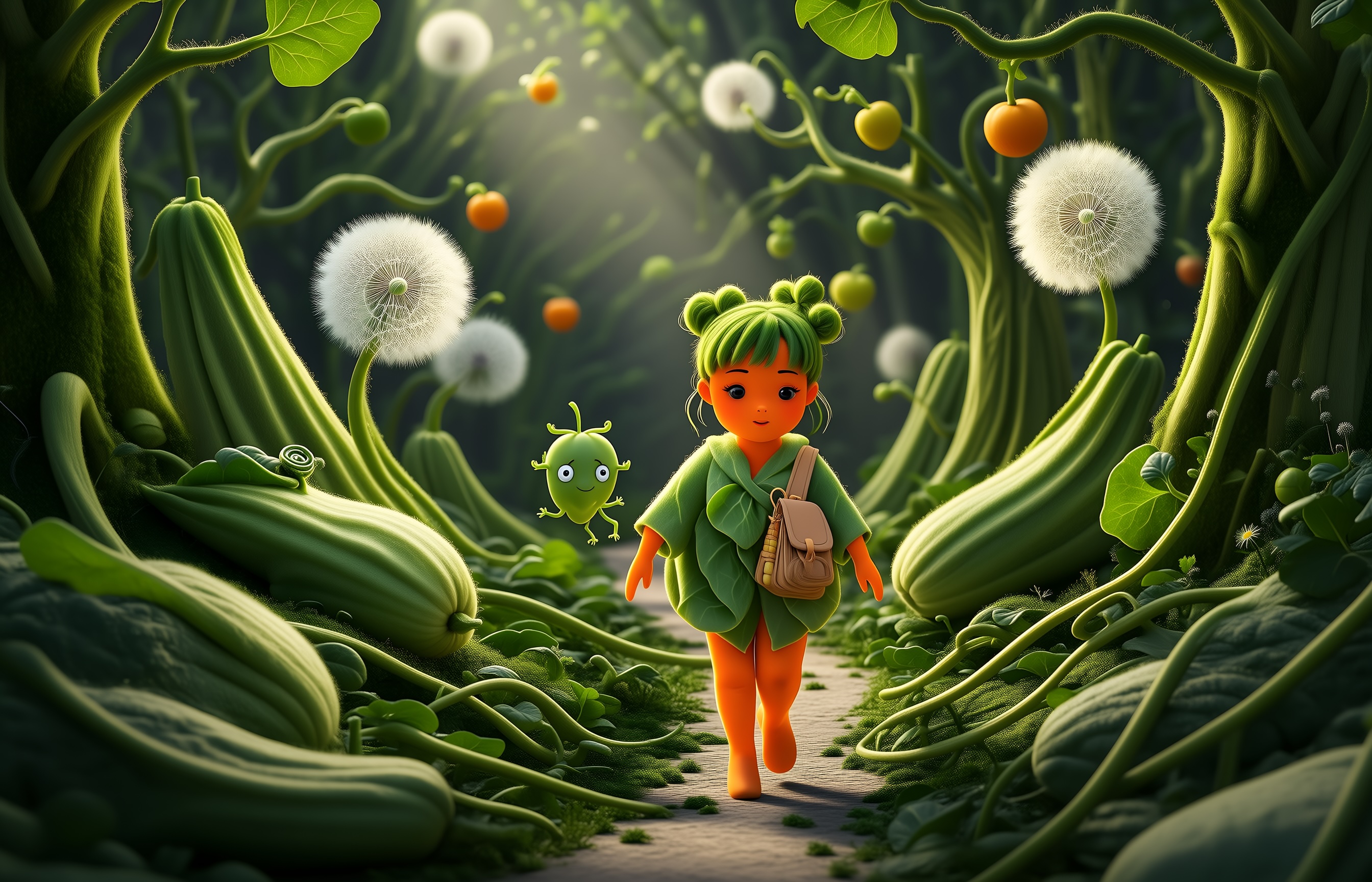}\par\nointerlineskip
        \includegraphics[width=\linewidth]{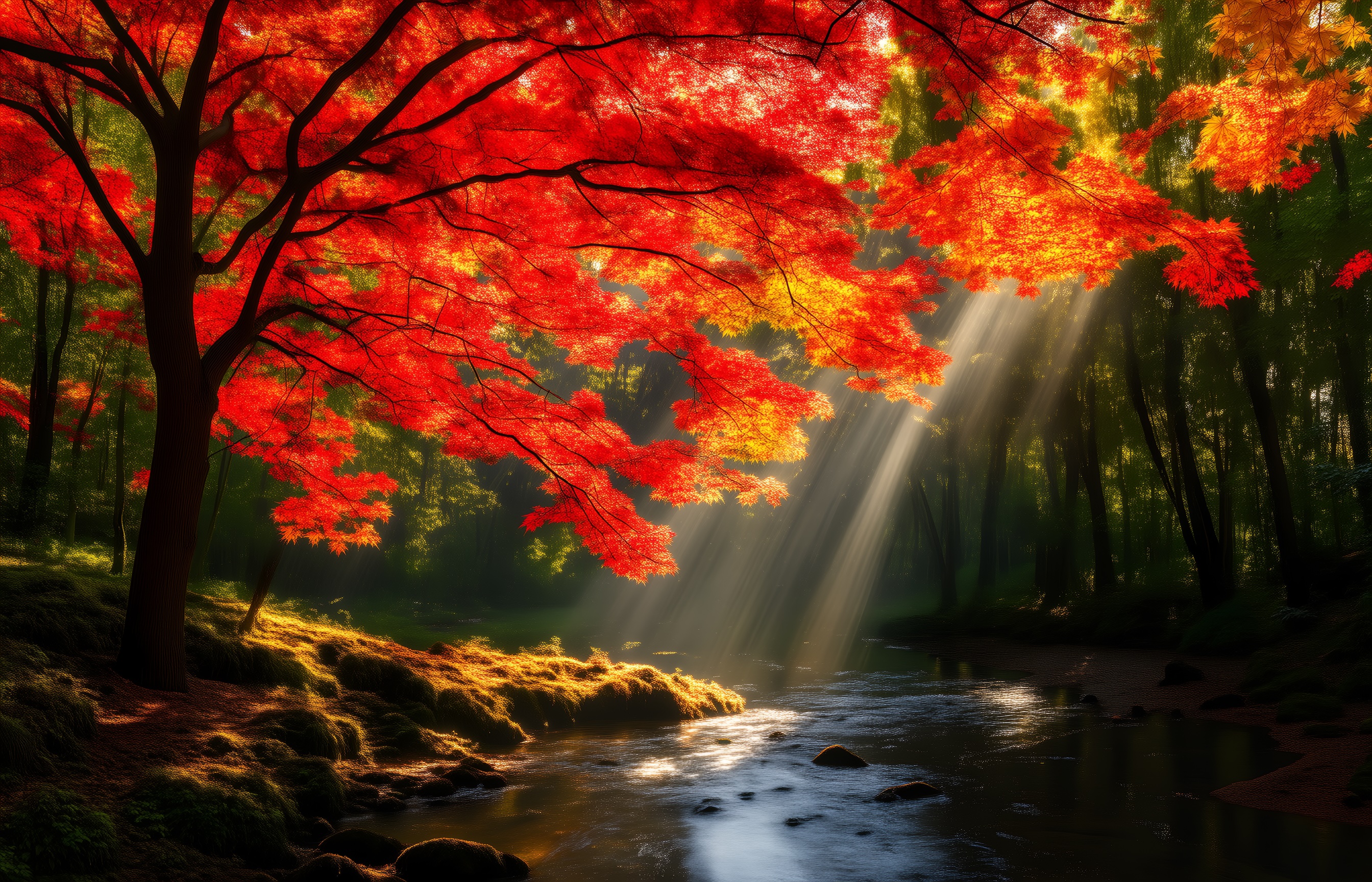}
    \end{minipage}

    \vspace{3pt}

    \begin{minipage}[c]{0.662\textwidth}
        \includegraphics[width=\linewidth]{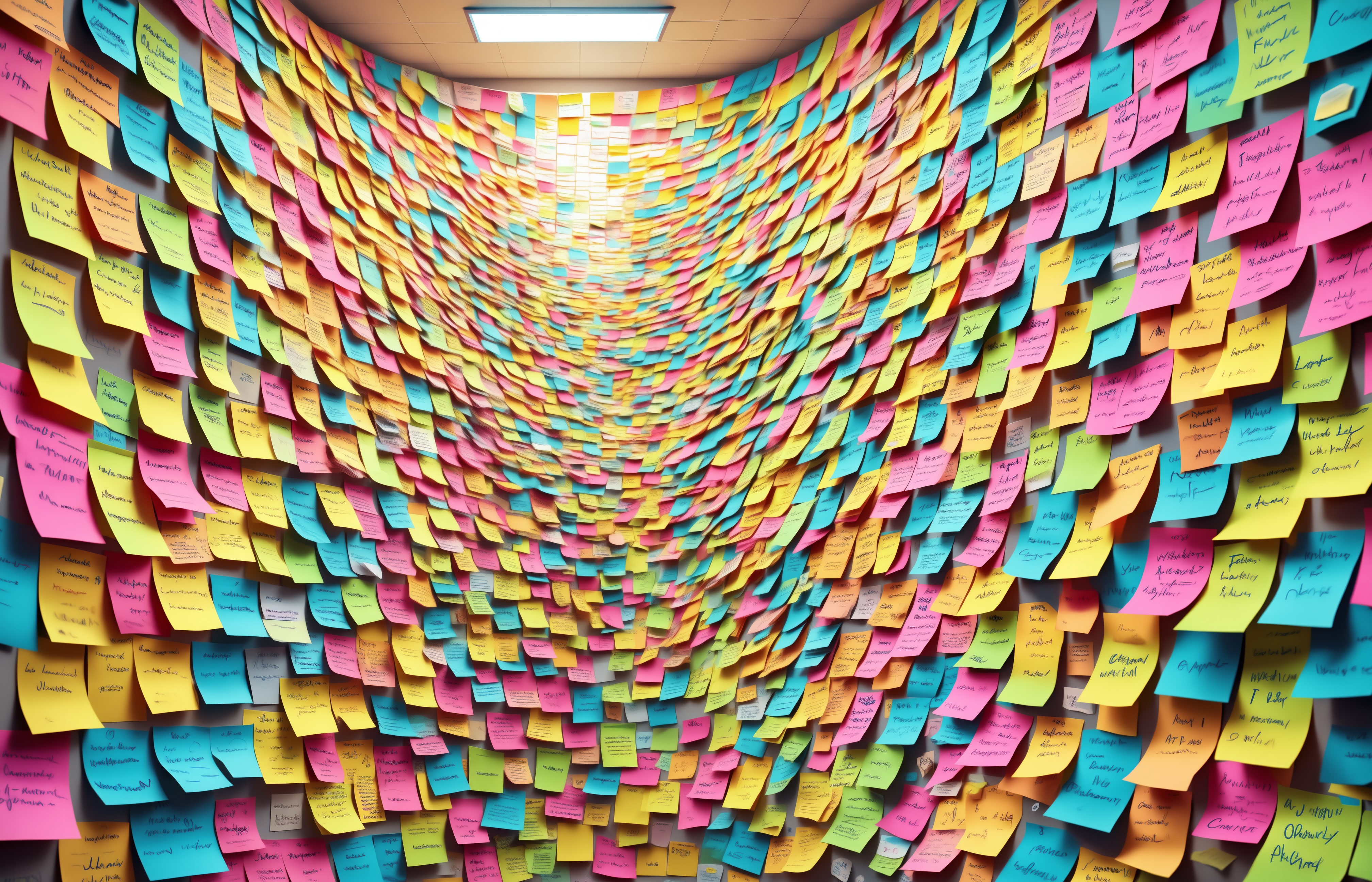}
    \end{minipage}\hfill%
    \begin{minipage}[c]{0.331\textwidth}
        \includegraphics[width=\linewidth]{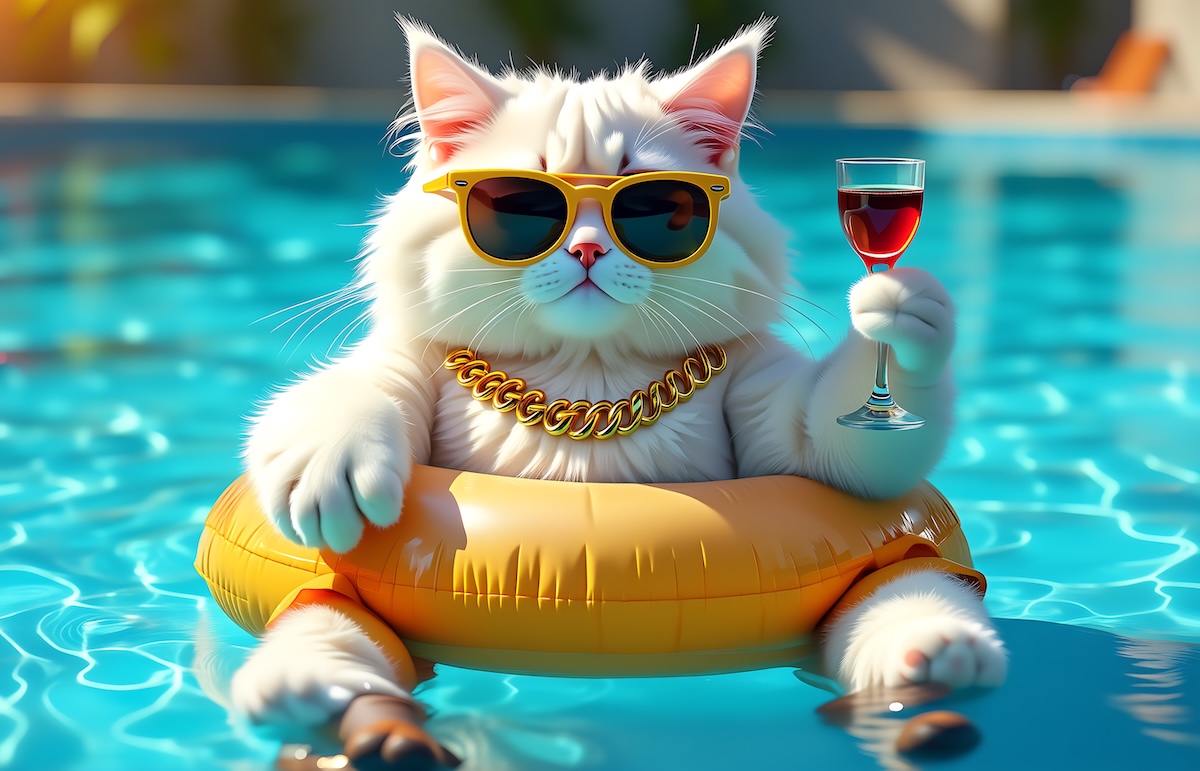}\par\nointerlineskip
        \includegraphics[width=\linewidth]{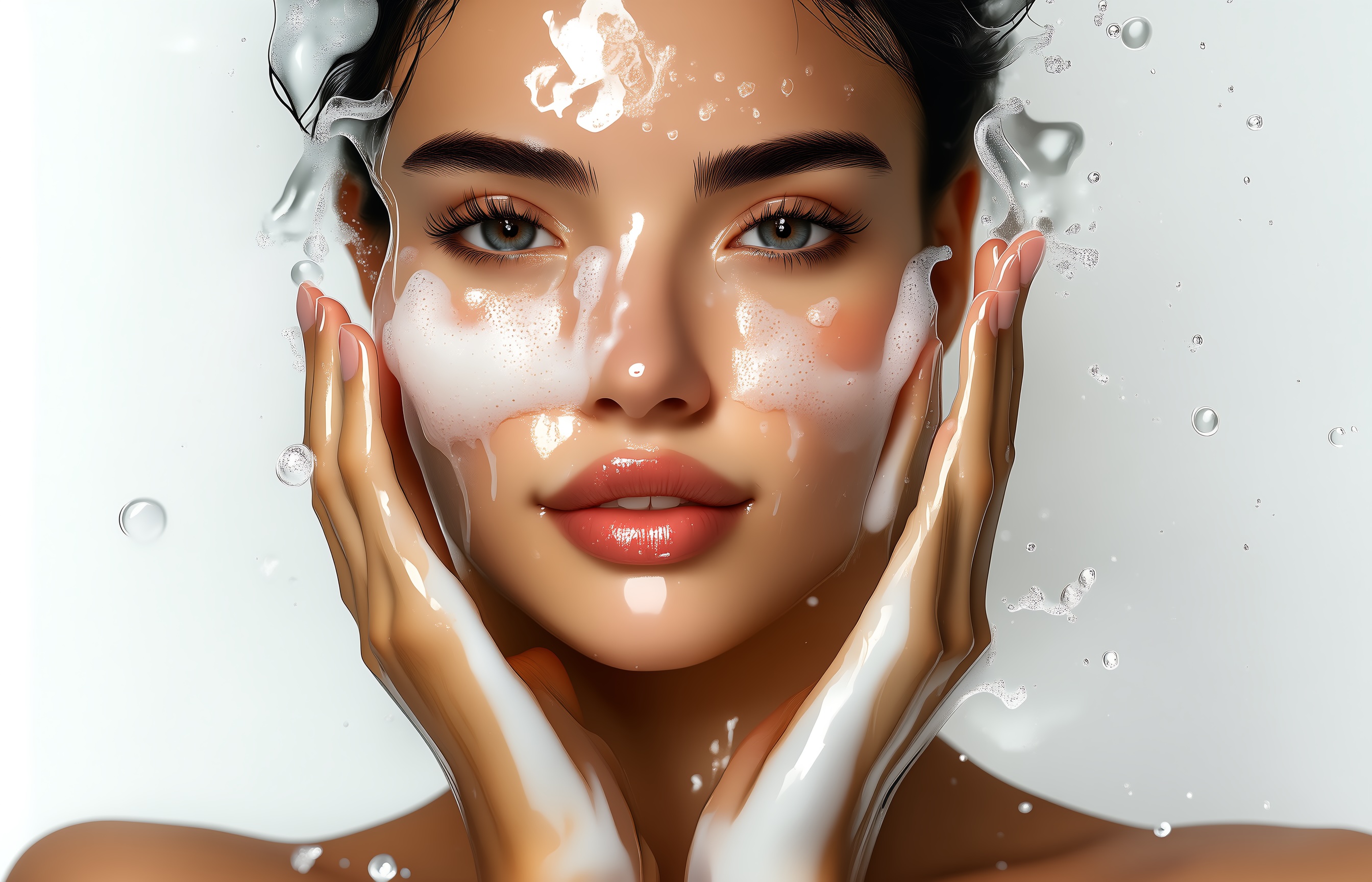}
    \end{minipage}
    \caption{\textbf{Zero-shot high-resolution generation.} Our model generate images at 4K (left panel) and 2K (right panel) resolutions in a zero-shot manner.}
    \label{fig:zero_shot_2k_4k}
\end{figure*}

\begin{figure*}[t]
    \vspace{-4em}
    \centering
    \includegraphics[width=0.19\textwidth]{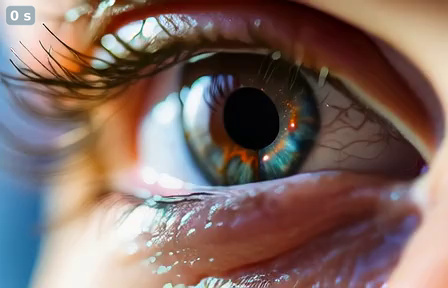}\hfill
    \includegraphics[width=0.19\textwidth]{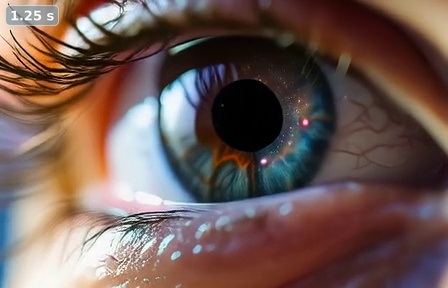}\hfill
    \includegraphics[width=0.19\textwidth]{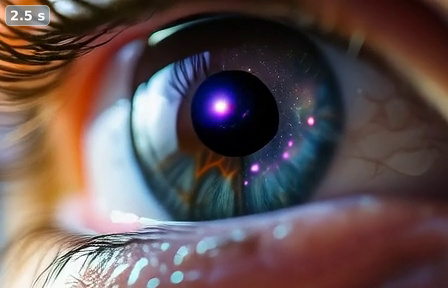}\hfill
    \includegraphics[width=0.19\textwidth]{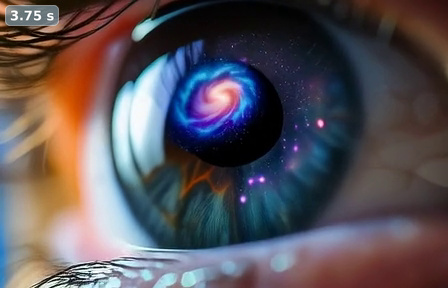}\hfill
    \includegraphics[width=0.19\textwidth]{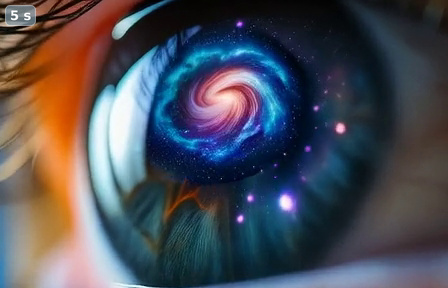}\\[3pt]
    \includegraphics[width=0.19\textwidth]{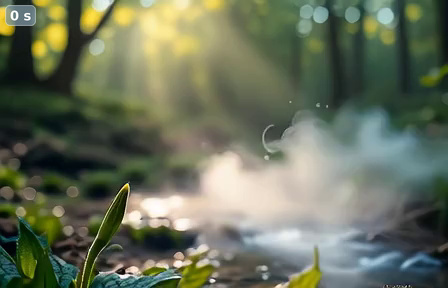}\hfill
    \includegraphics[width=0.19\textwidth]{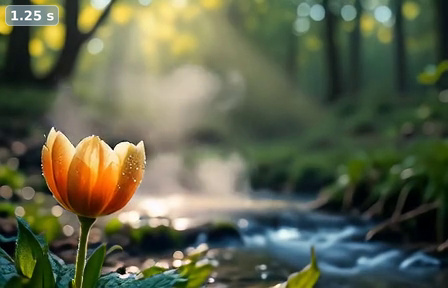}\hfill
    \includegraphics[width=0.19\textwidth]{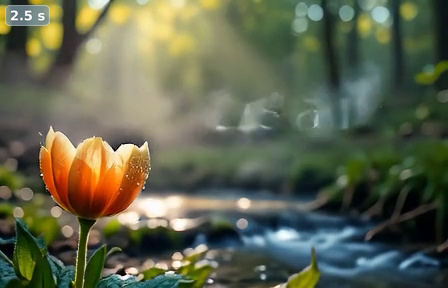}\hfill
    \includegraphics[width=0.19\textwidth]{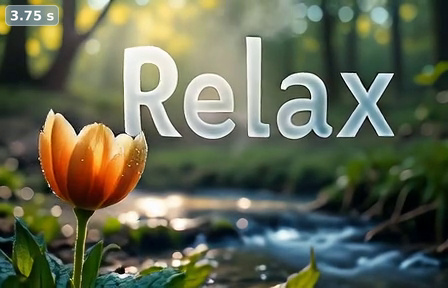}\hfill
    \includegraphics[width=0.19\textwidth]{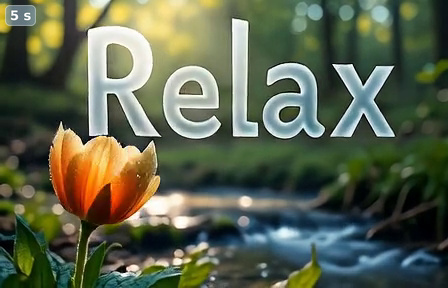}\\[3pt]
    \includegraphics[width=0.19\textwidth]{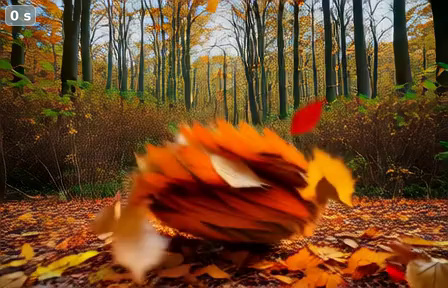}\hfill
    \includegraphics[width=0.19\textwidth]{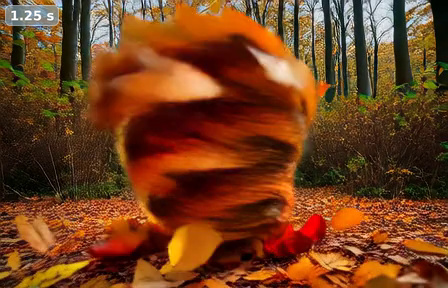}\hfill
    \includegraphics[width=0.19\textwidth]{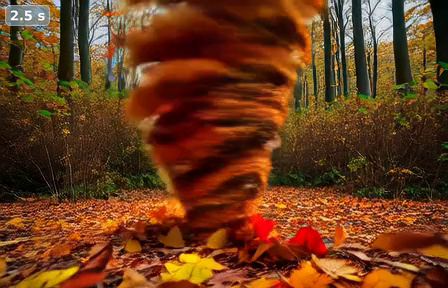}\hfill
    \includegraphics[width=0.19\textwidth]{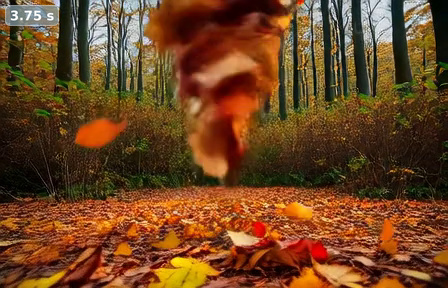}\hfill
    \includegraphics[width=0.19\textwidth]{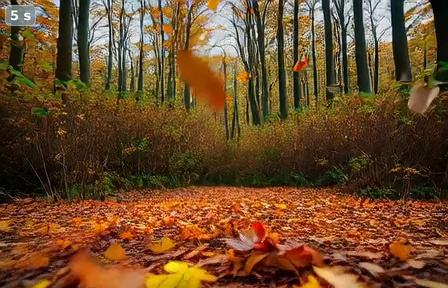}\\[3pt]
    \includegraphics[width=0.19\textwidth]{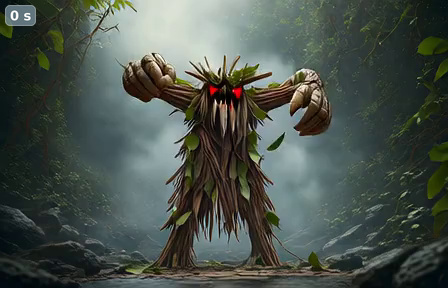}\hfill
    \includegraphics[width=0.19\textwidth]{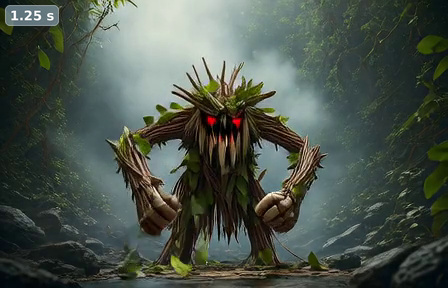}\hfill
    \includegraphics[width=0.19\textwidth]{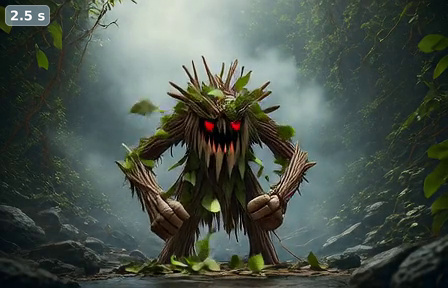}\hfill
    \includegraphics[width=0.19\textwidth]{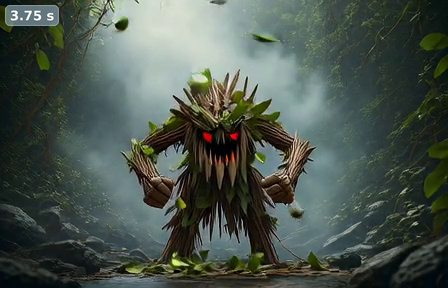}\hfill
    \includegraphics[width=0.19\textwidth]{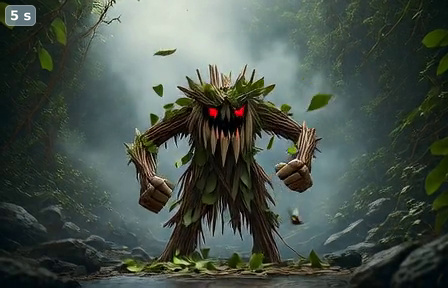}\\[3pt]
    \includegraphics[width=0.19\textwidth]{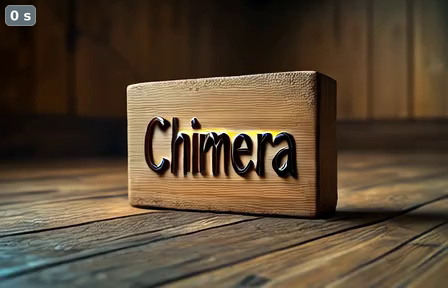}\hfill
    \includegraphics[width=0.19\textwidth]{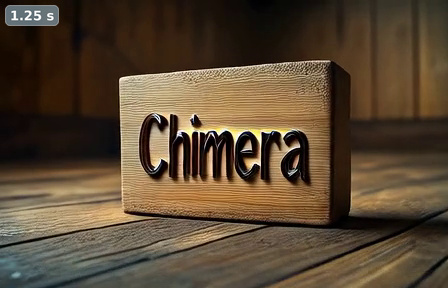}\hfill
    \includegraphics[width=0.19\textwidth]{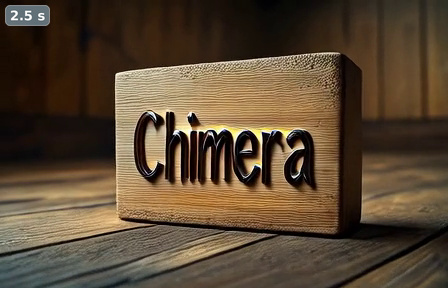}\hfill
    \includegraphics[width=0.19\textwidth]{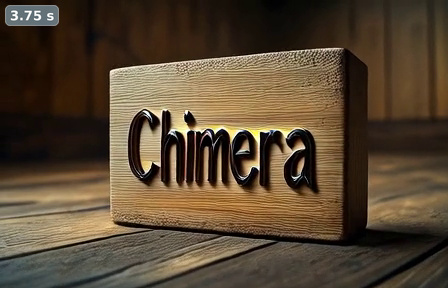}\hfill
    \includegraphics[width=0.19\textwidth]{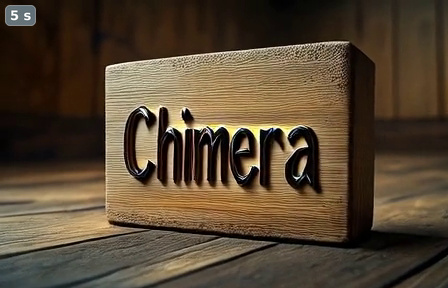}
    \caption{\textbf{Qualitative text-to-video samples from \ours{}.}
    Each row shows five evenly spaced frames of one $81$-frame video generated at $448\times288$ resolution with classifier-free guidance scale $8$, with time proceeding from left to right.
    }
    \label{fig:gen_samples_video}
\end{figure*}

\begin{figure*}[!t]
    \centering

    \begin{subfigure}[t]{0.242\textwidth}
        \centering
        \includegraphics[width=\linewidth]{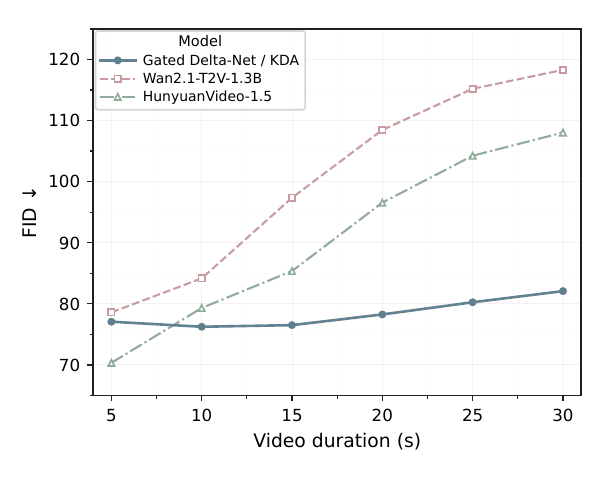}
        \caption{Absolute FID.}
        \label{fig:length_extrapolation_fid_absolute}
    \end{subfigure}
    \hfill
    \begin{subfigure}[t]{0.242\textwidth}
        \centering
        \includegraphics[width=\linewidth]{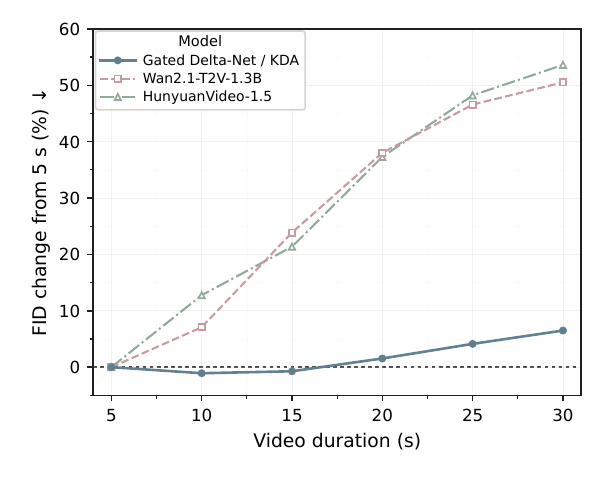}
        \caption{FID change from 5\,s.}
        \label{fig:length_extrapolation_fid_relative}
    \end{subfigure}
    \hfill
    \begin{subfigure}[t]{0.242\textwidth}
        \centering
        \includegraphics[width=\linewidth]{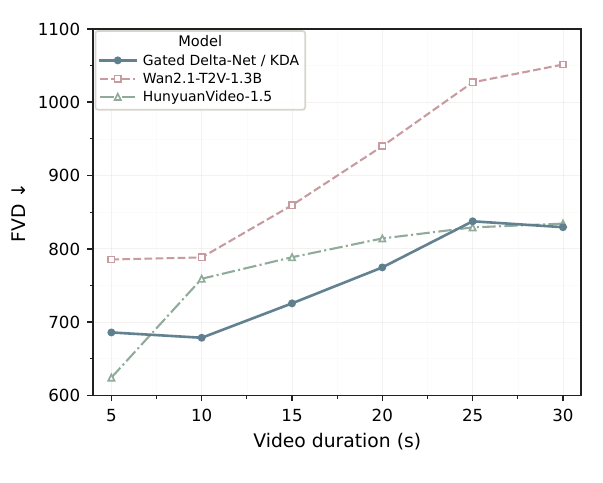}
        \caption{Absolute FVD.}
        \label{fig:length_extrapolation_fvd_absolute}
    \end{subfigure}
    \hfill
    \begin{subfigure}[t]{0.242\textwidth}
        \centering
        \includegraphics[width=\linewidth]{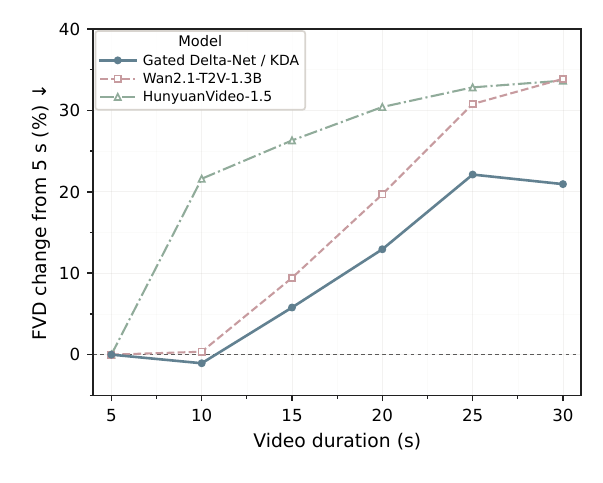}
        \caption{FVD change from 5\,s.}
        \label{fig:length_extrapolation_fvd_relative}
    \end{subfigure}
    \vspace{-0.1in}
    \caption{
    \textbf{Zero-shot video-length extrapolation beyond the 5-second training horizon.}
    We generate videos of 5--30 seconds without length-specific finetuning and evaluate only the final 5 seconds of every video.
    KDA denotes our \ours{} model.
    Panels (a) and (c) report absolute FID and FVD, while panels (b) and (d) report the percentage change relative to each model's own 5-second result.
    Evaluation uses 512 videos per model and duration, with $512{\times}81$ tail frames for FID and 512 tail clips for FVD.
    Lower is better, and a positive relative change denotes degradation.
    }
    \vspace{-0.1in}\label{fig:zero_shot_length_extrapolation}
\end{figure*}

Fig.~\ref{fig:gen_samples} presents text-to-image samples generated by \ours{} at a resolution of $1344\times864$, using prompts ranging from short phrases to detailed descriptions. The samples cover diverse visual styles, including photorealistic macro and portrait photography, studio-style food photography, watercolor and ink illustration, and anime-inspired artwork. Across these examples, \ours{} captures fine-grained structures such as petal and stamen geometry, skin texture, and reflections on liquid and glass surfaces, while maintaining coherent lighting, depth of field, object boundaries, and artistic style. The last row further examines typography-aware generation, with each example combining a mythical chimera with the word ``\textit{Chimera}.'' The word is rendered using materials specified by the prompts, including glowing bubbles, drizzled sauce, and marker strokes, while remaining legible and visually consistent with the geometry and illumination of each scene. Overall, these examples suggest that \ours{} can produce high-resolution images with strong local detail, global coherence, and text--image alignment despite relying predominantly on linear attention. Despite being trained exclusively on 1K images, \ours{} can directly generate coherent 2K and 4K images in a zero-shot manner, without resolution-specific finetuning.

Fig.~\ref{fig:gen_samples_video} presents qualitative results for text-to-video generation, showing five uniformly sampled frames from each 81-frame (5-second) video. Across diverse scenarios involving object emergence, natural and articulated motion, and camera movement, \ours{} produces temporally consistent content while preserving object identity, scene composition, and lighting. The examples containing the words \textit{Relax}'' and \textit{Chimera}'' further show that rendered text remains legible and geometrically consistent across frames. These results suggest that \ours{} maintains strong semantic and visual consistency over time.

\paragraph{\textbf{Zero-shot length extrapolation.}}

\begin{figure*}[t]
    \centering
    \newcommand{\longvideorow}[7]{%
        \raisebox{0.043\textwidth}{\makebox[0.028\textwidth][c]{\rotatebox[origin=c]{90}{\scriptsize #1}}}\hfill%
        \includegraphics[width=0.158\textwidth]{#2}\hfill%
        \includegraphics[width=0.158\textwidth]{#3}\hfill%
        \includegraphics[width=0.158\textwidth]{#4}\hfill%
        \includegraphics[width=0.158\textwidth]{#5}\hfill%
        \includegraphics[width=0.158\textwidth]{#6}\hfill%
        \includegraphics[width=0.158\textwidth]{#7}%
    }

    \begin{subfigure}[t]{\textwidth}
        \centering
        \longvideorow{Wan2.1}%
            {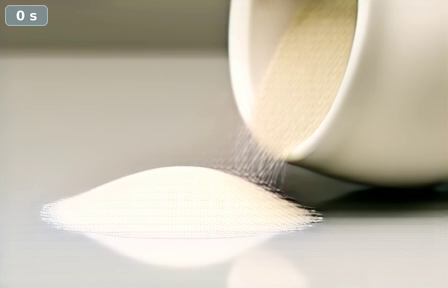}%
            {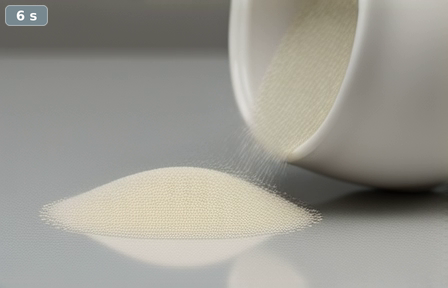}%
            {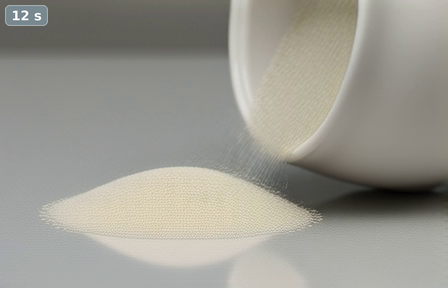}%
            {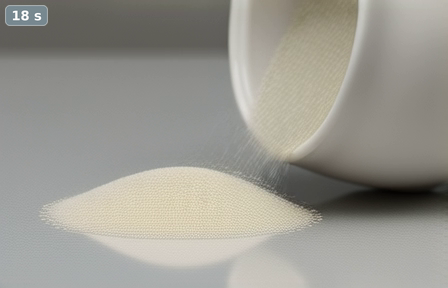}%
            {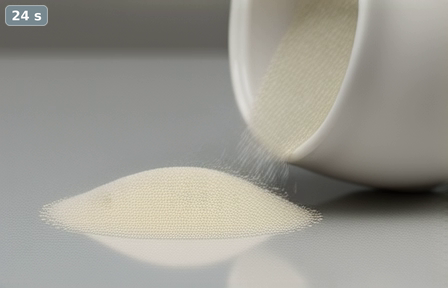}%
            {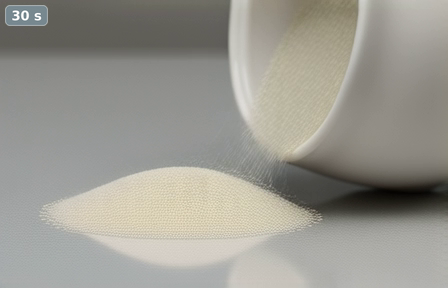}\\[1pt]
        \longvideorow{HunyuanVideo-1.5}%
            {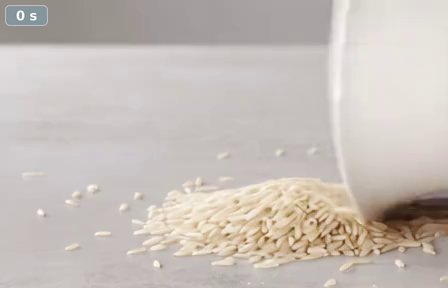}%
            {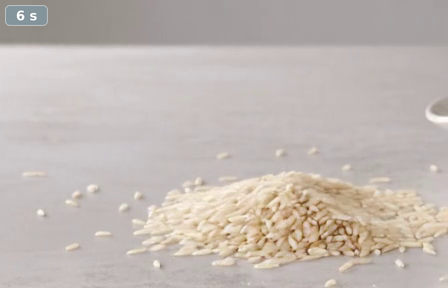}%
            {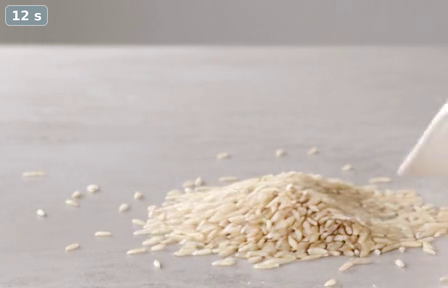}%
            {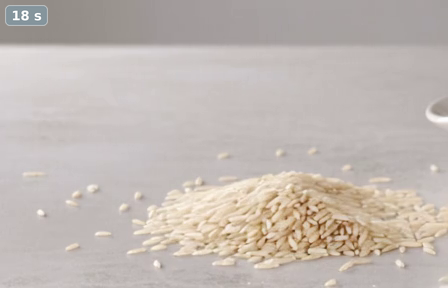}%
            {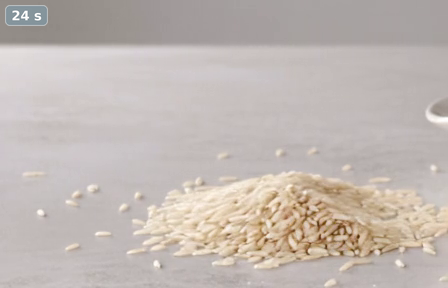}%
            {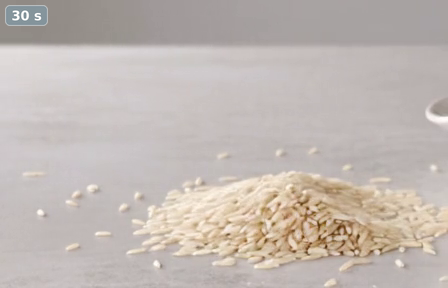}\\[1pt]
        \longvideorow{\ours{}}%
            {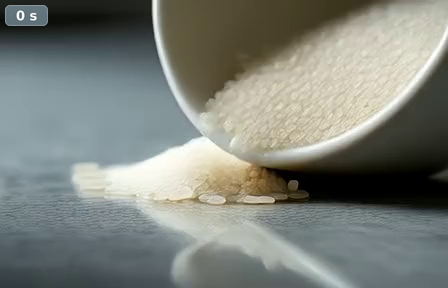}%
            {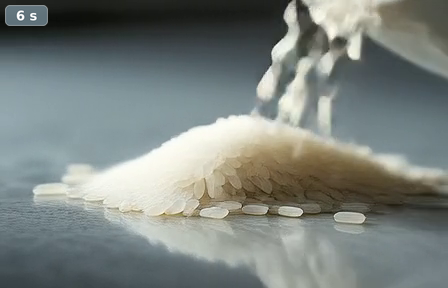}%
            {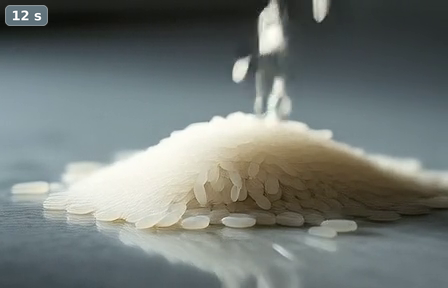}%
            {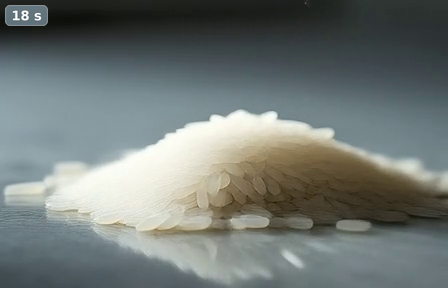}%
            {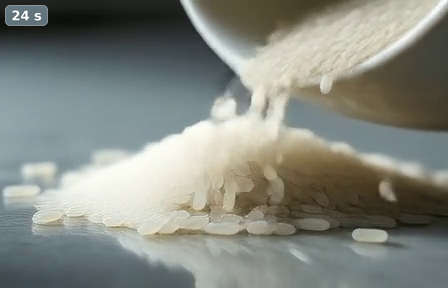}%
            {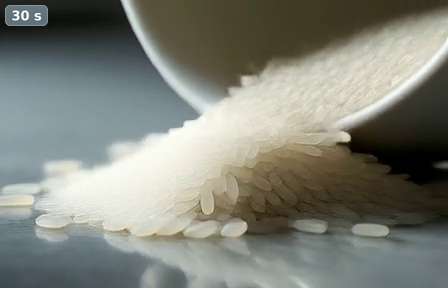}
    \caption{\footnotesize\textit{The bowl slowly tips from upright to its side and remains at rest while rice grains pour out in a steady stream, cascading downward and spreading forward across the counter, some grains bouncing slightly before settling into a small mound; the countertop stays stationary}.}
    \end{subfigure}

    \vspace{2pt}

    \begin{subfigure}[t]{\textwidth}
        \centering
        \longvideorow{Wan2.1}%
            {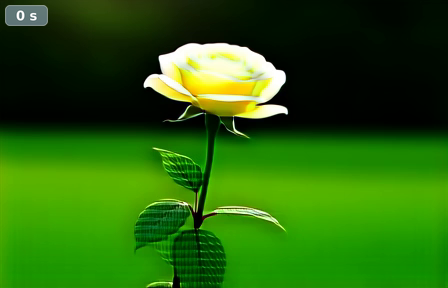}%
            {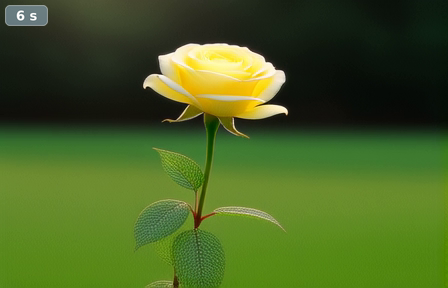}%
            {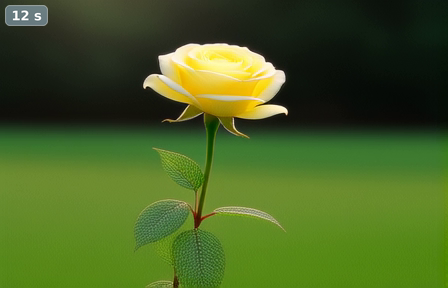}%
            {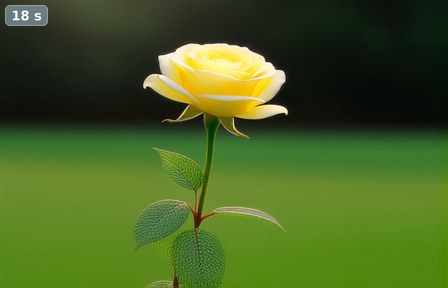}%
            {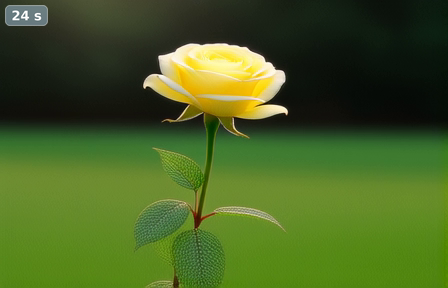}%
            {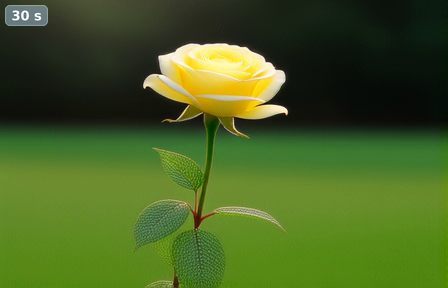}\\[1pt]
        \longvideorow{HunyuanVideo-1.5}%
            {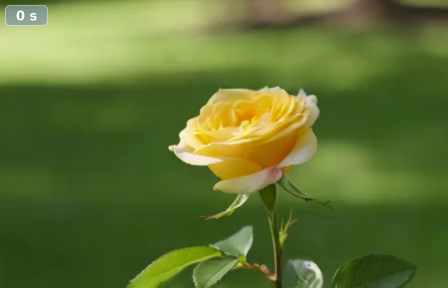}%
            {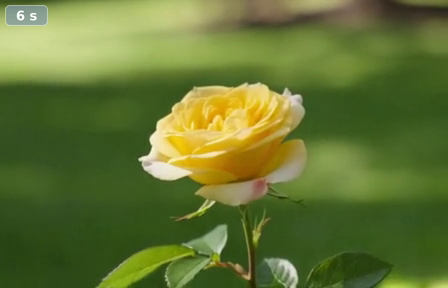}%
            {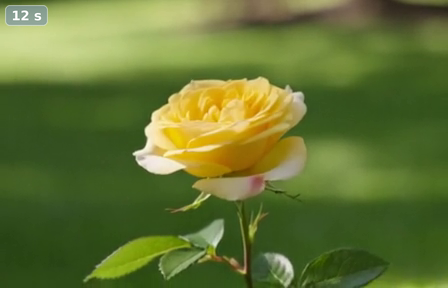}%
            {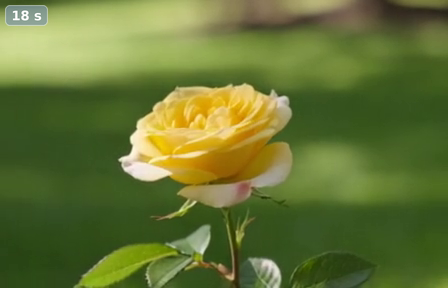}%
            {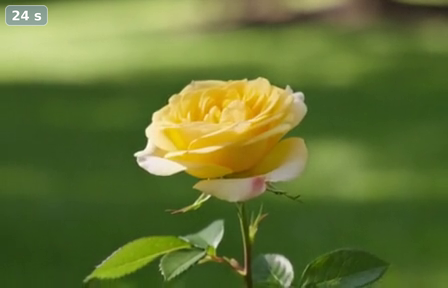}%
            {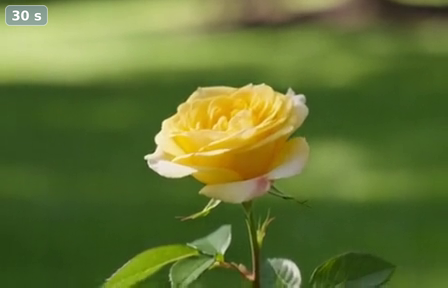}\\[1pt]
        \longvideorow{\ours{}}%
            {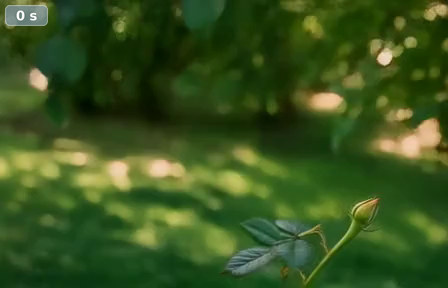}%
            {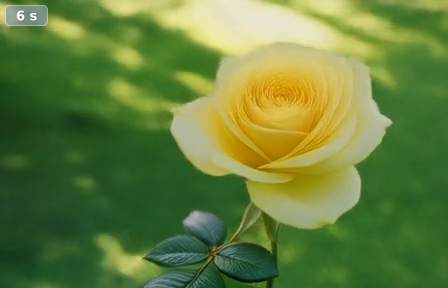}%
            {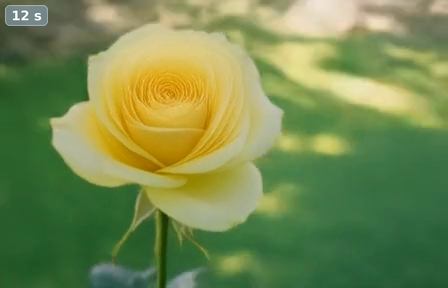}%
            {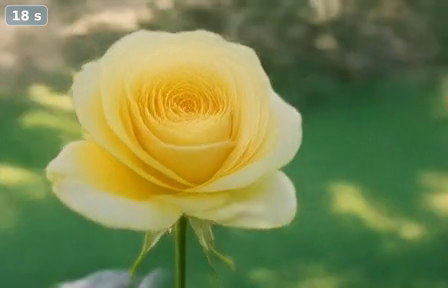}%
            {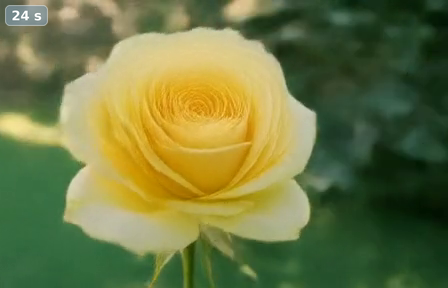}%
            {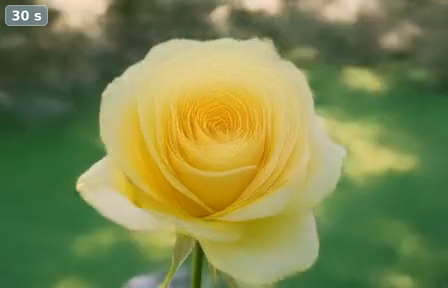}
        \caption{\footnotesize\textit{The rose’s petals gradually unfurl outward from the center in a slow, natural opening motion while the background stays fixed; no other movement occurs.}}
    \end{subfigure}
    \vspace{-0.2em}
    \caption{\textbf{Qualitative zero-shot video-length extrapolation to 30 seconds.}
    For each prompt, we show six frames sampled at 6-second intervals from videos generated by Wan2.1, HunyuanVideo-1.5, and \ours{}, with time proceeding from left to right.
    All models use the same prompt, seed, resolution, frame rate, and duration within each comparison.
    }
    \label{fig:long_video_comparisons}
    \vspace{-1.2em}
\end{figure*}

To test whether KDA and our modality-aware causal convolution provide a natural mechanism for length extrapolation, we evaluate a \ours{} checkpoint trained with 5-second video clips at inference lengths of 5, 10, 15, 20, 25, and 30 seconds, without length-specific finetuning or positional interpolation. We compare against Wan2.1-T2V-1.3B~\citep{wan2025wan} and HunyuanVideo-1.5~\citep{hunyuanvideo2025}. For every model and duration, we generate 512 videos and evaluate them against 512 reference videos sampled from the validation set. To keep the evaluation window fixed while measuring quality progressively farther beyond the training horizon, we compute both metrics only on the final 5 seconds of each video. This yields $512{\times}81$ generated frames for FID and 512 generated tail clips for FVD. We use the same prompts, seeds, resolution, frame rate, and duration grid across all models, although the shared $448{\times}288$ and 16-fps setting lies outside the baselines' native configurations.

Fig.~\ref{fig:zero_shot_length_extrapolation} shows that \ours{} preserves quality substantially better as the generation horizon increases. From 5 to 30 seconds, its FID increases only from $77.1$ to $82.1$, a relative degradation of $6.5\%$, compared with $50.5\%$ for Wan2.1 and $53.6\%$ for HunyuanVideo-1.5. Its FVD increases from $685.8$ to $829.5$, corresponding to $20.9\%$ degradation, whereas Wan2.1 and HunyuanVideo-1.5 degrade by $33.9\%$ and $33.7\%$, respectively. At 30 seconds, \ours{} also attains the lowest measured absolute FID and FVD among the three models. These results are consistent with the architectural motivation of our NoPE design: KDA repeatedly applies the same content-adaptive recurrent update beyond the training horizon, while the temporally causal convolution supplies local structure through fixed-support kernels that introduce no length-dependent positional phases. The resulting backbone thus exhibits natural zero-shot temporal extrapolation.

The qualitative comparisons in Fig.~\ref{fig:long_video_comparisons} reveal the same trend: Wan2.1 becomes blurry and nearly static during long-horizon inference, while HunyuanVideo-1.5 produces sharper frames but likewise exhibits little motion. In contrast, \ours{} remains sharp and continues to evolve throughout the 30-second sequence, avoiding both failure modes.

\paragraph{\textbf{Benchmark evaluation.}}
\begin{table}[ht]
\centering
\small
\setlength{\tabcolsep}{5pt}
\renewcommand{\arraystretch}{1.08}
\caption{
Text-to-image generation results on GenEval and DPG-Bench.
Higher values are better. Note that our \ours{} is trained with a budget of only 600 H100 days.
}
\vspace{-0.05in}
\label{tab:geneval_dpg}
{%
\begin{tabular*}{\textwidth}
  {@{\extracolsep{\fill}}lcccccccc@{}}
\toprule
& \multicolumn{7}{c}{\textbf{GenEval}} & \textbf{DPG} \\
\cmidrule(lr){2-8}\cmidrule(l){9-9}
\textbf{Model}
& \textbf{Single Obj.}
& \textbf{Two Obj.}
& \textbf{Counting}
& \textbf{Colors}
& \textbf{Position}
& \textbf{Color Attri.}
& \textbf{Overall}$\uparrow$
& \textbf{Overall}$\uparrow$ \\
\midrule
\multicolumn{9}{@{}l}{\textit{Generation Only}} \\
\midrule
PixArt-$\alpha$ \citep{chen2024pixart}
    & 0.98 & 0.50 & 0.44 & 0.80 & 0.08 & 0.07 & 0.48 & 71.11 \\
SD v2.1 \citep{rombach2022ldm}
    & 0.98 & 0.51 & 0.44 & 0.85 & 0.07 & 0.17 & 0.50 & 68.09 \\
DALL-E 2 \citep{ramesh2022hierarchical}
    & 0.94 & 0.66 & 0.49 & 0.77 & 0.10 & 0.19 & 0.52 & -- \\
SDXL
    & 0.98 & 0.74 & 0.39 & 0.85 & 0.15 & 0.23 & 0.55 & 74.65 \\
DALL-E 3
    & 0.96 & 0.87 & 0.47 & 0.83 & 0.43 & 0.45 & 0.67 & 83.50 \\
SD3-Medium \citep{esser2024scaling}
    & 0.99 & 0.94 & 0.72 & 0.89 & 0.33 & 0.60 & 0.74 & 84.08 \\
FLUX.1-dev\textsuperscript{\ensuremath{\dagger}}
    & 0.98 & 0.93 & 0.75 & 0.93 & 0.68 & 0.65 & 0.82 & 84.00 \\
Seedream 3.0
    & 0.99 & 0.96 & 0.91 & 0.93 & 0.47 & 0.80 & 0.84 & 88.27 \\
Z-Image-Turbo \citep{cai2025zimage}
    & 1.00 & 0.95 & 0.77 & 0.89 & 0.65 & 0.68 & 0.82 & 84.86 \\
\midrule

\ours
    & 0.98 & 0.88 & 0.85 & 0.88 & 0.62 & 0.68 & 0.82 & 85.12 \\
\bottomrule
\end{tabular*}%
}
\end{table}

To quantitatively assess the image-generation capability of \ours{}, we evaluate it on two widely used image benchmarks, GenEval and DPG-Bench. Together, these benchmarks measure compositional accuracy across attributes such as object count, color, and spatial relations, as well as overall text--image alignment. Table~\ref{tab:geneval_dpg} reports the results.

\ours{} achieves an overall GenEval score of $0.82$ and a DPG-Bench score of $85.12$. It matches FLUX.1-dev and Z-Image-Turbo on GenEval while outperforming both on DPG-Bench, and it surpasses every other evaluated baseline except Seedream~3.0 on both aggregate metrics. Although our limited training budget prevents \ours{} from matching the strongest large-scale image-generation model, its consistent advantage over the remaining baselines demonstrates strong compositional understanding and prompt alignment for a unified model trained with much less compute.

\subsection{Ablation Studies}
\label{sec:exp_ablation}

\begin{wrapfigure}{r}{0.4\textwidth}
    \centering
    \vspace{-1.7em}
    \includegraphics[width=0.4\textwidth]{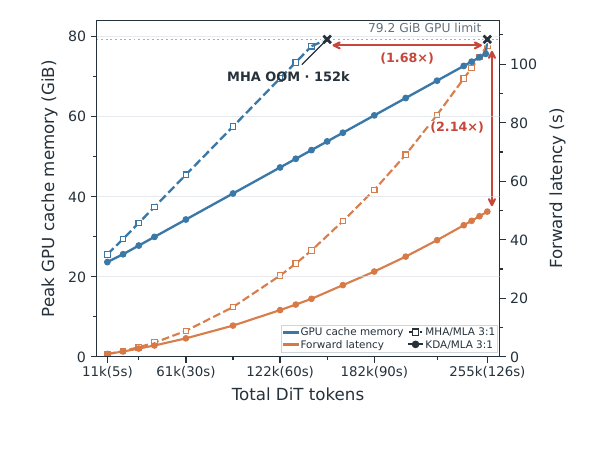}
    \vspace{-3.3em}
    \caption{
    \textbf{Memory and latency growth with tokens.}
    We compare $3{:}1$ KDA/MLA and MHA/MLA backbones with 2B activated parameters.
    }
    \label{fig:kda_mha_memory_latency}
    \vspace{-1.0em}
\end{wrapfigure}

\paragraph{\textbf{Hybrid KDA--MLA attention improves memory efficiency and latency.}}
To isolate the system-level benefit of hybrid attention, we compare our KDA/MLA backbone with an otherwise matched MHA/MLA counterpart, both with approximately 2B activated parameters. Both use the same $3{:}1$ schedule, but every KDA layer in the counterpart is replaced by conventional multi-head self-attention, while the periodic MLA layers and all other components remain unchanged. We use batch size one and BF16 activations on an NVIDIA A100-SXM4-80GB, with $512$ text tokens and $18{\times}28$ visual tokens per temporal slice, and vary the total sequence length. The memory experiment measures peak allocated CUDA memory: we allocate, touch, and retain all cache tensors on device during one complete denoiser forward. The 24 MHA layers store sequence-dependent BF16 key--value caches of size $\mathcal{O}(NHd_h)=\mathcal{O}(ND)$, whereas the 24 KDA layers retain fixed-size FP32 recurrent states of shape $H\times d_h\times d_h$, with memory $\mathcal{O}(Hd_h^2)$ independent of $N$. Both backbones retain the same eight BF16 MLA latent key--value caches. MHA is implemented via FlashAttention. For latency, we time only the denoiser forward, exclude cache construction, synchronize CUDA, and report the median of three post-warm-up runs.

Fig.~\ref{fig:kda_mha_memory_latency} shows that the MHA/MLA backbone reaches the GPU memory limit and first runs out of memory at approximately $152$k tokens, whereas the KDA/MLA backbone completes the $255$k-token test on the same GPU, supporting more than $1.68\times$ the sequence length under the roll-out-cache setting. Although FlashAttention removes the quadratic attention workspace, MHA still incurs quadratic attention arithmetic and its latency therefore grows more rapidly with sequence length. At $255$k tokens, KDA/MLA is $2.14\times$ faster than MHA/MLA. Thus, replacing most MHA layers with KDA substantially reduces persistent cache growth and long-sequence latency, while the shared periodic MLA layers preserve direct global token interactions.
\WFclear

\paragraph{\textbf{MoE, iHC, and \heterop{} improve pre-training compute efficiency.}}
We ablate the three components cumulatively from a 2B Chimera-dense baseline. We first replace its dense FFNs with sparsely activated MoE layers while matching the activated parameter count. We then add iHC to provide adaptive read and write mappings over multiple residual streams. Finally, we apply the \heterop{} (Sec.~\ref{sec:hp_transfer}), which transfers the near-optimal base learning rate and the associated module-wise optimizer settings.

\begin{wrapfigure}{l}{0.4\textwidth}
    \centering
    \vspace{-2.0em}
    \includegraphics[width=0.4\textwidth]{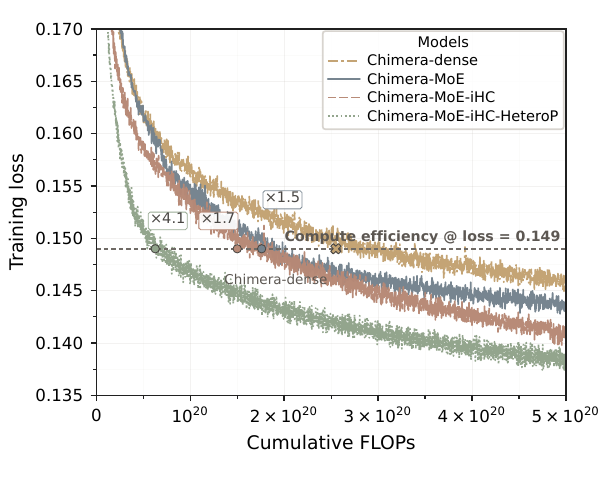}
    \vspace{-2.5em}
    \caption{
    \textbf{Component-wise pre-training compute efficiency.}
    }
    \label{fig:flops_loss_chimera_ablation}
    \vspace{-1.2em}
\end{wrapfigure}

Fig.~\ref{fig:flops_loss_chimera_ablation} shows that every addition improves the loss--compute trajectory. At a common training loss of $0.149$, Chimera-dense requires $2.55\times10^{20}$ FLOPs. Introducing MoE reduces this requirement to $1.76\times10^{20}$ FLOPs, an approximately $1.5\times$ compute-efficiency gain. Adding iHC further reduces it to $1.50\times10^{20}$ FLOPs, yielding a $1.7\times$ gain over the dense baseline. The complete configuration with \heterop{} reaches the same loss with $6.27\times10^{19}$ FLOPs, corresponding to a $4.1\times$ gain. The progression indicates that sparse capacity and multi-stream residual routing provide complementary gains, while \heterop{} contributes the largest additional improvement. Consistent with Fig.~\ref{fig:heterop_lr_transfer}, the \heterop{} parameterization allows the larger base learning rate to remain stable and near-optimal, accelerating convergence without sacrificing training stability.

\paragraph{\textbf{Learning-rate transfer fails without \heterop{}.}}
We first ablate the transfer recipe at the level of individual runs. The bottom row of Fig.~\ref{fig:heterop_lr_transfer} repeats the width and depth sweeps of Sec.~\ref{sec:exp_hp} under standard parameterization (SP), using the same numerical grid for the unscaled global learning rate but without our transfer rules. In contrast to the aligned minima in the top row, the optimal learning rate under SP is no longer consistent across scales. The fitted optima in Figs.~\ref{fig:sp_width} and~\ref{fig:sp_depth} spread over a $6\times$ range along the width family, from $10^{-4}$ for the smallest model to about $6\times10^{-4}$ for the larger ones, and shift toward smaller values as depth grows. No single global learning rate is simultaneously near-optimal for the whole family, so a global  rate tuned on the proxy is necessarily mis-tuned at most other scales.

\begin{figure*}[!t]
    \centering

    \begin{subfigure}[t]{0.32\textwidth}
        \centering
        \includegraphics[width=\linewidth]{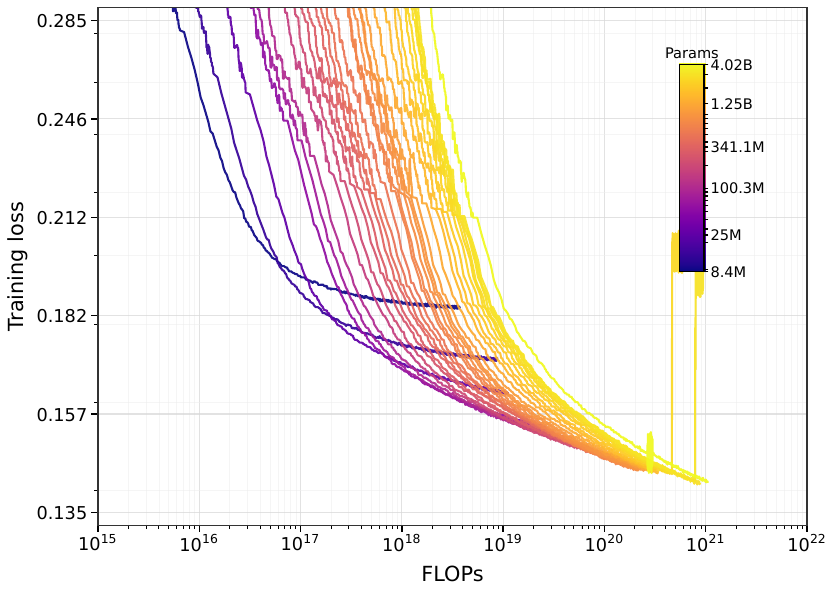}
        \caption{SP training-loss envelope.}
        \label{fig:nomup_loss_flops_256}
    \end{subfigure}
    \hfill
    \begin{subfigure}[t]{0.32\textwidth}
        \centering
        \includegraphics[width=\linewidth]{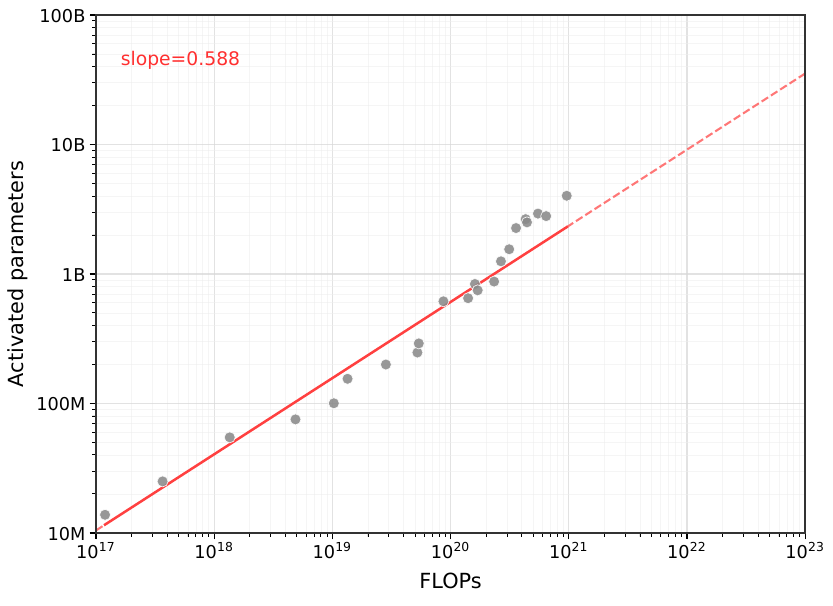}
        \caption{Envelope-optimal model size.}
        \label{fig:nomup_params_flops_256}
    \end{subfigure}
    \hfill
    \begin{subfigure}[t]{0.32\textwidth}
        \centering
        \includegraphics[width=\linewidth]{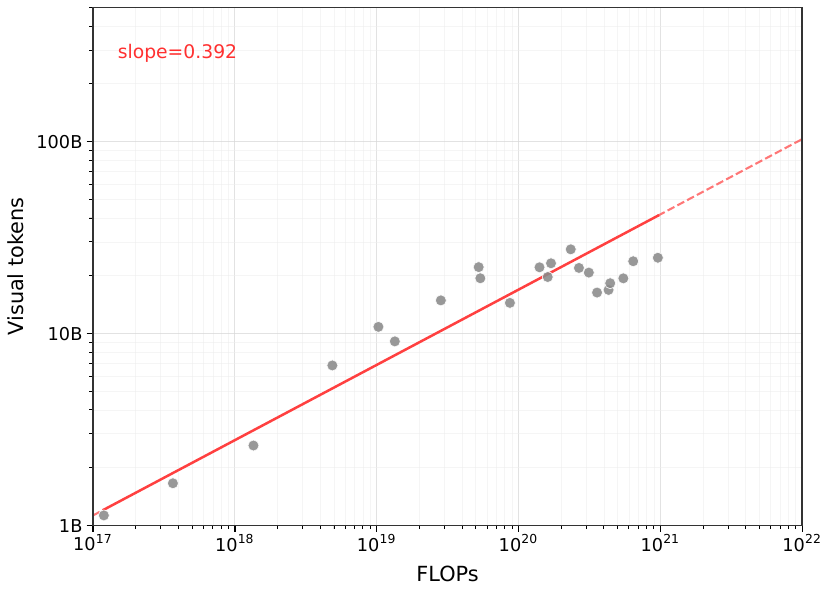}
        \caption{Envelope-optimal training tokens.}
        \label{fig:nomup_tokens_flops_256}
    \end{subfigure}

    \par\vspace{0.75em}

    \begin{subfigure}[t]{0.32\textwidth}
        \centering
        \includegraphics[width=\linewidth]{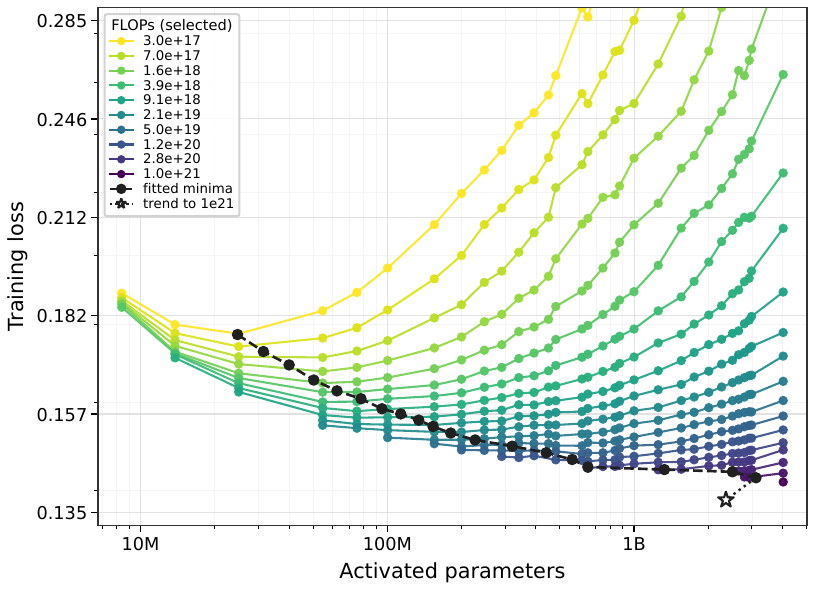}
        \caption{SP IsoFLOP loss profiles.}
        \label{fig:iso_nomup_loss_params_256}
    \end{subfigure}
    \hfill
    \begin{subfigure}[t]{0.32\textwidth}
        \centering
        \includegraphics[width=\linewidth]{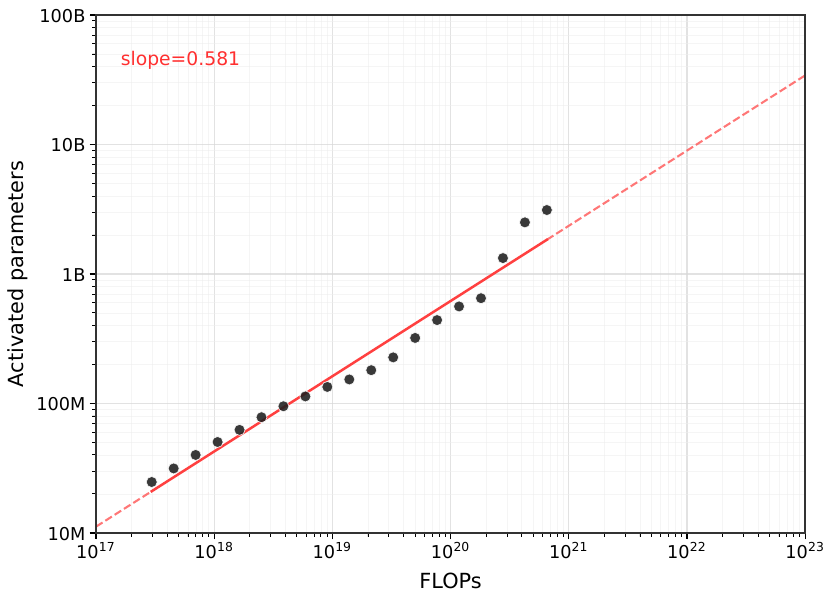}
        \caption{IsoFLOP-optimal model size.}
        \label{fig:iso_nomup_params_flops_256}
    \end{subfigure}
    \hfill
    \begin{subfigure}[t]{0.32\textwidth}
        \centering
        \includegraphics[width=\linewidth]{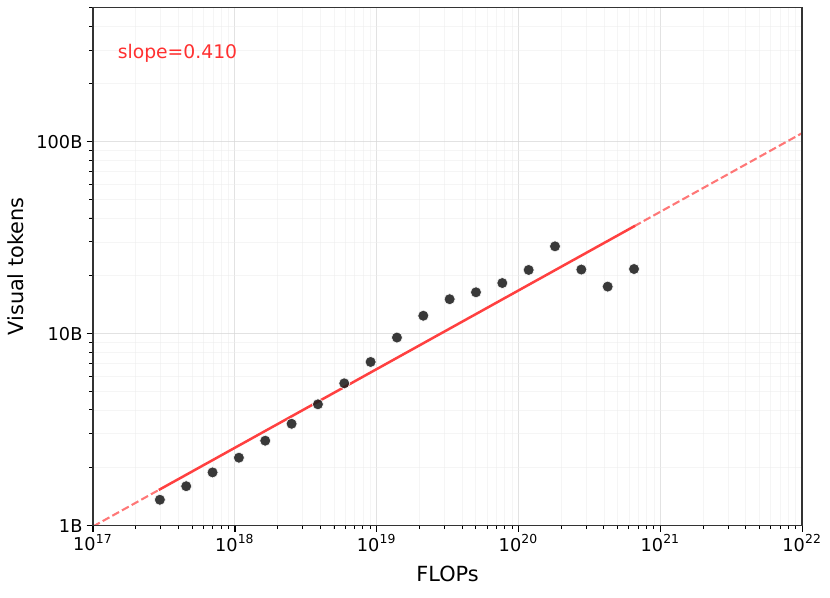}
        \caption{IsoFLOP-optimal training tokens.}
        \label{fig:iso_nomup_tokens_flops_256}
    \end{subfigure}

    \caption{
    \textbf{Compute-optimal scaling \underline{without}  \heterop{} ($256^2$ image pre-training), leading to conservative scaling recipes compared with  Fig.~\ref{fig:chinchilla_256img}.}
    All runs use standard parameterization (SP) with the proxy-tuned global learning rate.
    \textbf{Top row, training-loss envelope:}
    (a)~Diffusion training loss versus cumulative training FLOPs for the same model family as Fig.~\ref{fig:params_flops_256}; the trajectories are noisier and several large models exhibit late-training loss spikes.
    (b)~Envelope-optimal model size $N_{\mathrm{opt}}(C)$, whose fitted exponent increases to $0.588$ from $0.505$ under \heterop{}.
    (c)~The corresponding optimal training tokens $D_{\mathrm{opt}}(C)$, whose exponent decreases to $0.392$ from $0.484$.
    \textbf{Bottom row, IsoFLOP profiles:}
    (d)~Fixed-compute loss profiles, with the fitted minimum of each profile highlighted.
    (e)~IsoFLOP-optimal model size, with exponent $0.581$ instead of the transferred value $0.481$ in Fig.~\ref{fig:iso_params_flops_256}.
    (f)~The corresponding optimal training tokens, with exponent $0.410$ instead of $0.511$.
    Both estimators shift the allocation toward oversized and undertrained models, demonstrating that scale-dependent learning-rate mis-tuning biases the fitted scaling law.
    }
    \label{fig:nomup_chinchilla_256img}
    \label{fig:iso_nomup_chinchilla_256img}
\end{figure*}

The sweeps further reveal a stability failure at high learning rates. Under \heterop{} recipe, every curve keeps a wide and flat basin around $\eta_{\mathrm{base}} = 10^{-3}$ and degrades gracefully up to $\eta_{\mathrm{base}} = 3\times10^{-3}$. Under SP, the loss instead rises abruptly once the global learning rate exceeds each model's optimum by only a small factor, and several runs blow up outright. The smallest width model leaves the plotted loss range already near $4\times10^{-4}$, mid-sized widths explode between $10^{-3}$ and $3\times10^{-3}$, and the deepest models spike sharply at $2\times10^{-3}$. These explosions also break the clean loss ordering by scale that \heterop{} maintains at every swept base learning rate. SP therefore fails in two compounding ways, drifting optima and high-learning-rate divergence, which makes proxy-based hyperparameter tuning unreliable.

\paragraph{\textbf{Scaling-law bias without \heterop{}.}}
The mis-tuning above is not only a per-run nuisance: it biases the scaling law fitted on such runs. To quantify this, we repeat the $256^2$ image study of Sec.~\ref{sec:exp_law} on a matched model family trained under SP, where every model reuses the proxy-tuned global learning rate and therefore carries a scale-dependent mis-tuning penalty. The effect is directly visible in Fig.~\ref{fig:nomup_loss_flops_256}: the loss trajectories are noisier than their \heterop{} counterparts in Fig.~\ref{fig:chinchilla_256img}, and several large models exhibit loss spikes late in training.

The distortion propagates into the fitted law. The envelope estimate on the SP runs gives $N_{\mathrm{opt}}(C)\propto C^{0.588}$ and $D_{\mathrm{opt}}(C)\propto C^{0.392}$ (Fig.~\ref{fig:nomup_chinchilla_256img} (\subref{fig:nomup_loss_flops_256}-\subref{fig:nomup_tokens_flops_256})), and the IsoFLOP estimate gives $C^{0.581}$ and $C^{0.410}$ (Fig.~\ref{fig:iso_nomup_chinchilla_256img} (\subref{fig:iso_nomup_loss_params_256}-\subref{fig:iso_nomup_tokens_flops_256})). Both estimators inflate the model-size exponent by about $0.08$--$0.10$ relative to the \heterop{} fits ($0.505$ and $0.481$) and deflate the token exponent by a similar margin. Because the mis-tuning penalty varies systematically with scale, the per-budget optima are shifted rather than merely noisier, and the near-balanced allocation of Sec.~\ref{sec:exp_law} is replaced by a spurious preference for parameters over data. Taken at face value, the SP law prescribes a compute-optimal model that grows about $1.2\times$ larger per decade of compute relative to the \heterop{} law, reaching roughly $2\times$ oversized, and correspondingly undertrained, when extrapolated three decades beyond the fitted range. \heterop{} is therefore not an optional refinement but a prerequisite for reliable compute-optimal scaling laws.



\subsection{Discussions}
We discuss several techniques, including Muon optimizer, MoE, and details of time embedding.

\textbf{Muon optimizer.} Following language model research, we experiment with using the Muon optimizer~\cite{jordan2024muon} for hidden layers. Although we investigated sufficient efforts in tuning hyper-parameters and splitting the correct parameter groups, we report that the loss under Muon optimizer is systematically worse than using AdamW. The only case where Muon outperforms is the beginning of training under large learning rate. This conclusion holds across different model architectures, scales, and whether \heterop{} and MoE are used. We hypothesize that Muon underperforms Adam in diffusion visual generation because it lacks Adam's implicit low-rank bias. Diffusion training combines denoising objectives across diverse timesteps and noise levels, so the useful optimization signal may concentrate in a small set of stable directions, while many other directions are noisy or inconsistent. To test this, we compare the spectra of parameter updates and trained weights from Adam and Muon models. We find that Adam concentrates most of its update and weight energy in a much smaller number of singular directions, whereas Muon produces substantially more full-rank and spectrally uniform updates. This supports the hypothesis that Adam implicitly filters for persistent low-dimensional denoising directions, while Muon spreads update energy more evenly across the matrix. Consequently, Muon can improve early training under large learning rates, but later makes slower progress because its update energy is diluted away from the stable low-rank subspace important for diffusion optimization.

\textbf{MoE.} We observe that MoE scaling provides much weaker compute efficiency in diffusion-based visual generation than what is commonly observed in language modeling. In our experiments, increasing sparsity yields only around a $1.5 \times$ improvement in compute efficiency, substantially below the gain predicted by the common $\sqrt{\mathrm{sparsity}}$ heuristic. This suggests that simply increasing the number of inactive experts does not translate into proportional effective capacity for diffusion models. We hypothesize that this gap arises because visual denoising offers less exploitable expert specialization: visual tokens are continuous, spatially redundant, and highly correlated, while diffusion timesteps share a large common denoising component that must be modeled consistently. Consistent with this hypothesis, we find that expert routing exhibits weak specialization and often becomes more uniform across tokens or timesteps. These results suggest that MoE sparsity in diffusion models is constrained by the intrinsic shared structure of visual denoising, rather than only by routing or implementation details.

\textbf{Time embedding.} We also observe that the timestep embedding network is a model-size-sensitive component in diffusion training. In our default design, sinusoidal timestep features are mapped to the model dimension through an MLP. However, using an overly wide hidden dimension for this MLP can destabilize training; for example, a 4096-dimensional hidden layer is unstable for a 1024-width backbone. This suggests that the timestep MLP should not be treated as a fixed auxiliary module independent of model scale. We hypothesize that the reason is that timestep embeddings act as a global conditioning signal, often controlling layer-wise modulation such as scale, shift, or gating. If the timestep MLP is too wide relative to the backbone, it can produce an overly strong or sharp timestep-dependent modulation, amplifying gradient conflicts across noise levels and disrupting the shared denoising representation. We therefore scale the hidden dimension of the timestep MLP with the backbone width, which improves training stability.

\section{Related Works} 
\label{sec:relatedwork}

\paragraph{\textbf{Diffusion Models for Visual Generation.}}

Diffusion models have been repeatedly adapted to the structure, scale, and controllability requirements of visual data for improving viusal generation quality. Early work adopt convolutional U-Net backbones~\citep{ho2020denoising, song2020score}, improvements in noise schedules~\citep{nichol2021improved}, and guidance~\citep{dhariwal2021diffusion} to improve generation fidelity.
Later systems adopted hierarchical~\citep{saharia2022photorealistic, ramesh2022hierarchical} and latent-space designs~\citep{rombach2022ldm}. Combined with classifier-free guidance~\citep{ho2022classifier} and strong language encoders~\citep{nichol2021glide}, these models enabled high-quality controllable generation.
Following work has revisited both the denoising backbone and the latent representation. DiT replace convolutional U-Nets with Transformer architectures, enabling stronger scaling~\citep{peebles2023scalable, chen2024pixart, esser2024scaling}. At the same time, the VAE latent space used by latent diffusion has been identified as a bottleneck for visual quality, motivating representation autoencoders~\citep{zheng2025rae, chen2025aligning} and renewed interest in pixel-space or hybrid pixel-latent generation~\citep{li2026back, yu2026pixeldit}.

Beyond image generation, Diffusion models for video generation have evolved through several stages. Early work primarily adapted image diffusion models to videos by adding temporal modeling modules, reusing strong image priors while learning motion~\citep{ho2022video, singer2022make, ho2022imagen, blattmann2023align, blattmann2023stable}. More recently, the focus has shifted from generating short clips to long-horizon video. One line of work reformulates video diffusion as autoregressive or causal rollout~\citep{chen2024diffusion, song2025history, huang2026self}. Another line of work, which is more relevant to \ours{}, is adopting efficient Transformer variants, by leveraging spatial-temporal attention patterns~\citep{xi2025sparse, ding2025efficient, wu2025vmoba}, adopting linear attention~\citep{wang2025lingen, chen2025sana},  and sparse attention~\citep{zhan2025bidirectional, zhang2026faster, tan2025dsv}. In this paper, we 
develop a novel hybrid attention architecture with minimal inductive bias, enabling native joint training of image and video generation.
We showcase efficient, stable training and a scaling law that attains the Chinchilla-optimal allocation.

\paragraph{\textbf{Efficient Attention Mechanisms.}}
The efficient attention mechanisms broadly categorized by what they reduce: the representation being attended to, the attention operator itself, or the set of token interactions. The first line compresses the key-value representation while preserving the standard softmax attention formulation~\citep{shazeer2019mqa, ainslie2023gqa, liu2024deepseekv2}. The second line removes the softmax operation, providing linear-time or constant-state inference while preserving parallelizable training~\citep{katharopoulos2020transformers, sun2023retentive, gu2023mamba, yang2024parallelizing}. The third line sparsifies attention by computing only a subset of token interactions, by designing attention patterns~\citep{child2019generating,zhu2026flare} or making it learnable~\citep{yuan2025native, liu2025deepseekv3p2, zheng2026scaling}. In this work, we adapt the architecture of Kimi Linear~\citep{team2025kimilinear} to visual generation while injecting minimal inductive bias of spatial-temporal modeling.

\paragraph{\textbf{Model Scaling.}}
Model scaling research studies how model quality changes as we allocate more parameters, data, and compute. Early scaling laws showed that language-model loss follows predictable power-law trends with model size, dataset size, and training compute~\citep{kaplan2020scaling}. Compute-optimal scaling later systematically studies the optimal allocation of data and model size given a fixed compute budget~\citep{hoffmann2022chinchilla}. These scaling principles have also been extended beyond language modeling to autoregressive image, video, and multimodal generation~\citep{henighan2020scaling, liang2024scaling}. To reduce the cost of hyperparameter search, a complementary line studies parameterization and hyperparameter transfer~\citep{yang2021tuning, dey2026completep, mlodozeniec2025completed}. We propose \heterop{} that builds on these principles but targets a different structural issue: heterogeneous backbones require module-specific width ratios rather than one global model-width ratio.
Besides, model scaling is not limited to dense parameter growth.
Sparse activation and mixture-of-experts architectures scale total capacity while keeping activated compute relatively low~\citep{lepikhin2020gshard, fedus2022switch, fei2407scaling}. In this work, we incorporate these techniques for a systemically study of scaling laws of visual Diffusion model in the contemporary era.

\section{Future Exploration and Conclusion} 
\label{sec:conclusion}

\textbf{Conclusion.} 
We introduced \ours{}, a hybrid, heterogeneous, single-stream diffusion backbone co-designed with a principled scaling recipe for token-extensive visual generation. Architecturally, \ours{} combines KDA for efficient long-context state tracking, periodic MLA for global interaction, and modality-aware short convolutions for local spatio-temporal structure, enabling a unified text--image--video sequence without positional embeddings. Sparse MoE, iHC, and sandwich normalization further improve capacity and training stability. To scale this heterogeneous architecture, \heterop{} transfers proxy-tuned hyperparameters and produces a consistently tuned model family from which we derive compute-optimal laws over activated model size, visual training tokens, and image--video composition. We hope that the architectural principles embodied by \ours{} provide useful guidance for designing future long-context visual generators. More broadly, we hope \heterop{} and the compute-optimal framework offer a reusable way to turn inexpensive proxy experiments into reliable model, data, and compute-allocation decisions for multimodal generators.

\textbf{Future Exploration.} 
There are some extensions for \ours{}'s future explorations.
First, our study deliberately isolates width and depth transfer. Although \heterop{} parameterizes the heterogeneous modules according to their functional fan-in, the scaling family still holds several structural axes fixed, including the iHC residual-stream count, the MoE expert and top-$K$ routing configuration (and hence its sparsity), and the MLA KV-compression ratio as detailed in Sec.~\ref{sec:hp_transfer}. A more complete and flexible parameterization should transfer from an inexpensive dense proxy to sparse target models and make residual-stream count, expert count, routing sparsity, and related structural choices scale-dependent rather than preset.
Second, while \ours{} already processes text, image, and video tokens in one stream, the present study evaluates only text-to-image and text-to-video generation. Incorporating the successful design choices in \ours{} to single-stream hybrid-linear-global architectures (e.g., Qwen3.5~\citep{team2026qwen3}, K3~\citep{kimiteam2026kimik3openfrontier}) is a promising path toward a single model for both long-context multimodal understanding and generation.
Third, \ours{} still relies on a diffusion-specific AdaLN modulation. Our preliminary experiments suggest that this can be absorbed into iHC by conditioning its stream read and write mappings directly on the diffusion timestep. Removing the separate AdaLN modulation would further align hybrid language-model and hybrid diffusion architectures toward a shared backbone.

\textbf{Acknowledgments.}
We thank Hongwu Peng for insightful discussions of the Muon optimizer and \(\mu\)P; Bi Sai for constructive feedback; and Tong Sun, Rajiv Jain, Kalyan Sunkavalli, and John Yang for supporting this project.

\clearpage
\newpage
\bibliographystyle{assets/plainnat}
\bibliography{paper}

\clearpage
\appendix
\section{Notation}
\label{app:notation}

Table~\ref{tab:notation} summarizes the main notation used throughout the paper. Symbols are grouped by component, and the scope of section-local or reused symbols is stated explicitly in the table.

\begin{table}[!ht]
    \centering
    \footnotesize
    \renewcommand{\arraystretch}{1.08}
    \caption{Main notation used in this paper.}
    \label{tab:notation}
    \begin{minipage}[t]{0.495\linewidth}
    \begin{tabular}[t]{@{}>{\raggedright\arraybackslash}p{0.25\linewidth}@{\hspace{4pt}}>{\raggedright\arraybackslash}p{0.72\linewidth}@{}}
    \toprule
    \textbf{Symbol} & \textbf{Meaning} \\
    \midrule
    \multicolumn{2}{@{}l}{\emph{Diffusion objective (Sec.~\ref{sec:model_arch_preliminary})}} \\
    $z_1$ & clean visual tokens (patchified VAE latents) \\
    $z_0$ & Gaussian noise \\
    $\tau$, $z_\tau$ & diffusion timestep; noised visual tokens at $\tau$ \\
    $v_\tau$ & rectified-flow velocity target $z_1 - z_0$ \\
    $c$ & text-token sequence from the frozen text encoder \\
    $\theta$ & trainable parameters of the denoiser \\
    $\mathcal{L}$ & diffusion training objective evaluated at visual-token positions \\
    \midrule
    \multicolumn{2}{@{}l}{\emph{Sequence and layout (Secs.~\ref{sec:single_stream}, \ref{sec:attention})}} \\
    $x$ & packed input sequence; reused after iHC expansion for the sequence-level multi-stream state \\
    $P_{\mathrm{text}}$, $P_{\mathrm{vis}}$ & modality-specific input projections \\
    $e_\tau$ & timestep embedding \\
    $T$, $H$, $W$ & temporal, height, width sizes of the latent grid \\
    $m$ & flattened index of the temporal-major raster order \\
    $L$ & sequence length \\
    \midrule
    \multicolumn{2}{@{}l}{\emph{Block computation (Secs.~\ref{sec:single_stream}, \ref{sec:ihc})}} \\
    $d$ & shared hidden dimension \\
    $\delta_A$, $\gamma_A$ & AdaLN shift and scale of the attention input \\
    $g^{A}$, $g^{F}$ & timestep-conditioned residual gates \\
    $M$ & number of parallel residual streams ($M=4$) \\
    $R$, $R_m$ & per-token multi-stream state; its $m$-th stream (stream-local index) \\
    $h_{\mathrm{pre}}(R)$, $h_{\mathrm{post}}(R)$ & token-dependent iHC read and write vectors \\
    $\alpha_h$ & learnable scale in the iHC coefficient generator \\
    $\Delta$ & gated attention or FFN update written to the streams \\
    $H_{\mathrm{res}}$ & residual-stream transition matrix (identity in iHC) \\
    \midrule
    \multicolumn{2}{@{}l}{\emph{KDA (Sec.~\ref{sec:attention})}} \\
    $t$ & scan index $1,\ldots,L$ (distinct from diffusion timestep $\tau$) \\
    $q_t$, $k_t$, $v_t$ & query, key, and value of token $t$ \\
    $d_q$, $d_k$, $d_v$ & per-head query, key, and value dimensions \\
    $S_t$ & recurrent state ($d_k \times d_v$) after token $t$ \\
    $\alpha_t$ & channel-wise forget gate (state decay) \\
    $\beta_t$ & head-wise erase/write strength \\
    $\operatorname{mShortConv}$ & modality-aware short conv.\ (Sec.~\ref{sec:short_conv}) \\
    \bottomrule
    \end{tabular}
    \end{minipage}\hfill
    \begin{minipage}[t]{0.495\linewidth}
    \begin{tabular}[t]{@{}>{\raggedright\arraybackslash}p{0.25\linewidth}@{\hspace{4pt}}>{\raggedright\arraybackslash}p{0.72\linewidth}@{}}
    \toprule
    \textbf{Symbol} & \textbf{Meaning} \\
    \midrule
    \multicolumn{2}{@{}l}{\emph{MLA (Sec.~\ref{sec:attention})}} \\
    $n_h$ & number of attention heads \\
    $U$ & compressed key--value latent \\
    $K^{\mathrm{direct}}$ & direct-key branch projected from the MLA input and shared across heads \\
    $K^{\mathrm{lat}}$, $V$ & latent keys and values decoded from $U$ \\
    $K_i$ & head-$i$ key $[K^{\mathrm{direct}},K_i^{\mathrm{lat}}]$ \\
    \midrule
    \multicolumn{2}{@{}l}{\emph{MoE (Sec.~\ref{sec:moe})}} \\
    $u$ & token input to the MoE FFN \\
    $N$, $K$ & section-local expert count ($N=56$); activated experts per token ($K=8$) \\
    $E_i$ & the $i$-th routed expert \\
    $r(\cdot)$, $g_i(u)$ & router; routing weight of expert $i$ \\
    $b \in \mathbb{R}^{N}$ & non-gradient load-balancing bias \\
    $\mathcal{T}(u)$ & top-$K$ expert set selected for token $u$ \\
    \makecell[l]{$\operatorname{Load}_i$, $\overline{\operatorname{Load}}$\\$\operatorname{MaxVio}$} & per-expert load; mean expert load; maximal load violation (Eq.~\ref{eq:maxvio}) \\
    \midrule
    \multicolumn{2}{@{}l}{\emph{RoPE analysis (Sec.~\ref{sec:attention}, App.~\ref{app:rope_probe})}} \\
    $\theta_j$ & rotation frequency of rotary pair $j$ (rad/token) \\
    $\delta$ & token offset, query position minus key position \\
    $q^{(i)}$, $k^{(i)}$ & components of $q$, $k$ in rotary channel pair $i$ \\
    $\|q^{(i)}\|\!\|k^{(i)}\|$ & amplitude of pair $i$ (window-averaged $q$, $k$) \\
    $E_i$ & per-pair energy $\mathbb{E}_{\ell}\,\|q^{(i)}_{\ell}\|\|k^{(i)}_{\ell}\|$ over token positions (distinct from expert $E_i$) \\
    local / far mass & attention fraction at Chebyshev distance $\leq 1$ / $\geq 5$ \\
    \midrule
    \multicolumn{2}{@{}l}{\emph{Scaling (Secs.~\ref{sec:scaling}, \ref{sec:exp_law})}} \\
    $\mathcal{M}^{(0)}$, $\mathcal{M}$ & proxy and target models \\
    $n_{\mathrm{blk}}^{(0)}$, $n_{\mathrm{blk}}$ & proxy and target block counts \\
    $W^{(0)}$, $W$ & corresponding proxy and target parameter groups \\
    $m_W$, $m_L$ & target-to-proxy fan-in ratio; target-to-proxy block-count ratio \\
    $\bar{\mathbf{h}}$ & tuple of proxy-tuned base hyperparameters \\
    $\mathcal{P}_W$, $\rho(W)$ & transferred settings for $W$; its functional role \\
    $\mathcal{T}_{\rho(W)}$ & role-dependent transfer map \\
    $N$ & activated parameter count in the scaling law \\
    $D$ & cumulative number of visual latent positions processed \\
    $C$ & training compute, approximated by $C \approx 6ND$ \\
    $\widehat{\mathcal{L}}$ & fitted diffusion-loss surface \\
    $E$; $A,B$; $a,b$ & irreducible loss; positive coefficients; scaling exponents \\
    \bottomrule
    \end{tabular}
    \end{minipage}
\end{table}

\section{Representative Model Configurations}
\label{app:model_family}

Table~\ref{tab:model_family} reports the exact configurations visualized in Fig.~\ref{fig:model_family}. The proxy is used for hyperparameter tuning, the scaling family supports the compute-optimal analyses, and the final model is selected for full training.

\begin{table}[t]
\centering
\small
\setlength{\tabcolsep}{5pt}
\renewcommand{\arraystretch}{1.1}
\caption{
Representative \ours{} configurations used in our experiments. All models instantiate the architecture of Sec.~\ref{sec:model_arch}; structural ratios (MoE experts, KDA-to-MLA ratio, residual streams, KV-compression ratio, patchification) are held fixed across scales (Sec.~\ref{sec:hp_transfer}). We report both per-token \emph{activated} parameters and total denoiser parameters, where total parameters count all routed MoE experts. The full sweep contains additional intermediate sizes.
}
\label{tab:model_family}
\begin{tabular*}{\linewidth}{@{\extracolsep{\fill}}lccccc@{}}
\toprule
\textbf{Model} & \textbf{Act. Params} & \textbf{Total Params} & \textbf{Width} & \textbf{FFN Hidden} & \textbf{Depth} \\
\midrule
Proxy          & 22M    & 59M    & 512   & 2048  & 4  \\
\midrule
\multirow{17}{*}{\shortstack[l]{Scaling\\family}}
               & 55M    & 253M   & 384   & 1536  & 16 \\
               & 200M   & 893M   & 896   & 2432  & 16 \\
               & 291M   & 1.35B  & 1024  & 4096  & 16 \\
               & 393M   & 1.88B  & 1152  & 5120  & 16 \\
               & 482M   & 2.46B  & 896   & 4096  & 32 \\
               & 748M   & 3.58B  & 1280  & 4096  & 32 \\
               & 873M   & 4.41B  & 1280  & 5120  & 32 \\
               & 999M   & 5.25B  & 1280  & 6144  & 32 \\
               & 1.55B  & 7.99B  & 1664  & 6656  & 32 \\
               & 2.01B  & 9.93B  & 2048  & 6400  & 32 \\
               & 2.26B  & 10.33B & 2432  & 5248  & 32 \\
               & 2.50B  & 11.92B & 2432  & 6400  & 32 \\
               & 2.65B  & 10.79B & 2944  & 4736  & 32 \\
               & 2.80B  & 14.26B & 2304  & 9216  & 32 \\
               & 2.93B  & 14.25B & 2560  & 7296  & 32 \\
               & 2.99B  & 15.09B & 2432  & 9088  & 32 \\
               & 4.02B  & 19.31B & 3072  & 8448  & 32 \\
\midrule
Final Training Model           & 2.22B  & 11.28B & 2048  & 8192  & 32 \\
\bottomrule
\end{tabular*}
\end{table}

\section{Visual RoPE Analysis: Protocol and Results}
\label{app:rope_probe}

\begin{figure*}[h]
    \centering
    \begin{subfigure}[b]{0.345\linewidth}
        \centering
        \includegraphics[width=\linewidth]{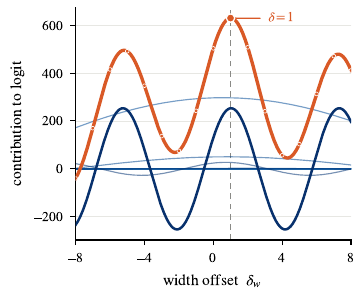}
        \caption{}
        \label{fig:rope_visual_wan_decomp}
    \end{subfigure}
    \hfill
    \begin{subfigure}[b]{0.345\linewidth}
        \centering
        \includegraphics[width=\linewidth]{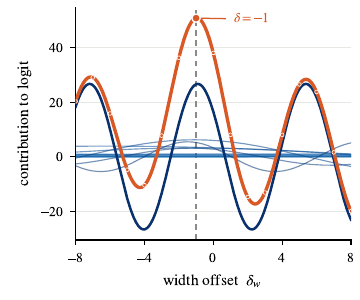}
        \caption{}
        \label{fig:rope_visual_flux_decomp}
    \end{subfigure}
    \hfill
    \begin{subfigure}[b]{0.285\linewidth}
        \centering
        \includegraphics[width=\linewidth]{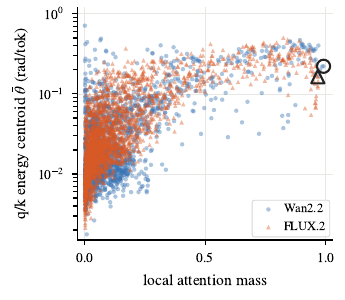}
        \caption{}
        \label{fig:rope_visual_division}
    \end{subfigure}
    \\[2pt]
    \begin{subfigure}[b]{0.49\linewidth}
        \centering
        \includegraphics[width=\linewidth]{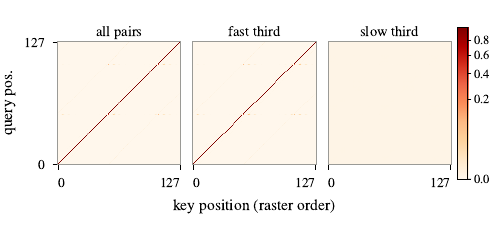}
        \caption{}
        \label{fig:rope_visual_wan_bands}
    \end{subfigure}
    \hfill
    \begin{subfigure}[b]{0.49\linewidth}
        \centering
        \includegraphics[width=\linewidth]{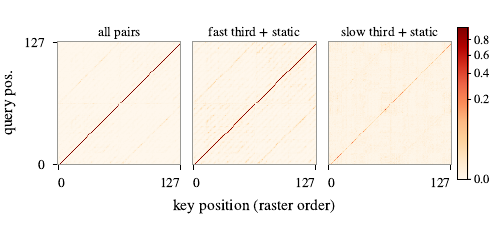}
        \caption{}
        \label{fig:rope_visual_flux_bands}
    \end{subfigure}
    \caption{
    \textbf{RoPE frequency usage in visual diffusion Transformers.}
    \textbf{(a)}~Per-pair logit decomposition of the strongest local head of Wan2.2 (layer 33, head 30) along the width axis: one curve per width-axis rotary pair (darker blue indicates faster rotation), whose sum (orange) peaks at offset $+1$, the raster-order previous token; pairs on other axes add offset-independent constants (not shown).
    \textbf{(b)}~The same decomposition for the strongest local head of FLUX.2 (layer 46, head 33), peaking at the raster-order next token ($\delta_w=-1$).
    \textbf{(c)}~Energy-weighted rotation speed of the query-key channels versus local attention mass, for every head of both models: local heads concentrate their energy at fast rotations, far-field heads at near-zero rotations (black outlines mark the heads from (a) and (b)).
    \textbf{(d)}~Attention of the Wan2.2 local head on a $128$-token raster crop, recomputed from frequency subsets: the fastest third of each axis's pairs reproduces the local pattern, the slowest third erases it.
    \textbf{(e)}~The same test for the FLUX.2 local head; here the fast pairs are combined with the axes that FLUX.2 leaves unrotated for image tokens, which carry the head's content gating.
    }
    \label{fig:rope_visual}
\end{figure*}

This appendix details the FLUX.2 and Wan2.2 analysis summarized in Sec.~\ref{sec:attention}. All statistics are computed from the models' own sampling trajectories, so the probed activations match the regime in which the models operate. Fig.~\ref{fig:rope_visual} visualizes the per-pair decompositions, the frequency-band ablations, and the division of labor across all heads.

\paragraph{\textbf{Models and Sampling.}}
We probe the public weights of Wan2.2-T2V-A14B and FLUX.2-dev. Wan2.2 has $40$ layers with $40$ self-attention heads of dimension $128$; axial RoPE splits each head into $22$ temporal, $21$ height, and $21$ width channel pairs with per-axis frequencies $\theta_j = 10000^{-2j/d_a}$, spanning $1$ down to ${\sim}1.5\times10^{-4}$ rad/token. Its two denoising experts (applied above and below $t=875$) are probed separately. FLUX.2 has $8$ double-stream and $48$ single-stream blocks with $48$ heads of dimension $128$; its axial RoPE assigns $16$ pairs to each of four axes (time, height, width, text index) with base $2000$, spanning $1$ down to ${\sim}8\times10^{-4}$ rad/token. Image tokens carry coordinates only on the height and width axes, so the time- and text-axis channels, half of every head, are unrotated between image tokens; text tokens attend in the same sequence and use the text axis. We sample Wan2.2 at $480\times832$ with $33$ frames (a $9\times30\times52$ token grid, $14{,}040$ visual tokens) using UniPC with $40$ steps and classifier-free guidance $4.0/3.0$, and FLUX.2 at $1024^2$ (a $64\times64$ grid) using its $50$-step flow schedule with embedded guidance $4.0$. Each model is probed on six shared prompts covering people, animals, nature, urban scenes, close-ups, and still-life motion, at three denoising timesteps ($t=\{964,750,347\}$ for Wan2.2, spanning both experts; noise levels $\sigma\approx\{0.99,0.88,0.46\}$ for FLUX.2), giving $18$ settings per model.

\paragraph{\textbf{Capture and Head Statistics.}}
At each captured step we record, for every self-attention layer, the per-head queries and keys after query-key normalization and before rotation (for Wan2.2, on the conditional branch of classifier-free guidance). For a fixed random subset of $512$ visual query tokens, shared across all layers and settings, we form the full attention rows and measure per head: the \emph{local attention mass} on the immediate spatio-temporal neighborhood, \textit{i.e.}, the fraction of a query's attention falling on tokens within Chebyshev distance $1$ on the token grid, self excluded ($26$ neighbors for video, $8$ for images); the \emph{far-field mass} at Chebyshev distance $\geq5$; the mean attended distance; and the rate at which the strongest non-self key is a neighbor. For FLUX.2, attention rows span the concatenated text and image tokens, and we additionally record the mass on text keys.

\paragraph{\textbf{Frequency Bands and Head Selection.}}
Within each positional axis we sort the channel pairs by rotation speed and split them into fastest, middle, and slowest thirds; for FLUX.2 the unrotated time- and text-axis channels form an additional static group. Band-restricted attention is obtained by recomputing the logits from a channel subset, keeping the $1/\sqrt{d}$ scale, and re-applying the softmax. For each model we select the eight heads with the highest local attention mass averaged over all $18$ settings; the ablation numbers below average over these heads and settings.

\paragraph{\textbf{Local Heads and Position Selection.}}
Both models contain sharp \emph{local heads}, the visual analogue of previous-token heads. In Wan2.2, layer $33$, head $30$ --- at the same position in both denoising experts, whose top-eight local-head lists share seven of eight entries --- attends to the immediately preceding token of the raster scan: its local attention mass is $0.99$, and $98.1\%$ of queries place their argmax exactly there. This rate equals the fraction of queries that do not start a row, \textit{i.e.}, selection is exact up to grid boundaries. In FLUX.2, head $33$ of layer $46$, a late single-stream block, mirrors this with the immediately following token ($98.4\%$ of queries, again saturating the boundary limit; local attention mass $0.93$). Each remains the top-ranked local head of its layer in all $18$ settings, indicating that locality is a stable head property rather than an artifact of a particular input. The per-pair decomposition traces this selection to the fast end of the spectrum, exactly as in Qwen3-4B: we average each head's pre-rotation queries and keys over all visual tokens and express every channel pair by its amplitude $\|q^{(i)}\|\|k^{(i)}\|$, phase, and frequency; the summed width-axis contributions peak at offset one (Figs.~\ref{fig:rope_visual_wan_decomp} and~\ref{fig:rope_visual_flux_decomp}), pairs on other axes contribute offset-independent constants, and the pairwise sum matches the directly rotated dot product to within $3{\times}10^{-12}$ (Wan2.2) and $3{\times}10^{-14}$ (FLUX.2) in float64. The band ablations confirm the attribution. In Wan2.2, retaining only the fastest third of each axis's pairs preserves the selected heads' local attention mass ($0.944$ versus $0.967$ in the low-noise expert; $0.934$ versus $0.957$ in the high-noise expert), while retaining only the slowest third collapses it to $0.002$ (Fig.~\ref{fig:rope_visual_wan_bands}). In FLUX.2, the unrotated channels select which tokens to attend to by content while the fast pairs place the peak at the target offset: attention recomputed from the two groups together matches the full heads almost exactly (cosine similarity $0.994$; local mass $0.776$ versus $0.926$), whereas either group alone fails ($0.190$ from the fast pairs alone, $0.051$ without them; Fig.~\ref{fig:rope_visual_flux_bands}).

\paragraph{\textbf{Far-Field Heads and Semantic Matching.}}
The slowly rotating channels serve the opposite role. Using full per-token dumps at one setting per model, we measure each head's per-pair energy $E_i = \mathbb{E}_t\,\|q^{(i)}_t\|\,\|k^{(i)}_t\|$ and summarize its spectrum by the energy-weighted geometric mean of the rotating pairs' frequencies. Heads whose attention lies far outside the local neighborhood (far-field mass $>0.8$; mean attended distance of ${\sim}20$ tokens in Wan2.2 and ${\sim}30$ in FLUX.2) concentrate their energy at near-zero rotation speeds: their median centroid is $0.012$ rad/token in Wan2.2 and $0.013$ in FLUX.2, more than $20\times$ slower than that of local heads (local mass $>0.5$; medians $0.26$ and $0.25$, respectively), so their dot products turn by only a few degrees over the attended range (Fig.~\ref{fig:rope_visual_division}). FLUX.2 makes the division explicit: its far-field heads park half of their energy (median $51\%$, versus $13\%$ for local heads) in the unrotated channels, and the slow channels alone (the slowest third plus the unrotated group) reproduce their attention almost exactly (cosine similarity $0.965$).

\end{document}